%% file: main.tex
\documentclass{article}

\usepackage{microtype}
\usepackage{graphicx}
\usepackage{subfigure}
\usepackage{booktabs} 

\usepackage{hyperref}

\usepackage[accepted]{icml2023} 

\usepackage{amsmath}
\usepackage{amssymb}
\usepackage{mathtools}
\usepackage{amsthm}
\input{Mis/Packages}

\usepackage[capitalize,noabbrev]{cleveref}

\theoremstyle{plain}
\newtheorem{theorem}{Theorem}[section]

\theoremstyle{definition}

\theoremstyle{remark}
\newtheorem{remark}[theorem]{Remark}

\usepackage[textsize=tiny]{todonotes}

\icmltitlerunning{A Reinforcement Learning Framework for Dynamic Mediation Analysis}

\input{Mis/Command}

\begin{document}

\twocolumn[
\icmltitle{A Reinforcement Learning Framework for Dynamic Mediation Analysis}




\begin{icmlauthorlist}
\icmlauthor{Lin Ge}{1}
\icmlauthor{Jitao Wang}{2}
\icmlauthor{Chengchun Shi}{3}
\icmlauthor{Zhenke Wu}{2}
\icmlauthor{Rui Song}{1}
\end{icmlauthorlist}

\icmlaffiliation{1}{North Carolina State University}
\icmlaffiliation{2}{University of Michigan, Ann Arbor}
\icmlaffiliation{3}{London School of Economics and Political Science}

\icmlcorrespondingauthor{Rui Song}{rsong@ncsu.edu}

\icmlkeywords{MDP, Off-Policy, Mediator}

\vskip 0.3in
]



\printAffiliationsAndNotice{}  

\begin{abstract}
Mediation analysis learns the causal effect transmitted via mediator variables between treatments and outcomes, and receives increasing attention in various scientific domains to elucidate causal relations. 
Most existing works focus on point-exposure studies where each subject only receives one treatment at a single time point. 
However, there are a 
number of applications (e.g., mobile health) where the treatments are sequentially assigned over time and the dynamic mediation effects are of primary interest. 
Proposing a reinforcement learning (RL) framework, we are the first to evaluate dynamic mediation effects in settings with infinite horizons. 
We decompose the average treatment effect into an immediate direct effect, an immediate mediation effect, a delayed direct effect, and a delayed mediation effect. Upon the identification of each effect component, we further develop robust and semi-parametrically efficient estimators under the RL framework to infer these causal effects. 
The superior performance of the proposed method is demonstrated through extensive numerical studies, theoretical results, and an analysis of a mobile health dataset. A Python implementation of the proposed procedure is available at https://github.com/linlinlin97/MediationRL.
\end{abstract}

 \input{1_Introduction}
 
\input{2_Related_Work}

\input{3_Data_and_Problem_Formulation}

\input{4_OPE}

\input{Appendix/Learning_Nuisance_Function}
\input{5_Theory_robust}
\input{6_Toy_Example}
\input{7_Real}
\input{8_Discussion}
\section*{Acknowledgements}
The research is partially supported by a grant from the NSF (DMS-2003637), a grant from EPSRC (EP/W014971/1), grants from the National Institutes of Health (R01 MH101459 to ZW; R01 NR013658 to JW $\&$ ZW), and an investigator grant from Precision Health Initiative at the University of Michigan to ZW. We thank Dr. Srijan Sen for generous support in the IHS data access.
\newpage
\bibliography{mycite}
\bibliographystyle{icml2023}

\newpage
\appendix
\onecolumn
\input{Appendix/Graphical_PO}

\input{Appendix/Alternative_Decomp}
\input{Appendix/Theorem1}
\input{Appendix/Robustness}
\input{Appendix/Efficiency}
\input{Appendix/Derivation}
\input{Appendix/Toy_Examples}
\input{Appendix/optimal_policy}
\input{Appendix/Additional_Experiments}

\end{document}

%% file: Mis/Packages.tex
\usepackage{mathtools}
\usepackage{enumitem}   

%% file: Mis/Command.tex
\newcommand{\blind}{0}

\newcommand{\Mean}{{\mathbb{E}}}

\newcommand{\prob}{{\mbox{Pr}}}

\usepackage{xr}

\newcommand{\mA}{\mathcal{A}}
\newcommand{\real}{\mathbb{R}}

\newcommand{\indep}{\perp \!\!\! \perp}

%% file: 1_Introduction.tex
\section{Introduction}\label{intro}
Mediation analysis aims to understand the causal pathway from an exposure (e.g., treatment or action) to an outcome variable of interest. It is gaining increasing popularity recently and has been frequently employed in a number of domains including epidemiology 
\citep{richiardi2013mediation,rijnhart2021mediation}, psychology \citep{rucker2011mediation}, 
genetics \citep{chakrabortty2018inference,zeng2021statistical, djordjilovic2022optimal}, 
economics \citep{celli2022causal} and neuroscience \citep{li2022sequential,shi2022testing}. 

Our paper is motivated by the need to 
learn the dynamic mediation effects in sequential decision making. 
One motivating example is given by the Intern Health Study \citep[IHS,][]{necamp2020assessing}, which focuses on sequential mobile health interventions to help improve the mental health of medical interns who work in stressful environments. Participants were randomly assigned to receive notifications (e.g., tips and insights) throughout the study. For example, some notifications remind participants to take a break or enjoy a tasty treat, while others summarize the trends of recent physical activity and sleep. All the notifications are designed to improve participants' mood scores (self-reported via a custom-made study App) either directly or indirectly through increased activity or sleep hours. In addition, it is essential to note that participants' recent behavior will not only influence their proximal mood but will also influence their behavior and mood scores in the following days.
To design a more effective intervention policy in IHS, it is necessary to understand how mobile prompts impact mood scores. In particular, the mobile prompts may directly impact the mood scores or encourage more physical activity and sleep, which may then impact the mood scores. In addition, an individual's past treatment sequence and behavior trajectory may impact the mood score. Teasing out these distinct sources of causal impacts on mood scores and their relative magnitudes needs new definitions, identification results, and inferential methods.

A fundamental question considered in this paper is how to infer the dynamic mediation effects in the aforementioned applications. Solving this question raises at least three challenges. First, the mediator at a given time affects both the current and future outcomes, inducing temporal carryover effects. As demonstrated in the case study in Section \ref{real}, the delayed direct 
 effect (DDE) and the delayed mediator effect (DME) are significant and dominate the average treatment effect for the intervention policy used in the IHS \citep{sen2010prospective,necamp2020assessing}. In contrast, the immediate direct effect (IDE) and immediate mediator effect (IME) are both insignificant. Nonetheless, most existing mediation analyses focus on estimating the indirect effect on the immediate reward and are hence inappropriate to our application. Second, the horizon (e.g., number of decision stages) in the aforementioned applications is typically very long or diverges with the sample size. Existing solutions developed in finite horizon settings typically suffer from the curse of horizon in the sense that the variances of the proposed estimators grow exponentially fast with respect to the horizon \citep{liu2018breaking} and are hence inapplicable; see Section \ref{sec:related} for details. Third, regardless of how the dynamic effects may change during the sequential treatments (or lack thereof), most works focus on examining the causal effects on the final outcome obtained at the end of the treatment process. However, in the context of behavioral change, the goal is to encourage and maintain small improvements to nudge individuals into generating sustained improvements in outcomes like mood scores. Currently, there is a dearth of methods to analyze causal effects for outcomes measured at every decision point in the sequence.

\begin{figure}[t]
\begin{center}
\centerline{\includegraphics[width=.75\columnwidth]{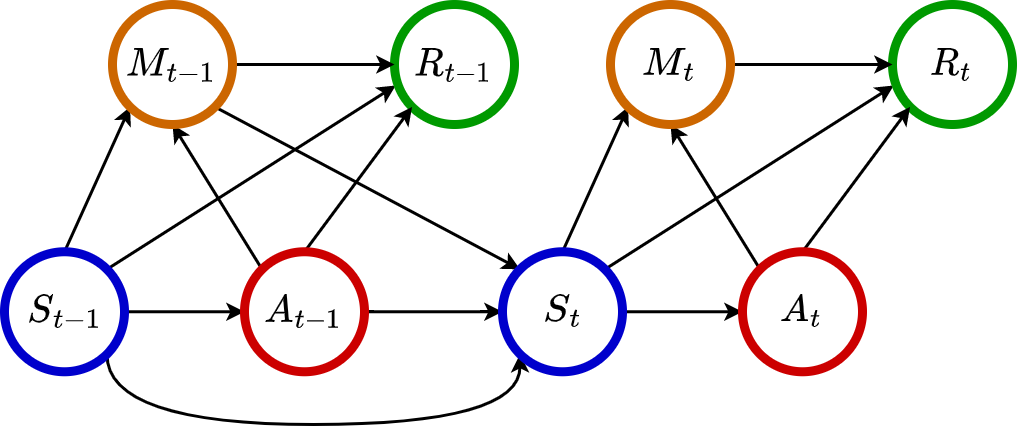}}
\vskip -0.15in
\caption{Mediated MDP.}
\label{fig:Proposed MDP}
\end{center}
\vskip -0.35in
\end{figure}

To address these limitations, we propose formulating the evaluation of dynamic mediation effects as a reinforcement learning (RL) problem. In particular, we use the Markov decision process (MDP) that is commonly employed in RL to model the mediated dynamic decision process over an infinite time horizon. Building upon the standard MDP, we introduce four additional sets of causal relationships, including state-mediator, action-mediator, mediator-state, and mediator-reward, as shown in Figure~\ref{fig:Proposed MDP}. To evaluate the effects of different treatment policies, we consider using the off-policy evaluation \citep[OPE,][]{dudik2014doubly,uehara2022review}, which is widely used to avoid the difficulty of rerunning trials by evaluating treatment policies based on observational data.


\textbf{Contributions.} The main contributions are as follows. Motivated by the mobile health applications, we first construct the mediation analysis within the framework of RL over an infinite time horizon. Second, we propose to decompose the average treatment effect between a target policy and a control policy into IDE, IME, DDE, and DME. While IDE and IME have been extensively studied in single-stage settings, we introduce the DDE and DME to quantify the carryover effects of past actions and mediators. 
Third, upon the identification result of each effect component, multiply-robust estimators are developed. In particular, each proposed estimator is consistent even when models such as mediator distribution and reward distribution are misspecified (See Section \ref{toy_1}). Furthermore, we theoretically show the semiparametric efficiency of the proposed estimators and confirm the theoretical prediction using numerical studies. Lastly, we conclude by analyzing the IHS data and providing new insights into guiding future designs of these behavioral interventions.

%% file: 2_Related_Work.tex
\section{Related Work}\label{sec:related}





Mediation analysis is widely studied in point-exposure studies under the classical structure consisting of a treatment, a mediator, and an outcome \citep{robins1992identifiability,pearl2022direct,petersen2006estimation,van2008direct,imai2010general,tchetgen2012semiparametric,tchetgen2014estimation,vanderweele2015explanation}, decomposing the average treatment effect into direct effect and indirect effect. Recently, to address commonly observed intermediate confounders that would be affected by the exposure and then affect both mediator and outcome, multiple methods have been developed to extend the classical mediation analysis \citep{robins2010alternative,tchetgen2014identification,vanderweele2014effect,vansteelandt2017interventional,diaz2021nonparametric,diaz2022causal}, among which the random intervention (RI)-based approach \citep{vanderweele2014effect,diaz2022causal} further sets the foundation for the recent advancement of longitudinal mediation analysis. 

There is a rich literature on longitudinal mediation analysis with no intermediate confounders \citep{selig2009mediation, roth2013mediation}. See also \citet{preacher2015advances} for a detailed review. However, time-varying intermediate confounders are ubiquitous in longitudinal data contexts. For example, in the IHS, doing exercises may result in a good mood, which may, in turn, increase the likelihood of engaging in more activities the next day and then subsequently affect the mood that follows. 

In the presence of time-varying intermediate confounders, there are two major RI-based approaches. 
\citet{vanderweele2017mediation} and 
\citet{diaz2022efficient} proposed to intervene in the mediator sequence by randomly drawing mediators from the corresponding \textit{marginal} distribution and defined the longitudinal interventional indirect/direct effect, which is different from the natural effect decomposition. Our work is primarily related to the work of \citet{zheng2017longitudinal}, which proposed to intervene in the mediator by randomly drawing the mediator from its \textit{conditional} distribution and provided a natural decomposition of the total effect. Using the efficient influence function (EIF), 
they developed a multiply-robust estimator with less reliance on the correct model specification. However, all the aforementioned methods only focused on the treatment impact on the final outcome in finite horizons and did not consider immediate outcomes or infinite horizon settings. In addition,  the estimator developed by \citet{zheng2017longitudinal} is based on the product of importance sampling ratios at all time points and suffers from the curse of horizon. \citet{zheng2012causal} also analyzed the longitudinal mediation effect by drawing mediators from \textit{conditional} distribution but with a focus on single-exposure settings.
Using an RL 
framework for dynamic mediation analysis over an infinite horizon, our work is also connected to the line of research on OPE. Existing OPE-related research evaluates the discounted sum of rewards or average rewards for a target policy using observational data gained by following a different behavior policy. 
In general, there are three types of estimation procedures. The first is known as the direct method \citep[DM,][]{le2019batch,feng2020accountable,luckett2020estimating, hao2021bootstrapping, liao2021off, Chen2022on,shi2020statistical}, which directly learns Q-functions and obtains value estimates based on their estimators. The second category of approaches utilizes importance sampling \citep[IS,][]{precup2000eligibility,thomas2015high, hallak2017consistent, hanna2017bootstrapping,liu2018breaking,xie2019towards, dai2020coindice, zhang2020gendice}, which re-weights the rewards to eliminate the bias due to distributional shift. 
The third category develops doubly robust (DR) estimators by appropriately integrating DM with IS estimators \citep{jiang2016doubly,thomas2016data,farajtabar2018more,liao2020batch,Tang2020Doubly,uehara2020minimax,kallus2022efficiently}. DR estimators are also known to achieve the semiparametric efficiency bound \citep{bickel1993efficient}. However, none of the above papers studied mediation analysis. Recently, \citet{shi2022off} proposed a consistent DR estimator for OPE in the presence of unmeasured confounders with the help of a mediator variable, which is used to intercept each directed path from treatments to reward/state. Our paper differs from theirs in that we decompose the off-policy value into the sum of IDE, IME, DDE, and DME and focus on settings without unmeasured confounding. 




%% file: 3_Data_and_Problem_Formulation.tex
\section{Preliminaries}

\subsection{Data Generating Process}
We consider the observational data generated from a mediated Markov decision process (MMDP), as illustrated in Figure~\ref{fig:Proposed MDP}. Suppose there exists an agent that tries to learn from the data and interact with a given environment. At each time $t$, the environment arrives at a state $S_t \in \mathcal{S}$, and the agent selects an action $A_t  \in \mA = \{0,1,\cdots, K-1\}$ according to a behavior policy $\pi_b(\bullet|S_t)$. Building upon the usual MDP, to further analyze the mediation effect, we consider an immediate mediator variable $M_t \in \mathcal{M}$ drawn according to $p_m(\bullet|S_t,A_t)$, which mediates the effect of $A_t$ on the environment. Subsequently, the agents would receive an immediate $R_t$ and the the environment transits to a next-state $S_{t+1}$ according to $p_{s',r}(\bullet,\bullet|S_t,A_t,M_t)$. 
Both $\mathcal{S}$ and $\mathcal{M}$ are finite dimensional vector spaces. 
To summarize, the observed data sequences consist of the state-action-mediator-reward tuples $(S_t,A_t,M_t,R_t)_{t\ge 0}$ satisfying the following Markov assumption: 
$(M_t,R_t,S_{t+1})\indep (S_j,A_j,M_j,R_j)_{j<t}|(S_t,A_t)$ for any $t$.


\subsection{Problem Formulation}
Let $N$ denote the number of trajectories. The $i$th trajectory contains 
$\{(S_{i,t},A_{i,t},M_{i,t},R_{i,t})\}_{1\le i \le N, 0\le t \le T}$ where $T$ is the termination time. 
We assume that all these trajectories are i.i.d. and follow the MMDP. Let $\pi$ 
denote a generic (stationary) policy which maps from $\mathcal{S}$ to a probability mass function on $\mA$, and $\Mean^{\pi}[\cdot]$ denote the expectation of a random variable under 
the policy $\pi$. 
Based on the observed data, our goal is to analyze the average treatment effect (ATE) of a target policy $\pi_e$ relative to a control policy $\pi_0$, 
given by
\begin{align*}
\textrm{ATE($\pi_e,\pi_0$)}=\lim_{T\to \infty} \frac{1}{T}\sum_{t=0}^{T-1} \textrm{TE}_t\textrm{($\pi_e,\pi_0$)},
\end{align*}where $\textrm{TE}_t\textrm{($\pi_e,\pi_0$)} = \Mean^{\pi_e}[R_t] - \Mean^{\pi_0}[R_t]$.


To gain a better understanding of the mediated and delayed effects, we consider decomposing $\textrm{TE}_t(\pi_e,\pi_0)$ into
\begin{equation*}
\textrm{IDE}_t(\pi_e,\pi_0)+\textrm{IME}_t(\pi_e,\pi_0)+\textrm{DDE}_t(\pi_e,\pi_0)+\textrm{DME}_t(\pi_e,\pi_0).
\end{equation*} Averaging over $t$ for each component, we obtain a four-way decomposition of ATE as IDE + IME + DDE + DME. We formally define each of these effects in the next section.


\section{Effect Decomposition} 
This section begins with a decomposition of $\textrm{TE}_t(\pi_e,\pi_0)$, from which we define each component in $\textrm{ATE}(\pi_e,\pi_0)$. 
We first notice that $\textrm{TE}_t$ can be decomposed into two components: i) the immediate treatment effect ($\textrm{ITE}_t$) measuring the impact of 
the current action-mediator pair $(A_t,M_t)$ on the immediate outcome $R_t$; 
ii) the delayed treatment effect ($\textrm{DTE}_t$) that measures the carryover effects of 
the historical action-mediator sequences $(A_j,M_j)_{j<t}$ 
on $R_t$ that pass through $S_t$. 

We next consider $\textrm{ITE}_t$. Let $\pi^t_{e,0}$ denote a nonstationary policy 
that follows $\pi_e$ at the first $t-1$ steps and then follows $\pi_0$ at $t$. Mathematically, 
$\textrm{ITE}_t$ is defined as $\Mean^{\pi_e}[R_t] - \Mean^{\pi^t_{e,0}}[R_t]$. Notice that $\pi^t_{e,0}$ differs from the stationary policy $\pi_e$ only at the current time $t$, then $\textrm{ITE}_t$ indeed measures the immediate effect. Under the Markov assumption, we obtain
\begin{align*}
    \Mean^{\pi_e}[R_t]&=\sum_{s,a,m}p_t^{\pi_e}(s)\pi_e(a|s)p_m(m|s,a)r(s,a,m),\\
    \Mean^{\pi^t_{e,0}}[R_t]&=\sum_{s,a,m} p_t^{\pi_e}(s)\pi_0(a|s) p_m(m|s,a) r(s,a,m),
\end{align*}
where $p_t^{\pi}(s)$ denotes the distribution of $S_t$ under a policy $\pi$, and $r(\bullet,\bullet,\bullet)$ denotes the conditional expectation of the reward given the state-action-mediator triplet.

\begin{figure}[t]
\begin{center}
\centerline{\includegraphics[width=.75\columnwidth]{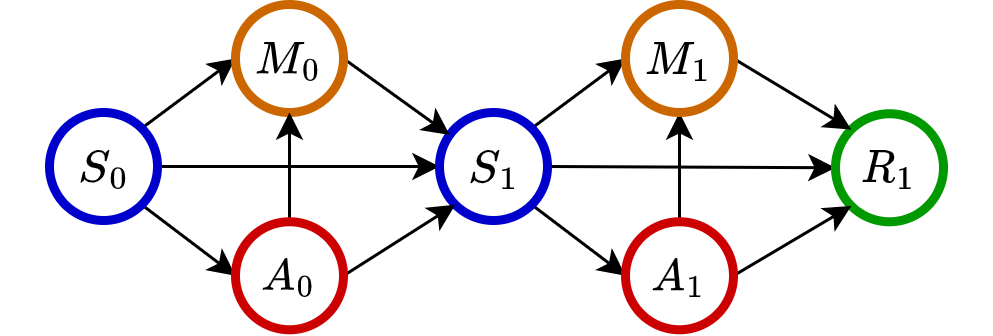}}
\vskip -0.05in
\caption{Causal paths from actions to reward received in $t=1$.}
\label{fig: 2-stage MMDP}
\end{center}
\vskip -0.35in
\end{figure}

Notice that $A_t$ has both a direct effect and an indirect effect (mediated by $M_t$) on $R_t$. This motivates us to further decompose $\textrm{ITE}_t$ into $\textrm{IDE}_t$ and $\textrm{IME}_t$. Let $G^t_e$ denote the process in which $\pi_e$ is applied at the first $t-1$ steps to generate $S_t$, $A_t$ is then generated according to $\pi_0$, and $M_t$ is 
generated as if $A_t$ were assigned according to $\pi_e$, i.e., 
\begin{eqnarray}\label{eqn:drawM}
    M_t\sim \sum_a p_m(\bullet|a,S_t)\pi_e(a|S_t).
\end{eqnarray}
Then, $\Mean^{G^t_e}[R_t]$ equals
\begin{equation*}
    \sum_{s,a,m}p_t^{\pi_e}(s)\pi_e(a|s)p_m(m|s,a)\sum_{a'}\pi_0(a'|s)r(s,a',m).
\end{equation*}
It follows that 
\begin{align*}
    \textrm{ITE}_t = \underbrace{\Mean^{\pi_e}[R_t] - \Mean^{G^t_e}[R_t]}_{\textrm{IDE}_t(\pi_e,\pi_0)} + \underbrace{\Mean^{G^t_e}[R_t]-\Mean^{\pi^t_{e,0}}[R_t]}_{\textrm{IME}_t(\pi_e,\pi_0)}.
\end{align*} 
By definition, the $\textrm{IDE}_t$ quantifies the direct treatment effect 
on the proximal outcome $R_t$ whereas the $\textrm{IME}_t$ evaluates 
the indirect effect mediated by $M_t$. As an illustration, 
set $t=1$ and consider Figure~\ref{fig: 2-stage MMDP}. $\textrm{IDE}_1$ measures the causal effect along the path $A_1\to R_1$ whereas $\textrm{IME}_1$ corresponds to the effect along the path $A_1\to M_1 \to R_1$.

Next, we consider $\textrm{DTE}_t$, defined as $\Mean^{\pi^t_{e,0}}[R_t] - \Mean^{\pi_0}[R_t]$. By definition, $\pi_{e,0}^t$ differs from $\pi_0$ at the first $t-1$ time points. As such,  $\textrm{DTE}_t$ characterizes the delayed treatment effects on the current outcome $R_t$. 
Similarly, we further decompose $\textrm{DTE}_t$ into the sum of direct and mediation effects. 
To characterize the delayed mediation effects, we follow the RI-based approach developed by \citet{zheng2017longitudinal}. Specifically,  consider a 
stochastic process in which at the first $t-1$ time steps, the action is selected according to $\pi_0$, and the mediator is drawn assuming the action is assigned according to $\pi_e$ (see Equation \ref{eqn:drawM}), whereas at time $t$, the system follows $\pi_0$. We provide more details on this process in Appendix \ref{graphical PO}. Let $G_0^t$ denote the resulting process and $\Mean^{G^t_0}[R_t]$ the expected value of $R_t$ generated according to $G_0^t$. 
This allows us to decompose $\textrm{DTE}_t$ as follows, 
\begin{align*}
    \textrm{DTE}_t = \underbrace{\Mean^{\pi^t_{e,0}}[R_t] - \Mean^{G^t_0}[R_t]}_{\textrm{DDE}_t(\pi_e,\pi_0)} + \underbrace{\Mean^{G^t_0}[R_t]-\Mean^{\pi_{0}}[R_t]}_{\textrm{DME}_t(\pi_e,\pi_0)}.
\end{align*}
Notice that in the three processes, the action selection and mediator generation mechanisms at time $t$ are the same. As such, both $\textrm{DDE}_t$ and $\textrm{DME}_t$ characterize the delayed effects. At the first $t-1$ time steps, the action selection mechanism between $G_0^t$ and the process generated by $\pi_{e,0}^t$ are different whereas both processes have the same mediator generation mechanism. As such, $\textrm{DDE}_t$ quantifies how past actions directly impact the current outcome. On the contrary, $G_0^t$ and the process generated by $\pi_0$ have the same action selection mechanism. They differ in the way the mediator is generated. Hence, $\textrm{DME}_t$ measures the indirect past treatment effects mediated by $\{M_j\}_{j<t}$. To elaborate, let us revisit 
Figure~\ref{fig: 2-stage MMDP}. $\textrm{DDE}_1$ captures the causal effect along the path $A_0\to S_1\to\{A_1, M_1\}\to R_1$ whereas
$\textrm{DME}_1$ considers the path
$A_{0}\to M_{0}\to S_1\to\{A_1, M_1\}\to R_1$.



We also remark that the proposed effects are consistent with those in the existing literature. Specifically, when specialized to state-agnostic policies, $\textrm{IDE}_0$ and $\textrm{IME}_0$ are reduced to the total direct effect and the pure indirect effect \citep{robins1992identifiability} in single-stage decision-making. Meanwhile, 
$\textrm{DDE}_t$ and $\textrm{DME}_t$ are 
similar to those proposed by \citet{zheng2017longitudinal} developed in finite horizons.

Based on these effects, by aggregating $\textrm{IDE}_t$,  $\textrm{IME}_t$, $\textrm{DDE}_t$ and $\textrm{DME}_t$ over time, we obtain the following four-way decomposition of $\textrm{ATE}(\pi_e,\pi_0)$, 
\begin{align*}
    \underbrace{\eta^{\pi_e} - \eta^{G_e}}_{\textrm{IDE}(\pi_e,\pi_0)} + \underbrace{\eta^{G_e}-\eta^{\pi_{e,0}}}_{\textrm{IME}(\pi_e,\pi_0)}+\underbrace{\eta^{\pi_{e,0}} - \eta^{G_0}}_{\textrm{DDE}(\pi_e,\pi_0)} + \underbrace{\eta^{G_0}-\eta^{\pi_{0}}}_{\textrm{DME}(\pi_e,\pi_0)},
\end{align*} where $\eta^{\pi}$ is the average reward of policy $\pi$.

Finally, we remark that to simplify the presentation, we choose not to use the potential outcome framework \citep{rubin2005causal} to formulate these causal effects of interest in this section. 
The detailed potential outcome definitions are relegated to Appendix \ref{graphical PO}. In addition, we 
show that these potential outcomes are identifiable and summarize the results in the following theorem.
\begin{theorem}[Identification]\label{thm:identification}
Under standard assumptions including consistency, sequential randomization and positivity \citep{zheng2017longitudinal,luckett2019estimating}, 
$\textrm{IDE}(\pi_e,\pi_0)$, $\textrm{IME}(\pi_e,\pi_0)$, $\textrm{DDE}(\pi_e,\pi_0)$, and $\textrm{DME}(\pi_e,\pi_0)$ are all identifiable. 
\end{theorem}
We refer readers to Appendix \ref{proof:identification} for more details. 




%% file: 4_OPE.tex
\section{Dynamic Treatment Effects Evaluation}
In this section, we first develop DM and IS estimators for each defined dynamic treatment effect, whose consistencies require 
a given set of nuisance functions to be correctly specified. 
This motivates us to further develop doubly or triply robust estimators in section \ref{TRE}, whose consistencies only require either one of the two or three sets of nuisance functions to be correctly specified. 
Finally, we discuss the estimation methods for nuisance functions.
\subsection{Direct Method Estimators (DM)}
The direct estimators are built upon the Q-functions. 
For $\pi \in \{\pi_e,\pi_0\}$, we first define the conditional relative value function $Q^{\pi}(s,a,m)$ as
\begin{equation}\label{Q:pi_e}
\sum_{t\ge 0} \Mean^{\pi} [R_{t}-\eta^{\pi}|S_0=s, A_0=a, M_0=m],
\end{equation} which measures the expected total difference between rewards and the average reward of policy $\pi$, given that the initial state-action-mediator triplet 
equals $(s,a,m)$. 
Notably, (\ref{Q:pi_e}) deviates slightly from the standard definition in MDPs (i.e., $\sum_{t\ge 0} \Mean^{\pi_e} [R_{t}-\eta^{\pi_e}|S_0=s, A_0=a]$) by incorporating the mediator in the conditioning set.

Next, we define $Q^{G_e}(s,a,m)$ as 
\begin{equation*}
    \sum_{t\ge 0}\Mean^{\pi_{e}}  [
    r(S_{t},\pi_0,M_t)-\eta^{G_e}|S_0=s, A_0=a, M_0=m],
\end{equation*} 
where $r(s,\pi_0,m)$ is a shorthand for $\sum_a \pi_0(a|s) r(s,a,m)$. 
$Q^{G_e}$ aggregates the difference between the expected reward of the interventional process $G_e^t$ starting from a given state-action-mediator triplet and that averaged over different initial conditions. 
It is crucial to note that $Q^{G_e}$ differs from $Q^{\pi}$ defined in \eqref{Q:pi_e}, in that the observed reward $R_t$ in \eqref{Q:pi_e} is replaced by the reward function $r(S_{t},\pi_0,M_{t})$. 
This is necessary as $G_e^t$ uses different policies for action selection and mediator generation at $t$. 

Following the same logic, we define $Q^{\pi_{e,0}}(s,a,m)$ as
\begin{eqnarray*}
    \sum_{t\ge 0}\Mean^{\pi_{e}} 
[r(S_t,\pi_0)-\eta^{\pi_{e,0}}|S_0=s, A_0=a, M_0=m],
\end{eqnarray*}
where $r(s,\pi_0)=\sum_{a,m}\pi_0(a|s)p_m(m|a,s)r(s,a,m)$. 
We similarly define $Q^{G_0}(s,a,m)$ as
\begin{eqnarray*}
\sum_{t\geq 0} \Mean^{G_0^t}[r(S_t,\pi_0)-\eta^{G_0}|S_0=s, A_0=a, M_0=m].
\end{eqnarray*}
We remark that all $Q$-functions are finite 
under the assumption of aperiodicity even though the horizon is infinite 
\citep{puterman2014markov}. This is because aperiodic Markov chains would reach their steady-state exponentially fast. 
As such, after a few iterations, the differences become very close to zero. 
More importantly, the $\eta$s and $Q$s are closely related according to the well-known Bellman equation, which is fundamental 
to deriving the DM estimator.  To elaborate, take the estimation of $\eta^{\pi_e}$ as an example. According to the Bellman equation, we have that
\begin{eqnarray}\label{bellman_0}
     \eta^{\pi_e} + Q^{\pi_e}(S_{t},A_{t}, M_{t}) = \Mean^{\pi_e}[R_{t}+
    \sum_{a,m}\pi_e(a|S_{t+1})\nonumber\\\times p_m(m|a,S_{t+1})Q^{\pi_e}(S_{t+1},a,m)|S_t,A_t,M_t].
\end{eqnarray}
Plugging in $\hat{p}_m$ learned from observed data into \eqref{bellman_0}, we can construct estimation equations to learn $Q^{\pi_e}$ and $\eta^{\pi_e}$ jointly. See Section \ref{nuisance} for details.
Let $\eta^\pi_d$ denote the resulting DM estimator for $\eta^\pi$. The DM estimator of each effect component is then constructed by plugging in these $\eta_d$s, the consistency of which requires correct model specifications of $r$, the $Q$-function and $p_m$. 

\subsection{Importance Sampling (IS) Estimators}\label{sec:ISest} As commented earlier, standard IS estimators suffer from the curse of horizon. In this section, we utilize the marginal importance sampling (MIS) method proposed in \citet{liu2018breaking} to break the curse of horizon. 
For a given policy $\pi$, we first introduce the MIS ratio, given by
$$\omega^{\pi}(s) = p^{\pi}(s)/p^{\pi_b}(s),$$
where $p^{\pi}$ and $p^{\pi_b}$ denote the stationary state distribution under $\pi$ and $\pi_b$, respectively. 
Using the change of measure theorem, it is immediate to see that, for $\pi \in \{\pi_e,\pi_0\}$,
\begin{equation}\label{MIS_standard}
    \frac{1}{NT}\sum_{i,t} \omega^{\pi}(S_{i,t})\frac{\pi(A_{i,t}|S_{i,t})}{\pi_b(A_{i,t}|S_{i,t})}R_{i,t}
\end{equation}
is unbiased to $\eta^{\pi}$. Similarly, using the change of measure theorem again, it is 
straightforward to show that
\begin{eqnarray}
    \frac{1}{NT}\sum_{i,t} \omega^{\pi_e}(S_{i,t})\frac{\pi_0(A_{i,t}|S_{i,t})}{\pi_b(A_{i,t}|S_{i,t})}R_{i,t},\label{MIS_eta3}\\
    \frac{1}{NT}\sum_{i,t} \omega^{G_0}(S_{i,t})\frac{\pi_0(A_{i,t}|S_{i,t})}{\pi_b(A_{i,t}|S_{i,t})}R_{i,t},\label{MIS_eta4}
\end{eqnarray} are unbiased to $\eta^{\pi_{e,0}}$ and $\eta^{G_0}$, respectively. Here, $\omega^{G_0}$ is a version of $\omega^{\pi}$ with the numerator equal to the stationary state distribution when the data are generated according to $\{G_0^t\}_t$. 
These two MIS estimators (\ref{MIS_eta3} and \ref{MIS_eta4}) differ from \eqref{MIS_standard} in that their state and action ratios are associated with two different interventional policies.



Lastly, we consider $\eta^{G_e}$. Recall that at time $t$, 
$G_e^t$ 
selects action according to $\pi_0$ and generates the mediator as if $\pi_e$ were applied to determine $A_t$. 
To further account for 
this distributional shift, we introduce a mediator ratio,
$\rho(S,A,M) = p_m^{-1}(M|S,A)[\sum_a \pi_e(a|S) p_m(M|S,a)]$, built upon which the following unbiased estimator, denoted as $\textrm{MIS}_1$, can be derived,
\begin{align*}
\frac{1}{NT}\sum_{i,t} \omega^{\pi_e}(S_{i,t}) \frac{\pi_0(A_{i,t}|S_{i,t})}{\pi_b(A_{i,t}|S_{i,t})} \rho(S_{i,t},A_{i,t},M_{i,t})R_{i,t}.
\end{align*} 
An alternative way to handle the distributional shift is to use the reward function instead of the observed reward to derive the IS estimator. 
This motivates the following estimator for $\eta^{G_e}$,  
\begin{align*}
(\textrm{MIS}_2): \frac{1}{NT}\sum_{i,t} \omega^{\pi_e}(S_{i,t}) \frac{\pi_e(A_{i,t}|S_{i,t})}{\pi_b(A_{i,t}|S_{i,t})} r(S_{i,t},\pi_0,M_{i,t}),
\end{align*}
which avoids the use of the mediator ratio. 

So far, we have discussed the MIS estimators for those $\eta$s. 
The subsequent MIS estimators for $\textrm{IDE}(\pi_e,\pi_0)$, $\textrm{IME}(\pi_e,\pi_0)$, $\textrm{DDE}(\pi_e,\pi_0)$, and $\textrm{DME}(\pi_e,\pi_0)$ can be similarly defined. 
Their consistencies require correct specifications of $\pi_b$, $p_m$, $r$, $\omega^{\pi_e}$, $\omega^{\pi_0}$, and $\omega^{G_0}$. 




\subsection{Multiply Robust (MR) Estimators}\label{TRE}
This section develops the MR estimators 
that combine the DM and MIS estimators for efficient and robust OPE. These estimators are derived based on the classical semiparametric theory \citep[see e.g.,][]{tsiatis2006semiparametric}.  
See Appendix \ref{appendix:EIF} for the detailed derivation. 
Let $O$ denote a tuple $(S,A,M,R,S')$. For each $\eta$, the proposed MR estimator is built upon the estimating function 
$\eta_d + I_{\eta}(O)$, 
where $\eta_d$ is the DM estimator of $\eta$ and $I_{\eta}(O)$ denotes some augmentation term that involves the MIS ratio. The purpose of introducing these augmentation terms lies in debiasing the bias of the DM estimator, making the resulting estimator more robust against model misspecification. Given the estimating function, its empirical average over the data tuples produces the final MR estimator. We present the detailed forms of these estimating functions below. 

First, consider $\eta^{\pi_e}$ and $\eta^{\pi_0}$. For a given policy $\pi \in \{\pi_e,\pi_0\}$, $I_{\eta^{\pi}}(O)$ is given by
\begin{multline*}
 \omega^{\pi}(S)\frac{\pi(A|S)}{\pi_b(A|S)}\Big[R+  Q^{\pi}(S',\pi)-Q^{\pi}(S,A)-\eta^{\pi}_d\Big],
\end{multline*} where $Q^{\pi}(s,\pi) = \sum_{a,m}\pi(a|s)p_m(m|a,s)Q^{\pi}(s,a,m)$ and $Q^{\pi}(s,a) = \sum_{m}p_m(m|a,s)Q^{\pi}(s,a,m)$.
Under the MMDP model, the term in brackets corresponds to a temporal difference error. Therefore, when $Q^{\pi}$, $\eta^{\pi}_d$, and $p_m$ are correctly specified, it is of mean zero given $(A,S)$. Thus, the resulting estimator is equivalent to DM which is consistent under correct model specification. On the contrary, when $\omega^{\pi}$ and $\pi_b$ are correctly specified, the final estimator is equivalent to MIS, which is consistent under these configurations \citep{liao2020batch}. As such, the resulting estimator is doubly robust whose consistency relying on the correct specification of $(Q^{\pi}, \eta^{\pi}, p_m)$ or $(\omega^{\pi}, \pi_b)$.  

Next, consider $\eta^{G_e}$. 
Let $I_{\eta^{G_e}}(O)$ denote 
\begin{eqnarray*}
    \omega^{\pi_e}(S)\Big[\frac{\pi_0(A|S)}{\pi_b(A|S)} \rho(S,A,M)\{R-r(S,A,M)\}+
   \frac{\pi_e(A|S)}{\pi_b(A|S)}\\\times\Big\{r(S,\pi_0,M)+Q^{G_e}(S',\pi_e)-
	Q^{G_e}(S,A)-\eta^{G_e}_d\Big\}\Big],
\end{eqnarray*} where $\rho$ is the mediator ratio defined before. Similarly, 
the second line is the temporal difference error 
with a zero mean given $(S,A,M)$ when models in $(r, p_m, Q^{G_e},\eta^{G_e}_d)$ are correctly specified. In addition, when $r$ is correctly specified, conditional on $(S,A,M)$, $\{R-r(S,A,M)\}$ is of zero mean as well. As such, $I_{\eta^{G_e}}(O)$ has a zero mean when $(r, p_m, Q^{G_e},\eta^{G_e})$ are correctly specified. Further, one can show that the final estimator based on $\eta_d^{G_e}+I_{\eta^{G_e}}(O)$ is unbiased to $\textrm{MIS}_1$ or $\textrm{MIS}_2$ introduced in Section \ref{sec:ISest} when $(p_m,\omega^{G_e},\pi_b)$ or $(r,\omega^{G_e},\pi_b)$ are correctly specified. As such, the estimator is triply robust in the sense that its consistency requires $(r, p_m, Q^{G_e},\eta^{G_e})$, $(p_m,\omega^{G_e},\pi_b)$ or $(r,\omega^{G_e},\pi_b)$ to be correct. 

Next, we consider $\eta^{\pi_{e,0}}$ and introduce $I_{\eta^{\pi_{e,0}}}(O)$, defined as
\begin{eqnarray*}
	\omega^{\pi_e}(S)\Big[\frac{\pi_0(A|S)}{\pi_b(A|S)} \{R-r(S,A)\}+
	\frac{\pi_e(A|S)}{\pi_b(A|S)}\\\times\Big\{r(S,\pi_0)
 +Q^{\pi_{e,0}}(S',\pi_e)
	-
	Q^{\pi_{e,0}}(S,A)-\eta^{\pi_{e,0}}_d\Big\}\Big],
\end{eqnarray*} where $r(s,a) = \sum_{m}p_m(m|a,s)r(s,a,m)$.
Following the same logic, we can show that the resulting estimator is doubly robust and requires either models in $(r, p_m, Q^{G_e},\eta^{G_e}_d)$ or those in $(\omega^{\pi_e}, \pi_b)$ are correctly specified. 

Finally, we consider $\eta^{G_0}$ and introduce $I_{\eta^{G_0}}(O)$ as follows,
\begin{multline*}
    \omega^{G_0}(S)\frac{\pi_0(A|S)}{\pi_{b}(A|S)}\Big[\Big\{R-r(S,A)\Big\}+\rho(S,A,M)\Big\{r(S,\pi_0)\\+ Q^{G_0}(S',G_0)-\eta^{G_0}_d-Q^{G_0}(S,A,M)\Big\}\Big]
    + \omega^{G_0}(S)\frac{\pi_e(A|S)}{\pi_{b}(A|S)}\\\times\Big[Q^{G_0}(S,\pi_0,M)-\sum_{a,m}\pi_0(a|S)p_m(m|A,S)Q^{G_0}(S,a,m)\Big],
\end{multline*} 
where $Q^{G_0}(s,\pi_0,m)$ is a shorthand of $\sum_a \pi_0(a|s)Q^{G_0}(s,a,m)$ and $Q^{G_0}(s,G_0)$ equals $\sum_{a,a',m}\pi_0(a|s)\pi_e(a'|s)p_m(m|a',s)Q^{G_0}(s,a,m)$. The resulting estimator's doubly robustness property can be similarly established. 

So far, we have introduced 
all the MR estimators for estimating these average rewards $\eta$s. 
We can plug in these estimators to construct the corresponding MR estimators for those dynamic treatment effects (i.e., $\textrm{MR-IDE}(\pi_e,\pi_0)$, $\textrm{MR-IME}(\pi_e,\pi_0)$, $\textrm{MR-DDE}(\pi_e,\pi_0)$, $\textrm{MR-DME}(\pi_e,\pi_0)$). Their consistencies and robustness can be similarly derived. We summarize and formally prove their robustness properties in Theorem \ref{thm:robust}.

%% file: Appendix/Learning_Nuisance_Function.tex
\subsection{Learning Nuisance Functions}\label{nuisance}
Recall that the MR estimators require estimation of nuisance functions including $\pi_b$, $r$, $p_m$, $\omega$, $Q$, and $\eta$. While $\pi_b$, $r$, and $p_m$ can be estimated efficiently using state-of-the-art nonparametric methods (i.e., regression/classification tree \citep{breiman2017classification}, random forest \citep{breiman2001random}, deep learning \citep{schmidt2020nonparametric}) with convergence rates faster than $N^{-\frac{1}{4}}$, we focus on the methods used to learn $\omega$, $Q$, and $\eta$.

We first consider the estimation of $\omega^{\pi}$ for any stationary policy $\pi$. Following the arguments in \citet{liu2018breaking} and \citet{uehara2020minimax}, we can show that for any function $f$
\begin{align*}
    \Mean \left[\omega^{\pi}(S)\{f(S)-\frac{\pi(A|S)}{\pi_b(A|S)}f(S')\}\right] = 0,
\end{align*} 
where the expectation is taken over the observed stationary distribution of $(S,A,S')$. Therefore, estimating $\omega^{\pi}$ is equivalent to solving a mini-max problem such that 
\begin{align}\label{minimax_omega}
\min_{\omega^{\pi}\in\Omega}\max_{f\in \mathcal{F}}\Mean^2 \left[\omega^{\pi}(S)\{f(S)-\frac{\pi(A|S)}{\pi_b(A|S)}f(S')\}\right]
\end{align} for some function classes $\Omega$ and $\mathcal{F}$. In our implementation, we consider linear function classes $\Omega$ and $\mathcal{F}$, which yields closed-form expressions. Specifically, let $\omega^{\pi}(s) = \xi^{T}(s)\beta$ for some $d_\omega$-dimensional $\beta \in \real^{d_\omega}$, where $\xi(s)$ is the feature vector generated by RBF sampler \citep{rahimi2007random}. Then (\ref{minimax_omega}) is equivalent to obtain $\beta$ by solving the equation
\begin{align*}
    \frac{1}{NT}\sum_{i,t}\left[\xi(S_i,t)-\frac{\pi(A_{i,t}|S_{i,t})}{\pi_b(A_{i,t}|S_{i,t})}\xi(S_{i,t+1})\right]\xi^{T}(S_{i,t})\beta = 0.
\end{align*} 
Similarly, considering $\omega^{G}$, we can show that 
\begin{align*}
    \Mean \left[\omega^{G}(S)\{f(S)-\rho(S,A,M)\frac{\pi_0(A|S)}{\pi_b(A|S)}f(S')\}\right] = 0,
\end{align*}where the expectation is taken over the distribution of $(S,A,M,S')$. $\omega^{G}$ can then be estimated following the same steps.

We next consider the estimation of pairs of $(Q, \eta)$. Taking $(Q^{\pi_e},\eta^{\pi_e})$ as an example, the estimation procedure is motivated by the Bellman equation model, such that:
\begin{equation}\label{Bellman1}
    Q^{\pi_e}(S_{t},A_{t}, M_{t}) = \Mean^{\pi_e}[R_{t}+
    \Mean^{\pi_e}_{a,m}Q^{\pi_e}(S_{t+1},a,m)
    -\eta^{\pi}].
\end{equation} Similar to the work of \citet{shi2020statistical}, we approximate the $Q$ function using linear sieves. Specifically, we assume that
\begin{align*}
    Q^{\pi_e}(s,a,m) \approx \Phi_{L}^{T}(s,m)\beta_{a}, \forall s \in \mathcal{S}, a \in \mathcal{A}, m \in \mathcal{M},
\end{align*} where $\Phi_{L}^{T}(s,m)$ is a $L$-dimensional feature vector derived using $L$ sieve basis functions, such as splines \citep{huang1998projection}. 
Let $\beta^* = (\beta_0^T,\cdots,\beta_{K-1}^T,\eta^{\pi})^{T}$. Let $\boldsymbol{U}(s,a,m)$ denotes
\begin{align*}
    [\Phi_L^T(s,m)1(a=0),\cdots,\Phi_L^T(s,m)1(a=K-1),1]^T,
\end{align*} and $\boldsymbol{V}(s)$ denotes
\begin{multline*}
[\Mean_{m|s,a=0}\Phi_L^T(s,m)\pi_e(0|s),\cdots,\\
\Mean_{m|s,a=K-1}\Phi_L^T(s,m)\pi_e(K-1|s),0]^T,
\end{multline*} where $\Mean_{m|s,a}\Phi_{L}^{T}(s,m) = \int_{m}\Phi_{L}^{T}(s,m)p(m|s,a)$ can be approximated by Monte Carlo sampling in practice.
Then, the equation (\ref{Bellman1}) can be rewritten as 
\begin{align*}
    \Mean U(S,A,M)[R+V(S')^{T}\beta^*-U(S,A,M)^{T}\beta^*] = 0.
\end{align*}Let $U_{i,t} = U(S_{i,t},A_{i,t}, M_{i,t})$ and $V_{i,t} = V(S_{i,t})$. Based on the observational data, the closed-form solution of $\beta^*$ is
\begin{align*}
    \left[\frac{1}{NT}\sum_{i,t}U_{i,t}(U_{i,t}-V_{i,t+1})^{T}\right]^{-1}\frac{1}{NT}\sum_{i,t}U_{i,t}R_{i,t}.
\end{align*}
In practice, we add ridge penalty to the term within the bracket to prevent overfitting, and let $L$ grow with the sample size 
to improve the approximation precision.

%% file: 5_Theory_robust.tex
\section{Statistical Guarantees}
In this section, we prove the robustness and semi-parametric efficiency of the proposed MR estimator. We begin with some notations. Let $\mathcal{Q}^{(\cdot)}, \Omega^{(\cdot)}, \mathcal{H}_m, \mathcal{H}_r$, and $\Pi_b$ respectively denote the function class of $Q^{(\cdot)}, \omega^{(\cdot)}, p_m, r$, and $\pi_b$. 

\begin{theorem}
\label{thm:robust}
\textbf{Multiply Robustness.}
Suppose the conditions in Theorem \ref{thm:identification} holds, the process $\{S_{i,t}\}_{t\ge 0}$ is stationary, $\pi_{b}$, $\hat{\pi}_{b}$, $p_{m}$ and $\hat{p}_{m}$ are uniformly bounded away from $0$, and $\mathcal{Q}^{(\cdot)}, \Omega^{(\cdot)}, \mathcal{H}_m, \mathcal{H}_r$, and $\Pi_b$ are bounded VC-type classes \citep{chernozhukov2014gaussian} with VC indices upper bounded by $O(N^k)$ for some $k<1/2$. As $NT\to \infty$,
\vspace{-.1cm}
\begin{enumerate}
    \vspace{-.1cm}
    \item $\textrm{MR-IDE}(\pi_e,\pi_0)$ is consistent if  either the set of models in ($\omega^{\pi_e}$, $\pi_b$, $r$) or in ($\omega^{\pi_e}$, $\pi_b$, $p_m$) or in ($Q^{\pi_e}$, $Q^{G_e}$, $\eta^{\pi_e}_d$, $\eta^{G_e}_d$, $r$, $p_m$) are consistently estimated;
    \vspace{-.1cm}
    \item $\textrm{MR-IME}(\pi_e,\pi_0)$ is consistent if  either the set of models in ($\omega^{\pi_e}$, $\pi_b$, $r$) or in ($\omega^{\pi_e}$, $\pi_b$, $p_m$) or in ($Q^{G_e}$, $Q^{\pi_{e,0}}$, $\eta^{G_e}_d$, $\eta^{\pi_{e,0}}_d$, $r$, $p_m$) are consistently estimated;
    \vspace{-.1cm}
    \item $\textrm{MR-DDE}(\pi_e,\pi_0)$ is consistent if  either the set of models in ($\omega^{\pi_e}$, $\omega^{G_0}$, $\pi_b$, $p_m$) or in ($Q^{\pi_{e,0}}$, $Q^{G_0}$, $\eta^{\pi_{e,0}}_d$, $\eta^{G_0}_d$, $r$, $p_m$) are consistently estimated;
    \vspace{-.1cm}
    \item $\textrm{MR-DME}(\pi_e,\pi_0)$ is consistent if  either the set of models in ($\omega^{\pi_0}$, $\omega^{G_0}$, $\pi_b$, $p_m$) or in ($Q^{G_0}$, $Q^{\pi_0}$, $\eta^{G_0}_d$, $\eta^{\pi_0}_d$, $r$, $p_m$) are consistently estimated.
    \vspace{-.1cm}
\end{enumerate}
\end{theorem}



Theorem \ref{thm:robust} formally establish the triply robustness properties of $\textrm{MR-IDE}(\pi_e,\pi_0)$ and $\textrm{MR-IME}(\pi_e,\pi_0)$, as well as the doubly robustness properties of $\textrm{MR-DDE}(\pi_e,\pi_0)$ and $\textrm{MR-DME}(\pi_e,\pi_0)$, respectively. To save space, the proof of this theorem is deferred to the Appendix \ref{appendix:robust}. 

\begin{theorem}
\label{thm:efficiency}
\textbf{Efficiency.} Suppose the conditions in Theorem \ref{thm:robust} holds, and $\hat{Q}^{(\cdot)}, \hat{\omega}^{(\cdot)}, \hat{p}_m, \hat{r}, \hat{\pi}_b$, and $\hat{\eta}^{(\cdot)}_d$ converges to their oracle value in $L_2$ norm at a rate of $N^{-k^*}$ for some $k^*>1/4$, respectively. The MR estimators are asymptotically normal with an asymptotic variance achieving the semiparametric efficiency bound. 
\end{theorem}
To save the space, the proof of this theorem is differed to the Appendix \ref{appendix:efficiency} with a sketch of the proof at the beginning. 
A Wald-type Confidence Interval (CI) for each MR estimator can be derived from Theorem \ref{thm:efficiency}. 

%% file: 6_Toy_Example.tex
\section{Numerical Examples}
In this section, we evaluate the estimation performance of the proposed methods through three simulation studies. Specifically, we demonstrate the robustness of the proposed MR estimator to model misspecification in the first simulation. In the second simulation, we compare the DM, MIS, and MR estimators to the classic direct/indirect estimator \citep{pearl2022direct} to demonstrate the importance of longitudinal mediation analysis, considering the policy effect on state transition. The final simulation is a semi-synthetic study that simulates the generation process of real data and demonstrates the superiority of the proposed MR estimators. For any effect $X$, let $\hat{X}$ be an estimator. We define the logbias as $\log|\Mean(\hat{X}-X)|$
and logMSE as $\Mean[\log(\hat{X}-X)^2]$. 

\subsection{Toy Example I}\label{toy_1}
\begin{figure}[ht]
    \centering
    \vspace{-.3cm}
    \includegraphics[width=\columnwidth]{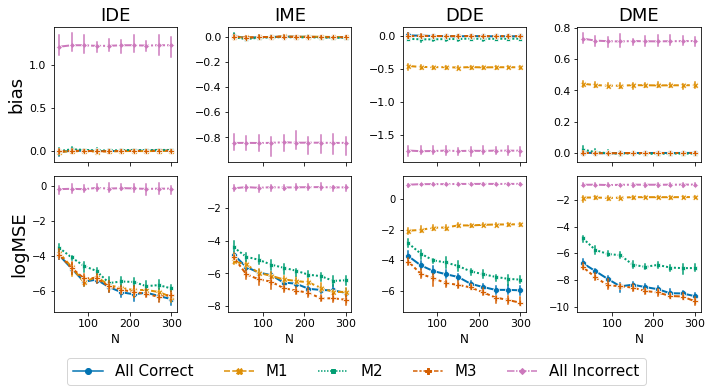}
    \vspace{-.5cm}
    \caption{Bias and the logMSE of MR estimators, aggregated over 200 random seeds. The error bars represent the 95\% CI.}
    \label{fig:robust}
    \vspace{-.3cm}
\end{figure}

We consider a simplified MMDP setting with binary states, actions, mediators, and rewards. See Appendix \ref{toy settings} for specific data generation settings. Let $\mathbb{M}_{1} = (\omega^{\pi_e}, \pi_b, r)$, $\mathbb{M}_{2} = (\omega^{\pi_e}, \omega^{\pi_0}, \omega^{G_0}, \pi_b, p_m)$, $\mathbb{M}_{3} = ( \{Q^{\pi}, \eta_d^{\pi}\}_{\pi \in \{\pi_e, G_e, \pi_{e,0}, G_0, \pi_0\}}, r, p_m)$. To investigate the robustness of the MR estimator, we test its performance in four scenarios:  i) $\mathbb{M}_{1}$, $\mathbb{M}_{2}$, and $\mathbb{M}_{3}$ are all correctly specified; ii) only $\mathbb{M}_{1}$ is correctly specified; iii) only $\mathbb{M}_{2}$ is correctly specified; iv) only $\mathbb{M}_{3}$ is correctly specified; and v) all the models in  $\mathbb{M}_{1}$, $\mathbb{M}_{2}$, and $\mathbb{M}_{3}$ are incorrectly specified by injecting non-negligible
random noises. As shown in Figure~\ref{fig:robust}, $\textrm{MR-IDE}(\pi_e,\pi_0)$ and $\textrm{MR-IME}(\pi_e,\pi_0)$ are consistent when either $\mathbb{M}_{1}$, $\mathbb{M}_{2}$, or $\mathbb{M}_{3}$ is correctly specified, and $\textrm{MR-DDE}(\pi_e,\pi_0)$ and $\textrm{MR-DME}(\pi_e,\pi_0)$ are consistent when either $\mathbb{M}_{2}$ or $\mathbb{M}_{3}$ is correctly specified.

\subsection{Toy Example II}\label{toy_2}
\begin{figure}[ht]
    \centering
    \vspace{-.2cm}
    \includegraphics[width=\columnwidth]{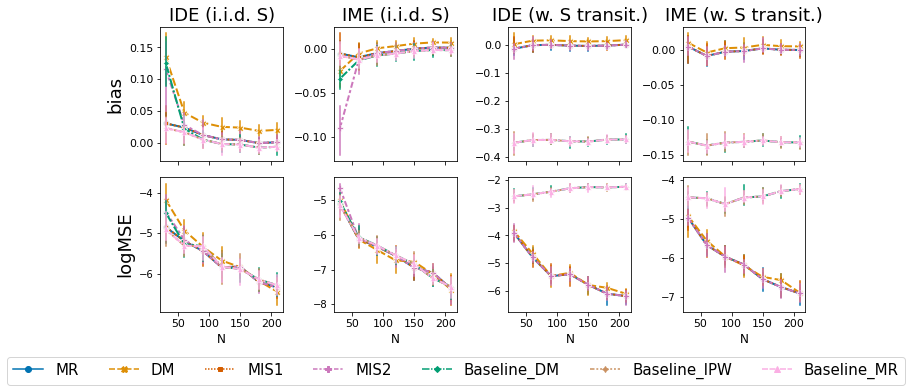}
    \vspace{-.5cm}
    \caption{Bias and the logMSE of estimators, under different data generation scenarios. The results are aggregated over 200 random seeds. The error bars represent the 95\% CI. Nuisance functions are estimated as discussed in Section \ref{nuisance}.}
    \label{fig:baseline}
    \vspace{-.5cm}
\end{figure}

As discussed in Section \ref{sec:related}, most existing works focus on a two-way decomposition of immediate treatment effects under the setting with a single stage. In this section, we compare the proposed estimators of IDE and IME to three baseline estimators assuming i.i.d. samples (See Appendix \ref{baseline} for details).
We first repeat the data generation process from Section \ref{toy_1}, in which the states are affected by the history observations for each trajectory. Then, by modifying the distribution of the next state, $S_{t+1}$, as $\prob(S_{t+1} = 1) = .2$, we consider a second scenario in which all observations of states are i.i.d sampled. Note that there are two versions of MIS estimators for IDE and IME. Let $\textrm{MIS2}$ denote the MIS estimators using the $\textrm{MIS}_2$ to estimate $\eta^{G_e}$. According to Figure~\ref{fig:baseline}, when states are i.i.d. sampled, all estimators produce consistent estimates. However, when policy-induced state transitions occur, all baseline estimators yield biased estimates, whereas the proposed estimators continue to provide consistent estimates, implying the necessity of accounting for the policy effect on the state transition.

\subsection{Semi-Synthetic Data}\label{semi_syn}
\begin{figure}[ht]
    \centering
    \vspace{-.3cm}
    \includegraphics[width=\columnwidth]{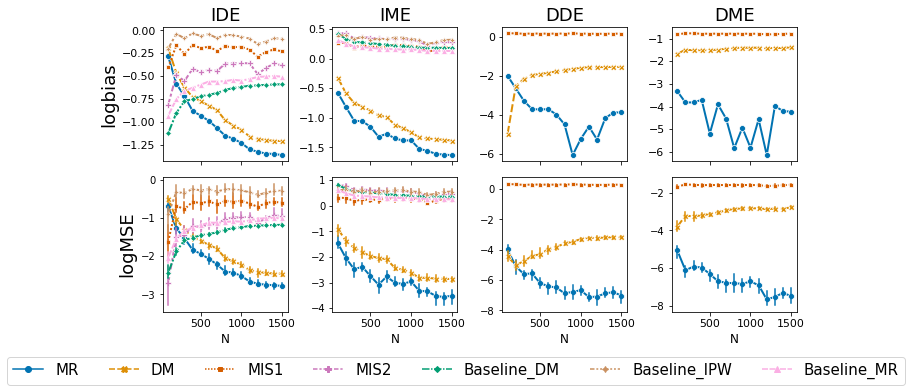}
    \vspace{-.5cm}
    \caption{The logbias and logMSE of various estimators, aggregated over 100 random seeds. The error bars represent the 95\% CI. Fix $T=25$.}
    \label{fig:T25}
    \vspace{-.3cm}
\end{figure}

\begin{figure}[ht]
    \centering
    \vspace{-.25cm}
    \includegraphics[width=\columnwidth]{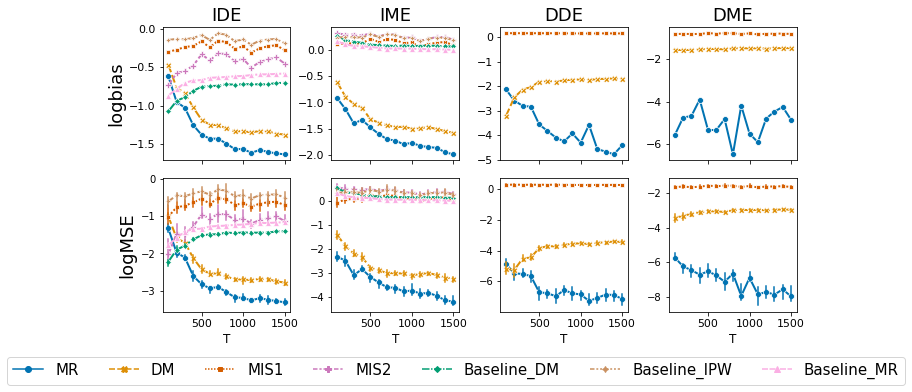}
    \vspace{-.5cm}
    \caption{The logbias and logMSE of various estimators, aggregated over 100 random seeds. The error bars represent the 95\% CI. Fix $N=50$.}
    \label{fig:N50}
    \vspace{-.3cm}
\end{figure}

In this section, we evaluate empirical performance of estimators using a semi-synthetic dataset structured similarly to the real dataset analyzed in Section \ref{real}. Specifically, we consider an MMDP setting with continuous reward, state, and mediator spaces and a binary action space.  See Appendix \ref{synthetic settings} for more information on the data-generation process. We compared the MR estimators to the DM estimators, the MIS estimators, and three baseline estimators. As shown in Figure~\ref{fig:T25} and Figure~\ref{fig:N50}, the MR estimators outperform all other estimators for all components of ATE, especially when the sample size is large. We first focus on $\textrm{IDE}(\pi_e,\pi_0)$ and $\textrm{IME}(\pi_e,\pi_0)$. On the one hand, the baseline and MIS estimators are all biased, whereas the bias and MSE of the proposed DM and MR estimators decay continuously as $N$ or $T$ increases. On the other hand, the DM estimators yield relatively more significant bias and MSE than MR estimators. Considering the $\textrm{DDE}(\pi_e,\pi_0)$ and $\textrm{DME}(\pi_e,\pi_0)$, both the DM and MIS estimators are biased with non-decreasing MSE, whereas the MR estimators continue to provide estimates with low bias and low MSE that decrease with $N$ and $T$. The results are in line with our theoretical findings. To further support the superior performance of the proposed MR estimators, additional simulation studies are conducted in Appendix \ref{additional_Exp} under different settings of data-generating mechanisms, all of which reach the same conclusion as in this section. 






%% file: 7_Real.tex
\section{Real Data Application}\label{real}
In this section, we apply the proposed MR estimators to analyze the real dataset from the IHS \citep{necamp2020assessing}, which was discussed as a motivating example in Section \ref{intro}.
The study involved 1565 interns and lasted six months. 
Every day, the participant would either receive a notification ($A_t = 1$) or no notification ($A_t = 0$). Meanwhile, participants' mood score ($R_t$), step count ($M_{t,1}$), and hours of sleep ($M_{t,2}$) were recorded. At each time step, we consider the previous time step's mood score as the current state (i.e., $S_t = R _{t-1}$).

Using the control policy $\pi_0$ of no intervention, we are interested in evaluating the treatment effects of the behavior policy $\pi_b$ used throughout the study, which sends notifications to individuals randomly with a constant probability of $.75$. According to \citet{necamp2020assessing}, pushing notifications has a negative impact on the mood condition when participants are already in a good mood (i.e., $S_t > 6$). Given that the majority of observations in the data have $S_t>6$, the ATE of $\pi_b$ is expected to be negative. As summarized in Table \ref{real_result}, the ATE of $\pi_b$ is significantly negative with an effect size of .1, which is consistent with our expectations. Further investigation of the ATE composition reveals that the immediate effects are all negligible. In contrast, the DDE and DME are both significant and account for the majority of the treatment effect, indicating the importance of learning the delayed effects and mediator effects to understand the entire mechanism from actions to outcomes.



Furthermore, given that the delayed effects are all passing through $S_t$, rather than simply abandoning the treatment proposal, it is recommended that we consider a state-dependent policy to make more informed decisions based on the $S_t$ and hence to improve the overall treatment effect. To support this claim, we further evaluate an optimal state-dependent policy, $\hat{\pi}_{opt}$, which is estimated by using single-stage policy estimation based on the observed data (See Appendix \ref{opt_policy} for more information). According to Table \ref{real_result}, in contrast to $\pi_b$, the estimated ATE of $\hat{\pi}_{opt}$ is $.090$, with significantly positive direct effects. This further demonstrates the necessity of analyzing dynamic treatment policies as opposed to fixed action sequences, which have been the main focus of most existing literature on mediation analysis.

\begin{table}[ht]
\centering
\resizebox{\columnwidth}{!}{%
\begin{tabular}{c c c c c c}
\hline
        $\pi_e$ & $\textrm{IDE}$ & $\textrm{IME}$ & $\textrm{DDE}$ & $\textrm{DME}$ & \textrm{ATE}\\ \hline
$\pi_b$  & -.007(.007)  & -.000(.001) & -.085(.034) & -.008(.004) &   -.100 (.041)    \\ \hline
$\hat{\pi}_{opt}$ & .018(.006)  & -.001(.001) & .077(.030) & -.005(.005)  &   .090 (.037)   \\ \hline
\end{tabular}%
}
\vspace{-.3cm}
\caption{Estimated treatments effects (standard error) for $\pi_b$ and $\hat{\pi}_{opt}$, compared to $\pi_0$ with no intervention.}
\label{real_result}
\vspace{-.2cm}
\end{table}

%% file: 8_Discussion.tex
\section{Conclusion}
Motivated by the growing number of applications (e.g., mobile health) with sequential decision-making over an infinite number of decision points, we propose an MMDP framework and a four-way decomposition of ATE of random policies to analyze the dynamic mediation effects. For each effect component, multiply-robust estimators with theoretical and numerical support are provided. 
The proposed framework can be extended in several aspects. First, the proposed methods are limited to applications with discrete action space. Meanwhile, problems such as dynamic pricing and personalized dose finding typically involve a continuous action space, which is worth studying in future work. Second, the no unmeasured confounder assumption can be violated from data collected from observational studies. Therefore, a confounded MMDP is worth investigating.



%% file: Appendix/Graphical_PO.tex
\section{More Details about Effect Decomposition} \label{graphical PO}
\subsection{Effect Decomposition in the Framework of Potential Outcomes}
Let $\bar{a}_t = (a_0, \cdots, a_t)$ denote a fixed treatment sequence up to time $t$. Let $M^*_t(\bar{a}_t)$ denote the potential mediator that would be observed at $t$ if $\bar{a}_t$ were taken, and $\bar{M}^*_t(\bar{a}_t) = (M^*_0(\bar{a}_0), \cdots, M^*_t(\bar{a}_t))$. Replacing the fixed action sequence by any random policy $\pi$, $M^*_t(\pi)$ denotes the potential mediator if the actions were taken under $\pi$. 

We first focus on the effects of action and mediator on their proximal outcome. Denotes $\pi_{e,0}^{t}$ a policy where the first $t-1$ steps follow $\pi_e$ and then follow $\pi_0$ at $t$. For $X \in \{S, R\}$, $X^*_t(\pi_1,\bar{M}^*_t(\pi_2))$ denotes the potential covariate if $\pi_1$ were used to determine actions and the mediators were set to levels as if $\pi_2$ were used. $\textrm{IDE}_t$ and $\textrm{IME}_t$ are defined as
\begin{align*}
    \textrm{IDE}_t(\pi_e,\pi_0) &= \Mean\left[R^{*}_t(\pi_e,\bar{M}^*_{t}(\pi_e))-R^{*}_t(\pi_{e,0}^{t},\bar{M}^*_{t}(\pi_e))\right],\\
    \textrm{IME}_t(\pi_e,\pi_0) &=\Mean[R^{*}_t(\pi_{e,0}^{t},\bar{M}^*_{t}(\pi_e))-R^{*}_t(\pi_{e,0}^{t},\bar{M}^*_{t}(\pi_{e,0}^{t}))]. 
\end{align*} Given that both $\bar{A}_{t-1}$ and $\bar{M}_{t-1}$ were set to levels as if $\pi_e$ were used, $\textrm{IDE}_t(\pi_e,\pi_0)$ contrasts the impact of $A_t$ generated by $\pi_e$ and $\pi_0$ on the proximal outcome $R_t$, fixing $M_t$ to $M^*_t(\pi_e)$. $\textrm{IME}_t(\pi_e,\pi_0)$ compares the effect of $M_t$ at levels $M^*_t(\pi_e)$ and $M^*_t(\pi_{e,0}^t)$ on $R_t$, when $A_t$ is set by $\pi_0$.

Next, we focus on the delayed effects of the historical action sequence $\bar{A}_{t-1}$ and mediator sequence $\bar{M}_{t-1}$ on $R_t$. Within the MMDP framework, $\bar{A}_{t-1}$ and $\bar{M}_{t-1}$ affect $R_t$ through $S_t$. Noticing that $\Mean [R^{*}_t(\pi_{0},\bar{M}^*_{t}(\pi_{e,0}^{t}))]$ is unidentifiable due to the presence of intermediate confounders $\bar{S}_t$ \citep{tchetgen2014identification}, we adopt the RI-based approach proposed in \citet{zheng2017longitudinal}. 

We first define the \textit{conditional} probability density of mediator at $t$, 
\begin{equation*}
    G_t^{\bar{a}'_t}(\cdot|\bar{m}_{t-1}, \bar{r}_{t-1}, \bar{s}_t) = p_{M^*_t(\bar{a}'_t)|\bar{M}^*_{t-1}(\bar{a}'_{t-1}),\bar{R}^*_{t-1}(\bar{a}'_{t-1}, \bar{M}^*_{t-1}(\bar{a}'_{t-1})),\bar{S}^*_t(\bar{a}'_{t-1}, \bar{M}^*_{t}(\bar{a}'_{t-1}))}(\cdot|\bar{m}_{t-1},\bar{r}_{t-1}, \bar{s}_t),
\end{equation*} if $\bar{a}'_t$ is assigned.
At time $t$, given the historical trajectories $\bar{m}_{t-1}$, $\bar{r}_{t-1}$, and $\bar{s}_t$, we intervene in the mediator by randomly drawing $M_t \sim G_t^{\bar{a}'_t}(\cdot|\bar{m}_{t-1}, \bar{r}_{t-1}, \bar{s}_t)$. For brevity, we omit the conditionality and let $\bar{G}_t^{\bar{a}'_t} = (G_0^{\bar{a}'_0},\cdots, G_t^{\bar{a}'_t})$ denote the process by which the mediator is set to a conditional random draw at each time $t$. Using a two-stage interventional process as an illustration, we set $\bar{A}_1 = \bar{a}_1$ and $\bar{M}_1 \sim \bar{G}_1^{\bar{a}'_1}$. 
%
The generating process of $R^{*}_1(\bar{a}_{1},\bar{G}_{1}^{\bar{a}'_{1}})$ is as follows:
After observing an initial state $s_0$, we would first assign a treatment $a_0$ and set $M_0$ by randomly drawing $m_0
\sim G_0^{a'_0}(\cdot|s_0)$, and then measure the resulting $R^{*}_0(a_0,\bar{G}_{0}^{a'_{0}}) = r_0$ and $S^{*}_1(a_0,\bar{G}_{0}^{a'_{0}}) = s_1$. At $t=1$, we then take action $a_1$ and set $M_1$ by randomly drawing $m_1\sim G_1^{\bar{a}'_1}(\cdot|s_0, s_1, m_0, r_0)$, and finally observe $R^{*}_1(\bar{a}_{1},\bar{G}_{1}^{\bar{a}'_{1}})$ as the outcome. 
Analogously, $R^{*}_t(\pi_1,\bar{G}_{t}^{\pi_2})$ is the potential reward if $\pi_1$ were used to determine $\bar{A}_t$ and $\bar{M}_t$ were set to have the $\pi_2$-driven conditional distributions $\bar{G}_t^{\pi_2}$. We then define the delayed effects as
\begin{align*}
    \textrm{DDE}_t(\pi_e,\pi_0) &= \Mean[R^{*}_t(\pi_{e,0}^{t},\bar{M}^*_{t}(\pi_{e,0}^{t}))-R^{*}_t(\pi_{0},\bar{G}_{t}^{\pi_{e,0}^{t}})],\\
    \textrm{DME}_t(\pi_e,\pi_0) &= \Mean[R^{*}_t(\pi_{0},\bar{G}_{t}^{\pi_{e,0}^{t}})-R^{*}_t(\pi_{0},\bar{M}^*_{t}(\pi_{0}))].
\end{align*}

Setting $A_t$ and $M_t$ to levels as if policy $\pi_0$ were used at $t$, $\textrm{DDE}_t(\pi_e,\pi_0)$ compares the effects of $\bar{A}_{t-1}$ generated by $\pi_e$ and $\pi_0$ on $R_t$ when $\bar{M}_{t-1}$ is generated by $\pi_e$, while $\textrm{DME}_t(\pi_e,\pi_0)$ contrasts the effects of $\bar{M}_{t-1}$ generated by $\pi_e$ and $\pi_0$ on $R_t$ when $\bar{A}_{t-1}$ is set by $\pi_0$. See Appendix \ref{nonidentifiable} for more discussion about the non-identifiability issue and Appendix \ref{PO graphs} for graphical representations of each component.

\begin{remark} 
As suggested in \citet{robins1992identifiability}, there are two ways to decompose the total effect.
The above definitions of direct effects and mediator effects are analogous to the Total Direct Effect (TDE) and the Pure Indirect Effect (PIE) \citep{robins1992identifiability}, while an alternative decomposition is provided in Appendix \ref{Alter Decomp}. 
By replacing $\pi_e$ and $\pi_0$ with $\bar{a}'_t$ and $\bar{a}_t$, IDE and IME are equivalent to TDE and PIE. Let $\tilde{\bar{a}}_t = \{\bar{a}'_{t-1},a_t\}$, we further replace $\pi_{e,0}^t$ with $\tilde{\bar{a}}_t$ to define DDE and DME. When $t>0$, if we set $\tilde{\bar{a}}_t = \bar{a}'_t$, DDE and DME are analogous to the effect components defined in \citet{zheng2017longitudinal}.

\end{remark}

\subsection{Graphical Representation of Potential Outcomes} \label{PO graphs}
\begin{table}[ht]
\centering
\begin{tabular}{c c c}
\hline
        $R^{*}_t(\pi_e,\bar{M}^{*}_{t}(\pi_e))$ & $R^{*}_t(\pi_{e,0}^{t},\bar{M}^{*}_{t}(\pi_{e}))$ & $R^{*}_t(\pi_{e,0}^{t},\bar{M}^{*}_{t}(\pi_{e,0}^{t}))$ \\ \hline
\includegraphics[scale=.4]{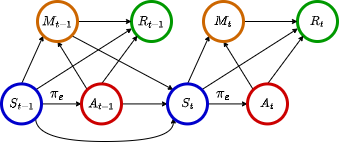}  & \includegraphics[scale=.4]{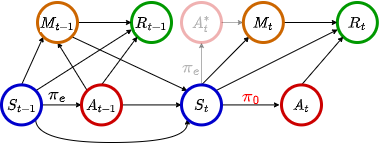} & \includegraphics[scale=.4]{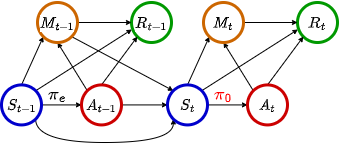}    \\ \hline
\end{tabular}
\caption{Potential Outcomes Related to Immediate Effects. }
\label{PO1}
\end{table}

\begin{table}[ht]
\centering
\begin{tabular}{|c|c|}
\hline
        $R^{*}_t(\pi_{e,0}^{t},\bar{M}^{*}_{t}(\pi_{e,0}^{t}))$ & \includegraphics[scale=.4]{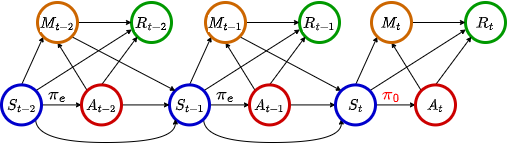}\\\hline
        $R^{*}_t(\pi_{0},\bar{G}_{t}^{\pi_{e,0}^{t}})$ & \includegraphics[scale=.4]{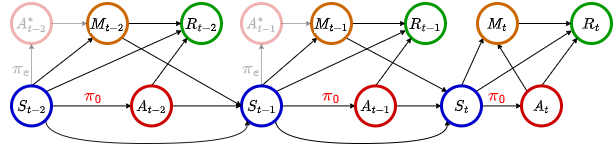}\\\hline
        $R^{*}_t(\pi_{0},\bar{M}^{*}_{t}(\pi_{0}))$ & \includegraphics[scale=.4]{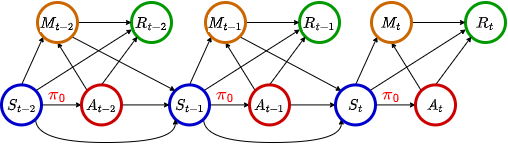}\\ \hline
\end{tabular}
\caption{Potential Outcomes Related to Delayed Effects. }
\label{PO2}
\end{table}

In Table \ref{PO1} and Table \ref{PO2}, using causal graphs, we explicitly depict the process generating the potential reward terms involved in the effect decomposition. Specifically, $R^{*}_t(\pi_e,\bar{M}^{*}_{t}(\pi_e))$ is the potential reward that would be observed if $\pi_e$ were used to determine $\bar{A}_t$ and $\bar{M}_t$; $R^{*}_t(\pi_{e,0}^{t},\bar{M}^{*}_{t}(\pi_{e}))$ is the potential reward that would be observed if $\pi_e$ were used to determine the historical sequences $\bar{A}_{t-1}$ and $\bar{M}_{t-1}$, while $A_t$ were determined by $\pi_0$ and $M_t$ were set to $M^*_t(\pi_e)$; $R^{*}_t(\pi_{e,0}^{t},\bar{M}^{*}_{t}(\pi_{e,0}^{t}))$ is the potential reward if $\pi_e$ were used to determine $\bar{A}_{t-1}$ and $\bar{M}_{t-1}$, while $A_t$ and $M_t$ are generated by $\pi_0$;  $R^{*}_t(\pi_{0},\bar{G}_{t}^{\pi_{e,0}^{t}})$ is the potential reward if $\pi_0$ were used to determine $A_{t}$ and $M_{t}$, while the historical sequences $\bar{A}_{t-1}$ and $\bar{M}_{t-1}$ were determined by $\pi_0$ and $\pi_e$ respectively; and $R^{*}_t(\pi_{0},\bar{M}^{*}_{t}(\pi_{0}))$ is the potential reward that would be observed if $\pi_0$ were used to determine $\bar{A}_t$ and $\bar{M}_t$.

By definition, $\textrm{IDE}(\pi_e,\pi_0)$ is the contrast between causal structures of $R^{*}_t(\pi_e,\bar{M}^{*}_{t}(\pi_e))$ and $R^{*}_t(\pi_{e,0}^{t},\bar{M}^{*}_{t}(\pi_{e}))$; $\textrm{IME}(\pi_e,\pi_0)$ is the contrast between causal structures of $R^{*}_t(\pi_{e,0}^{t},\bar{M}^{*}_{t}(\pi_{e}))$ and $R^{*}_t(\pi_{e,0}^{t},\bar{M}^{*}_{t}(\pi_{e,0}^{t}))$; $\textrm{DDE}(\pi_e,\pi_0)$ is the contrast between causal structures of $R^{*}_t(\pi_{e,0}^{t},\bar{M}^{*}_{t}(\pi_{e,0}^{t}))$ and $R^{*}_t(\pi_{0},\bar{G}_{t}^{\pi_{e,0}^{t}})$; and $\textrm{DME}(\pi_e,\pi_0)$ is the contrast between causal structures of $R^{*}_t(\pi_{0},\bar{G}_{t}^{\pi_{e,0}^{t}})$ and $R^{*}_t(\pi_{0},\bar{M}^{*}_{t}(\pi_{0}))$.

\subsection{Non-identifiability Issue} \label{nonidentifiable}
To understand the non-identifiability issue, let us focus on the identification of $\Mean [R^{*}_t(\pi_{0},\bar{G}_{t}^{\pi_{e,0}^{t}})]$. For simplicity, we consider two fixed action sequences $\bar{a}_t$ and $\bar{a}^*_{t}$ and let the mediator and state be discrete values. 
 Based on the definition, let $\tilde{\bar{a}}_t = (\bar{a}^*_{t-1},a_t)$ and $t=1$, we have that 
\begin{multline*}
    \Mean [R^{*}_1(\bar{a}_1,\bar{M}^*_{1}(\tilde{\bar{a}}_{1}))] = \sum_{a_0,a_1,a_0^*,m_0,m_1,s_0,s_1,s_1^*}\Mean(R_1^*(\bar{a}_1,\bar{m}_1)|a_0,a_1,m_0,m_1,s_0,s_1)\prob(M^*_1(\tilde{\bar{a}}_1) = m_1|a_0^*,a_1,m_0,s_0,s_1^*)\\ \times \prob(S_1^*(a_0,m_0)=s_1,S_1^*(a_0^*,m_0)=s_1^*|m_0,s_0)\prob(M^*_0(a_0^*) = m_0|a_0^*,s_0)\prob(S_0=s_0).
\end{multline*} While $\Mean(R_1^*(\bar{a}_1,\bar{m}_1)|a_0,a_1,m_0,m_1,s_0,s_1)$, $\prob(M^*_1(\tilde{\bar{a}}_1) = m_1|a_0^*,a_1,m_0,s_0,s_1^*)$, $\prob(M^*_0(a_0^*) = m_0|a_0^*,s_0)$, and $\prob(S_0=s_0)$ are identifiable from the observational data, the joint distribution of $\prob(S_1^*(a_0,m_0)=s_1,S_1^*(a_0^*,m_0)=s_1^*|m_0,s_0)$ is not identified, leading to the non-identifiability of $\Mean [R^{*}_1(\bar{a}_1,\bar{M}^*_{1}(\tilde{\bar{a}}_{1}))]$. The non-identifiability of $\Mean [R^{*}_t(\pi_{0},\bar{G}_{t}^{\pi_{e,0}^{t}})]$ is then followed.

%% file: Appendix/Alternative_Decomp.tex
\section{Alternative Decomposition of $\textrm{ATE}(\pi_e,\pi_0)$} \label{Alter Decomp}
In this section, we provide an alternative decomposition of $\textrm{ATE}(\pi_e,\pi_0)$. Let $\tilde{G}$ denote the stochastic process selecting actions according to $\pi_e$ and drawing mediators assuming $\pi_0$ was applied. Adopting the notations used in the main text, we have that 
\begin{align*}
    \textrm{ATE}(\pi_e,\pi_0) = \underbrace{\eta^{\pi_e} - \eta^{\tilde{G}_e}}_{\textrm{DME}^{(2)}(\pi_e,\pi_0)} + \underbrace{\eta^{\tilde{G}_e}-\eta^{\pi_{0,e}}}_{\textrm{DDE}^{(2)}(\pi_e,\pi_0)}+\underbrace{\eta^{\pi_{0,e}} - \eta^{\tilde{G}_0}}_{\textrm{IME}^{(2)}(\pi_e,\pi_0)} + \underbrace{\eta^{\tilde{G}_0}-\eta^{\pi_{0}}}_{\textrm{IDE}^{(2)}(\pi_e,\pi_0)}.
\end{align*}
In the following subsections, we further written the alternative decomposition in the framework of potential outcomes along with the corresponding MR estimators. 
\subsection{Decomposition in the Framework of Potential Outcomes}
We follow the notations used in the Appendix \ref{graphical PO}. Another classic decomposition of the total effect is well-known as natural effect decomposition, which divides the total effect into Natural Direct Effect (NDE) (also named as Pure Direct Effect) and Natural Indirect Effect (NIE) (also named as Total Indirect Effect) \citep{robins1992identifiability,pearl2022direct, vanderweele2013three}. Denotes $\pi_{0,e}^{t}$ a policy where the first $t-1$ steps follow $\pi_0$ and then follow $\pi_e$ at $t$. Following the natural effect decomposition , we alternatively decompose the $\textrm{TE}_t(\pi_e,\pi_0)$ as follows:
\begin{align*}
    \textrm{TE}_t(\pi_e,\pi_0) = 
\textrm{DME}^{(2)}_t(\pi_e,\pi_0)+\textrm{DDE}^{(2)}_t(\pi_e,\pi_0)+\textrm{IME}^{(2)}_t(\pi_e,\pi_0)+\textrm{IDE}^{(2)}_t(\pi_e,\pi_0),
\end{align*} where
\begin{align*}
    \textrm{DME}^{(2)}_t(\pi_e,\pi_0) &= \Mean[R^{*}_t(\pi_e,\bar{M}^*_{t}(\pi_e))-R^{*}_t(\pi_{e},\bar{G}_{t}^{\pi_{0,e}^{t}})],\\
    \textrm{DDE}^{(2)}_t(\pi_e,\pi_0) &=\Mean[R^{*}_t(\pi_{e},\bar{G}_{t}^{\pi_{0,e}^{t}})-R^{*}_t(\pi_{0,e}^{t},\bar{M}^*_{t}(\pi_{0,e}^{t}))],\\
    \textrm{IME}^{(2)}_t(\pi_e,\pi_0) &= \Mean[R^{*}_t(\pi_{0,e}^{t},\bar{M}^*_{t}(\pi_{0,e}^{t}))-R^{*}_t(\pi_{0,e}^{t},\bar{M}^*_{t}(\pi_{0}))],\\
    \textrm{IDE}^{(2)}_t(\pi_e,\pi_0) &= \Mean[R^{*}_t(\pi_{0,e}^{t},\bar{M}^*_{t}(\pi_{0}))-R^{*}_t(\pi_{0},\bar{M}^*_{t}(\pi_{0}))].
\end{align*}
Then, for $X\in\{\textrm{IDE}^{(2)},\textrm{IME}^{(2)},\textrm{DDE}^{(2)},\textrm{DME}^{(2)}\}$, we have that 
\begin{align}
    X = \lim_{T\to \infty}\frac{1}{T}\sum_{t=0}^{T-1}X_t.
\end{align}
By replacing $\pi_e$ and $\pi_0$ with $\bar{a}'_t$ and $\bar{a}_t$, $\textrm{IDE}^{(2)}$ and $\textrm{IME}^{(2)}$ are equivalent to NDE and NIE derived in \citet{pearl2022direct}. Let $\tilde{\bar{a}}_t = \{\bar{a}_{t-1},a'_t\}$, we further replace $\pi_{0,e}^t$ with $\tilde{\bar{a}}_t$ to define $\textrm{DDE}^{(2)}$ and $\textrm{DME}^{(2)}$ for fixed action sequnces. When $t>0$, if we set $\tilde{\bar{a}}_t = \bar{a}_t$, $\textrm{DDE}^{(2)}$ and $\textrm{DME}^{(2)}$ are equivalent to NDE/NIE defined in \citet{zheng2017longitudinal}.

\subsection{MR Estimators of the Alternative Decomposition}
Similar to Section \ref{TRE}, we first define three additional $Q$ functions: 
\begin{align*}
    Q^{\tilde{G}_0}(s,a,m) & = \sum_{t\ge 0}\Mean^{\pi_0} [\Mean^{\pi_e}_{a^*}r(S_{t},a^*,M_{t})-\eta^{\tilde{G}_0}|S_0=s, A_0=a, M_0=m],\\
    Q^{\pi_{0,e}}(s,a,m) & = \sum_{t\ge 0}\Mean^{\pi_0}[\Mean^{\pi_e}_{a^*,m^*} r(S_t,a^*,m^*)-
	\eta^{\pi_{0,e}}|S_0=s, A_0=a, M_0=m],\\
    Q^{\tilde{G}_e}(s,a,m) & = \sum_{t\geq 0} \Mean^{\tilde{G}}[\Mean^{\pi_e}_{a^*,m^*} r(S_t,a^*,m^*)-\eta^{\tilde{G}_e}|S_0=s, A_0=a, M_0=m], 
\end{align*} where $\eta^{\tilde{G}_0}$ is the expected value of $\Mean^{\pi_e}_{a^*}r(S_{t},a^*,M_{t})$ under policy $\pi_0$, $\eta^{\pi_{0,e}}$ is the expectation of $\Mean^{\pi_e}_{a^*,m^*} r(S_t,a^*,m^*)$ under $\pi_0$, and $\eta^{\tilde{G}_e}$ is the expectation of $\Mean^{\pi_e}_{a^*,m^*} r(S_t,a^*,m^*)$ under the treatment process and the intervened mediator process of $\tilde{G}$.

Next, we construct three additional augmentation terms similar to the augmentation terms defined in the main text. Let $\rho^{(2)}(S,A,M)=\frac{\sum_a \pi_0(a|S) p(M|S,a)}{p(M|S,A)}$. We define that
\begin{align*}
    I_6(O) & =  \omega^{\pi_0}(S)\frac{\pi_0(A|S)}{\pi_b(A|S)}\Big\{\Mean^{\pi_e}_{a'} r(S,a',M)+\Mean^{\pi_0}_{a,m} Q^{\tilde{G}_0}(S',a,m)-
	\Mean_{m}Q^{\tilde{G}_0}(S,A,m)-\eta^{\tilde{G}_0}\Big\}\\
    &+\omega^{\pi_0}(S)\frac{\pi_e(A|S)}{\pi_b(A|S)} \rho^{(2)}(S,A,M)\{R-r(S,A,M)\},\\
 I_7(O) & = \omega^{\pi_0}(S)\frac{\pi_0(A|S)}{\pi_b(A|S)}\Big\{\Mean^{\pi_e}_{a',m}r(S,a',m)+\Mean^{\pi_0}_{a,m} Q^{\pi_{0,e}}(S',a,m)
	- \Mean_{m} Q^{\pi_{0,e}}(S,A,m)-\eta^{\pi_{0,e}}\Big\}\\
 &+ \omega^{\pi_0}(S)\frac{\pi_e(A|S)}{\pi_b(A|S)} \{R-\Mean_{m}r(S,A,m)\},\\
 I_8(O) & = \omega^{\tilde{G}}(S)\frac{\pi_e(A|S)}{\pi_{b}(A|S)}\rho^{(2)}(S,A,M) \Big\{\Mean^{\pi_e}_{a',m}r(S,a',m)+ \Mean^{\tilde{G}}_{a,m}Q^{\tilde{G}_e}(S',a,m)-Q^{\tilde{G}_e}(S,A,M)-\eta^{\tilde{G}_e}\Big\}
    \\&+ \omega^{\tilde{G}}(S)\frac{\pi_e(A|S)}{\pi_{b}(A|S)}\Big\{R-\Mean_{m}r(S,A,m)\Big\} \\
    &+ \omega^{\tilde{G}}(S)\frac{\pi_0(A|S)}{\pi_{b}(A|S)}\times \sum_a \pi_e(a|S)\Big[Q^{\tilde{G}_e}(S,a,M)-\sum_m p(m|A,S)Q^{\tilde{G}_e}(S,a,m)\Big]
\end{align*}

Then the MR estimator of $\textrm{IDE}^{(2)}(\pi_e,\pi_0)$ is 
\begin{align*}
    \textrm{MR-IDE}^{(2)}(\pi_e,\pi_0) = \frac{1}{NT}\sum_{i,t}\eta^{\tilde{G}_0} - \eta^{\pi_0}+I_6(O_{i,t})-I_5(O_{i,t}).
\end{align*} 
The MR estimator of $\textrm{IME}^{(2)}(\pi_e,\pi_0)$ is 
\begin{align*}
    \textrm{MR-IME}^{(2)}(\pi_e,\pi_0) = \frac{1}{NT}\sum_{i,t}\eta^{\pi_{0,e}} - \eta^{\tilde{G}_0}+I_7(O_{i,t})-I_6(O_{i,t}).
\end{align*} 
The MR estimator of $\textrm{DDE}^{(2)}(\pi_e,\pi_0)$ is 
\begin{align*}
    \textrm{MR-DDE}^{(2)}(\pi_e,\pi_0) = \frac{1}{NT}\sum_{i,t}\eta^{\tilde{G}_e} - \eta^{\pi_{0,e}}+I_8(O_{i,t})-I_7(O_{i,t}).
\end{align*} 
The MR estimator of $\textrm{DME}^{(2)}(\pi_e,\pi_0)$ is 
\begin{align*}
    \textrm{MR-IDE}^{(2)}(\pi_e,\pi_0) = \frac{1}{NT}\sum_{i,t}\eta^{\pi_e} - \eta^{\tilde{G}_e}+I_1(O_{i,t})-I_8(O_{i,t}).
\end{align*} 
Following Theorem \ref{thm:robust} and Theorem \ref{thm:efficiency}, we can show that $\textrm{MR-IDE}^{(2)}$, $\textrm{MR-IME}^{(2)}$, $\textrm{MR-DDE}^{(2)}$, and $\textrm{MR-DME}^{(2)}$ are multiply robust and achieve the semi-parametric efficiency bound.

%% file: Appendix/Theorem1.tex
\section{Proof of Theorem \ref{thm:identification}} \label{proof:identification}

This proof adheres strictly to the definitions of potential outcomes discussed in Appendix \ref{graphical PO}. 

We first clarify three standard assumptions, and then identify the potential rewards $\Mean \left[R^{*}_t(\pi,\bar{M}^{*}_{t}(\pi))\right]$ for any arbitrary policy $\pi$ and $\Mean \left[R^{*}_t(\pi_{0},\bar{G}_{t}^{\pi_{e,0}^t})\right]$ using the observed data distribution, followed by the identification function for each of the $\textrm{IDE}(\pi_e,\pi_0)$, $\textrm{IME}(\pi_e,\pi_0)$, $\textrm{DDE}(\pi_e,\pi_0)$, and $\textrm{DME}(\pi_e,\pi_0)$. 

\subsection{Standard Assumptions}
The decomposed effects are identifiable under three standard assumptions \citep{zheng2017longitudinal,luckett2019estimating}:

\textbf{Assumption 1 (Consistency).} 
$\forall t, M_t = M_t^*(\bar{A}_t)$, $R_t = R^{*}_t(\bar{A}_{t},\bar{M}_{t})$, and $S_{t+1} = S^{*}_{t+1}(\bar{A}_{t},\bar{M}_{t})$.


\textbf{Assumption 2 (Sequential Randomization).}  $\forall j\geq t$, i) $\{R^*_{j}(\bar{a}_{j},\bar{m}_j),S^*_{j+1}(\bar{a}_{j},\bar{m}_j)\}\indep A_{t}|\bar{A}_{t-1},\bar{M}_{t-1},\bar{R}_{t-1},\bar{S}_{t}$; 
ii) $M^*_{j}(\bar{a}_{j})\indep A_{t}|\bar{A}_{t-1},\bar{M}_{t-1},\bar{R}_{t-1},\bar{S}_{t}$; 
and iii) $\{R^*_{j}(\bar{a}_{j},\bar{m}_j), S^*_{j+1}(\bar{a}_{j},\bar{m}_j)\}\indep M_{t}|\bar{A}_{t},\bar{M}_{t-1},\bar{R}_{t-1},\bar{S}_{t}$

\textbf{Assumption 3 (Positivity).} Let $h_{t} = (\bar{m}_{t},\bar{r}_{t},\bar{s}_{t+1})$. For all $t\geq 0$ and all ($h_t,\bar{a}_t,\bar{a}'_t)$: i) if $p^{\pi_b}(\bar{a}_t,h_t)>0$, then $p^{\pi_b}(a_{t+1}|\bar{a}_t,h_t)>0$; ii) if $p^{\pi_b}(\bar{a}'_t,h_t)>0$, then $p^{\pi_b}(a'_{t+1}|\bar{a}'_t,h_t)>0$; iii) if $p^{\pi_b}(r_t,s_{t+1}|\bar{a}_t,h_{t-1},m_t)>0$, then $p^{\pi_b}(r_t,s_{t+1}|\bar{a}'_t,h_{t-1},m_t)>0$; and iv) if $p^{\pi_b}(m_t|\bar{a}'_t,h_{t-1})>0$, then $p^{\pi_b}(m_t|\bar{a}_t,h_{t-1})>0$.


Assumption 1 states that the observed mediator, state, and reward are equivalent to their counterfactuals, which would be observed if the observed actions were carried out, and that the observed reward and state are consistent with the potential reward and state if the observed sequences of actions and mediators were taken.
Assumption 2 requires that there are no unmeasured confounders between $A_t$ and all of its subsequent covariates and between $M_t$ and all of its subsequent covariates. Lastly, assumption 3 ensures that treatments and covariates are not exclusive to a specific stratum of covariates. The identification result is summarized as follows.

\subsection{Identification of $\Mean \left[R^{*}_t(\pi,\bar{M}^{*}_{t}(\pi))\right]$} \label{Identification 1}
Without loss of generality, we first consider the states and mediators in discrete values. By definition, we have that 
\begin{align*}
    \Mean \left[R^{*}_t(\pi,\bar{M}^{*}_{t}(\pi))\right] =& \sum_{\bar{a}_t,\bar{m}_t,\bar{s}_{t+1}, \bar{r}_t} r_t \prob(S_0=s_0) \prod_{j=0}^{t}\pi(a_j|S^*_j(\bar{a}_{j-1}, \bar{M}^{*}_{j-1}(\bar{a}_{j-1})) = s_j) \\
    &\times \prob[M^*_j(\bar{a}_j) = m_j|\bar{S}^*_j(\bar{a}_{j-1}, \bar{M}^{*}_{j-1}(\bar{a}_{j-1})) = \bar{s}_j, \bar{M}^{*}_{j-1}(\bar{a}_{j-1}) = \bar{m}_{j-1}]\\
    &\times \prob[S^*_{j+1}(\bar{a}_{j}, \bar{M}^{*}_{j}(\bar{a}_{j}))  = s_{j+1}, R^{*}_j(\bar{a}_j,\bar{M}^{*}_{j}(\bar{a}_j)) = r_j|\bar{S}^*_j(\bar{a}_{j-1}, \bar{M}^{*}_{j-1}(\bar{a}_{j-1})) = \bar{s}_j, \bar{M}^{*}_{j}(\bar{a}_{j}) = \bar{m}_{j}].
\end{align*}
To identify the potential reward, we first consider $t=0$ and observe that
\begin{align*}
    \pi(a_t|S^*_t(\bar{a}_{t-1}, \bar{M}^{*}_{t-1}(\bar{a}_{t-1})) = s_t) = \pi(a_0|S_0 = s_0).
\end{align*} Next, we show that
\begin{align*}
    \prob[M^*_0(a_0) = m_0|S_0 = s_0] &= \prob[M^*_0(a_0) = m_0|A_0 = a_0, S_0 = s_0]\\
    & =\prob[M_0 = m_0|A_0 = a_0, S_0 = s_0],
\end{align*} where the first equality holds by Assumption 2 and the second equality follows from the Assumption 1. Similarly, using the same arguments, we can show that
\begin{align*}
    &\prob[S^*_1(a_0, M^*_0(a_0)) = s_1, R^*_0(a_0, M^*_0(a_0)) = r_0|S_0 = s_0, M^*_0(a_0) = m_0] \\
    &= \prob[S^*_1(a_0, M^*_0(a_0)) = s_1, R^*_0(a_0, M^*_0(a_0)) = r_0|A_0 = a_0, S_0 = s_0, M^*_0(a_0) = m_0]\\
    & =\prob[S_1 = s_1, R_0 = r_0|A_0 = a_0, S_0 = s_0, M_0 = m_0].
\end{align*} Applying the same arguments for the subsequent potential covariates repeatedly, we can show that 
\begin{align*}
    \Mean \left[R^{*}_t(\pi,\bar{M}^{*}_{t}(\pi))\right] =& \sum_{\bar{a}_t,\bar{m}_t,\bar{s}_{t+1}, r_t} r_t \prob(S_0=s_0) \prod_{j=0}^{t}\pi(a_j|S_j = s_j) \prob[M_j = m_j|\bar{A}_j = \bar{a}_j, \bar{S}_j = \bar{s}_j, \bar{M}_{j-1} = \bar{m}_{j-1}] \\
    &\times\prob[S_{j+1} = s_{j+1}, R_j = r_j|\bar{A}_j = \bar{a}_j, \bar{S}_j = \bar{s}_j, \bar{M}_{j}= \bar{m}_{j}].
\end{align*}
Finally, under the assumption that the data generating process satisfied the Markov property, such that i) the distribution of $A_t$ is independent of all the past history observations given $S_t$, ii) the distribution of $M_t$ is independent of all the past history observations given $(S_t,A_t)$, and iii) the distributions of $R_t$ and $S_{t+1}$ are independent of all the past history observations given $(S_t,A_t, M_t)$, we have that
\begin{align*}
    \Mean \left[R^{*}_t(\pi,\bar{M}^{*}_{t}(\pi))\right] =& \sum_{\bar{a}_t,\bar{m}_t,\bar{s}_{t+1}, r_t} r_t \prob(S_0=s_0)\prod_{j=0}^{t}\pi(a_j|S_j = s_j) \prob[M_j = m_j|A_j = a_j, S_j = s_j] \\
    &\times \prob[S_{j+1} = s_{j+1}, R_j = r_j|A_j = a_j, S_j = s_j, M_{j}= m_{j}].
\end{align*}
Let $\tau_t$ denote the data trajectory $\{(s_j,a_j,m_j,r_j,s_{j+1})\}_{0\le j \le t}$. Replacing the probability mass functions by probability density functions, we have that
\begin{align*}
    \Mean \left[R^{*}_t(\pi,\bar{M}^{*}_{t}(\pi))\right] &= \sum_{\tau_t}r_t \prod_{j=0}^{t}p(s_{j+1},r_j|s_j,a_j,m_j)p(m_j|s_j,a_j)\pi(a_j|s_j)\nu(s_0)\\
    &= \sum_{\tau_t}r_t p(s_{t+1},r_t|s_t,a_t,m_t)p(m_t|s_t,a_t)\pi(a_t|s_t)\prod_{j=0}^{t-1}p^{\pi}(s_{j+1},r_j,m_j,a_j|s_j)\nu(s_0),
\end{align*} the identifiability of which is guaranteed by Assumption 3.

When $\pi = \pi_e$,
\begin{align*}
    \Mean \left[R^{*}_t(\pi_e,\bar{M}^{*}_{t}(\pi_e))\right] &=\sum_{\tau_t}r_t p(s_{t+1},r_t|s_t,a_t,m_t)p(m_t|s_t,a_t)\pi_e(a_t|s_t)\prod_{j=0}^{t-1}p^{\pi_e}(s_{j+1},r_j,m_j,a_j|s_j)\nu(s_0).
\end{align*} 
When $\pi = \pi_0$,
\begin{align*}
    \Mean \left[R^{*}_t(\pi_0,\bar{M}^{*}_{t}(\pi_0))\right] &=\sum_{\tau_t}r_t p(s_{t+1},r_t|s_t,a_t,m_t)p(m_t|s_t,a_t)\pi_0(a_t|s_t)\prod_{j=0}^{t-1}p^{\pi_0}(s_{j+1},r_j,m_j,a_j|s_j)\nu(s_0).
\end{align*} 
When $\pi = \pi_{e,0}^{t}$,
\begin{align*}
    \Mean \left[R^{*}_t(\pi_{e,0}^{t},\bar{M}^{*}_{t}(\pi_{e,0}^{t}))\right] &=\sum_{\tau_t}r_t p(s_{t+1},r_t|s_t,a_t,m_t)p(m_t|s_t,a_t)\pi_0(a_t|s_t)\prod_{j=0}^{t-1}p^{\pi_e}(s_{j+1},r_j,m_j,a_j|s_j)\nu(s_0).
\end{align*} 
Following the same arguments, we can show that 
\begin{align*}
    \Mean \left[R^{*}_t(\pi_{e,0}^{t},\bar{M}^{*}_{t}(\pi_e))\right] = \sum_{\tau_t}\sum_{s^*,r^*,a'}r^* p(s^*,r^*|s_t,a',m_t)\pi_0(a'|s_t)p(m_t|s_t,a_t)\pi_e(a_t|s_t)\prod_{j=0}^{t-1}p^{\pi_e}(s_{j+1},r_j,m_j,a_j|s_j)\nu(s_0).
\end{align*}

\subsection{Identification of $\Mean [R^{*}_t(\pi_{0},\bar{G}_{t}^{\pi_{e,0}^t})]$}

Without loss of generality, we first consider the states and mediators in discrete values. Let $\tilde{\bar{a}}_t = (\bar{a}'_{t-1},a_t)$. By definition, we have that 
\begin{align}
    \Mean [R^{*}_t(\pi_{0},\bar{G}_{t}^{\tilde{\bar{a}}_t})] =& \sum_{\bar{a}_t,\bar{a}'_{t-1},\bar{m}_t,\bar{s}_{t+1}, \bar{r}_t} r_t \prob(S_0=s_0) \prod_{j=0}^{t-1}\pi(a_j|S^*_j(\bar{a}_{j-1}, \bar{G}_{j-1}^{\tilde{\bar{a}}_t})) = s_j) \pi(a'_j|S^*_j(\bar{a}_{j-1}, \bar{G}_{j-1}^{\tilde{\bar{a}}_t})) = s_j)\\
    &\times \prob[ G_{j}^{\tilde{\bar{a}}_t} = m_j|\bar{S}^*_j(\bar{a}_{j-1},  \bar{G}_{j-1}^{\tilde{\bar{a}}_t}) = \bar{s}_j,  \bar{G}_{j-1}^{\tilde{\bar{a}}_t} = \bar{m}_{j-1}]\\
    &\times \prob[S^*_{j+1}(\bar{a}_{j},  \bar{G}_{j}^{\tilde{\bar{a}}_t})  = s_{j+1}, R^{*}_j(\bar{a}_j, \bar{G}_{j}^{\tilde{\bar{a}}_t}) = r_j|\bar{S}^*_j(\bar{a}_{j-1},  \bar{G}_{j-1}^{\tilde{\bar{a}}_t}) = \bar{s}_j,  \bar{G}_{j}^{\tilde{\bar{a}}_t} = \bar{m}_{j}] \label{probS}\\
    & \times \pi(a_t|S^*_t(\bar{a}_{t-1}, \bar{G}_{t-1}^{\tilde{\bar{a}}_t}) = s_t) \prob[ G_{t}^{\tilde{\bar{a}}_t} = m_t|\bar{S}^*_t(\bar{a}_{t-1},  \bar{G}_{t-1}^{\tilde{\bar{a}}_t}) = \bar{s}_t, \bar{G}_{t-1}^{\tilde{\bar{a}}_t} = \bar{m}_{t-1}]\\
    &\times \prob[S^*_{t+1}(\bar{a}_{t},\bar{G}_{t}^{\tilde{\bar{a}}_t})  = s_{t+1}, R^{*}_t(\bar{a}_t, \bar{G}_{t}^{\tilde{\bar{a}}_t}) = r_t|\bar{S}^*_t(\bar{a}_{t-1},  \bar{G}_{t-1}^{\tilde{\bar{a}}_t}) = \bar{s}_t,  \bar{G}_{t}^{\tilde{\bar{a}}_t} = \bar{m}_{t}].
\end{align}
For $j<t$, By the definition of $\bar{G}_{j}^{\tilde{\bar{a}}_t}$, we have that
\begin{align}\label{PO_M1}
    \prob[ G_{j}^{\tilde{\bar{a}}_t} = m_j|\bar{S}^*_j(\bar{a}_{j-1},  \bar{G}_{j-1}^{\tilde{\bar{a}}_t}) = \bar{s}_j,  \bar{G}_{j-1}^{\tilde{\bar{a}}_t} = \bar{m}_{j-1}] = \prob[ M^*_j(\bar{a}'_j) = m_j|\bar{S}^*_j(\bar{a}'_{j-1},\bar{M}^*_{j-1}(\bar{a}'_{j-1})) = \bar{s}_j, \bar{M}^*_{j-1}(\bar{a}'_{j-1}) = \bar{m}_{j-1}].
\end{align} Using the same arguments in \ref{Identification 1}, we can show that equation (\ref{PO_M1}) equals
\begin{align*}
    \prob[M_j = m_j|\bar{A}_j = \bar{a}'_j, \bar{S}_j = \bar{s}_j, \bar{M}_{j-1} = \bar{m}_{j-1}],
\end{align*} which is identifiable under Assumption 3. 

Further, to show the identification of equation (\ref{probS}), we prove it at $j=0$ as follows:
\begin{align*}
    &\prob[S^*_{1}(a_0,G_{0}^{\tilde{\bar{a}}_t})  = s_1, R^{*}_0(a_0, G_{0}^{\tilde{\bar{a}}_t}) = r_0|S_0=s_0,  G_{0}^{\tilde{\bar{a}}_t} = m_{0}]  \\
    &= \prob[S^*_{1}(a_0,m_0)  = s_1, R^{*}_0(a_0, m_0) = r_0|S_0=s_0,  G_{0}^{\tilde{\bar{a}}_t} = m_{0}]\\
    &= \prob[S^*_{1}(a_0,m_0)  = s_1, R^{*}_0(a_0, m_0) = r_0|S_0=s_0]\\
    & =\prob[S_1 = s_1, R_0 = r_0|A_0 = a_0, S_0 = s_0, M_0 = m_0].
\end{align*} The second equality holds by the definition of the process $G_{t}^{\tilde{\bar{a}}_t}$, in which we randomly draw $M_0$ from $G_{0}^{\tilde{\bar{a}}_t}$. Specifically, given $S_0 = s_0$, $G_{0}^{\tilde{\bar{a}}_t}$ is independent of $S^*_{1}(a_0,m_0)$ and $R^*_{1}(a_0,m_0)$. The last equality follows from Assumption 1 and 2. A similar proof can be found in \citet{zheng2017longitudinal}.

Then, following the steps in \ref{Identification 1}, we can show that
\begin{multline*}
    \Mean \left[R^{*}_t(\pi_{0},\bar{G}_{t}^{\pi_{e,0}^t})\right] =
        \sum_{\tau_t,\bar{a}^*_{t-1}}r_t p(s_{t+1},r_t,m_t|s_t,a_t)\pi_0(a_t|s_t)\\
        \prod_{j=0}^{t-1}p(s_{j+1},r_j|s_j,a_j,m_j)\pi_0(a_j|s_j)p(m_j|s_j,a_j^*)\pi_e(a_j^*|s_j)\nu(s_0),
\end{multline*} the identifiability of which is guaranteed by Assumption 3.

\subsection{Identification of $\textrm{IDE}(\pi_e,\pi_0)$, $\textrm{IME}(\pi_e,\pi_0)$, $\textrm{DDE}(\pi_e,\pi_0)$, $\textrm{DME}(\pi_e,\pi_0)$}
Using the above identification results, the identification functions of $\textrm{IDE}(\pi_e,\pi_0)$, $\textrm{IME}(\pi_e,\pi_0)$, $\textrm{DDE}(\pi_e,\pi_0)$, $\textrm{DME}(\pi_e,\pi_0)$ are directly induced. Specifically,
\begin{multline*}
\textrm{IDE}(\pi_e,\pi_0) = \lim_{T\to \infty}\frac{1}{T}\sum_{t=0}^{T-1} \sum_{\tau_t} \big\{r_t p(s_{t+1},r_t|s_t,a_t,m_t)-\sum_{s^*,r^*,a'}r^{*}p(s^*,r^*|s_t,a',m_t)\pi_0(a'|s_t)\big\}\\ \times p(m_t|s_t,a_t)\pi_e(a_t|s_t) \prod_{j=0}^{t-1}p^{\pi_e}(s_{j+1},r_j,m_j,a_j|s_j)\nu(s_0),
\end{multline*}
\begin{multline*}
    \textrm{IME}(\pi_e,\pi_0) = \lim_{T\to \infty}\frac{1}{T}\sum_{t=0}^{T-1} \sum_{\tau_t} r_t p(s_{t+1},r_t|s_t,a_t,m_t)\pi_0(a_t|s_t)[\sum_{a'}p(m_t|s_t,a')\pi_{e}(a'|s_t)-p(m_t|a_t,s_t)]\\
	\times \prod_{j=0}^{t-1}\left[p^{\pi_e}(s_{j+1},r_j,m_j,a_j|s_j)\right]\nu(s_0),
\end{multline*}
\begin{multline*}
    \textrm{DDE}(\pi_e,\pi_0) = \lim_{T\to \infty}\frac{1}{T}\sum_{t=0}^{T-1}\sum_{\tau_t}r_t p(s_{t+1},r_t|s_t,a_t,m_t)p(m_t|s_t,a_t)\pi_0(a_t|s_t)\\
    \times \Big\{\prod_{j=0}^{t-1}p^{\pi_e}(s_{j+1},r_j,m_j,a_j|s_j)
    -\sum_{\bar{a}^*_{t-1}}\prod_{j=0}^{t-1}p(s_{j+1},r_j|s_j,a_j,m_j)\pi_0(a_j|s_j)p(m_j|s_j,a_j^*)\pi_e(a_j^*|s_j)\Big\}\nu(s_0),
\end{multline*} and 
\begin{multline*}
    \textrm{DME}(\pi_e,\pi_0) = \lim_{T\to \infty}\frac{1}{T}\sum_{t=0}^{T-1} \sum_{\tau_t}r_t p(s_{t+1},r_t|s_t,a_t,m_t)p(m_t|s_t,a_t)\pi_0(a_t|s_t)\\
    \times \Big\{\sum_{\bar{a}^*_{t-1}}\prod_{j=0}^{t-1}p(s_{j+1},r_j|s_j,a_j,m_j)\pi_0(a_j|s_j)p(m_j|s_j,a_j^*)\pi_e(a_j^*|s_j)-
    \prod_{j=0}^{t-1}p^{\pi_0}(s_{j+1},r_j,m_j,a_j|s_j)\Big\}\nu(s_0).
\end{multline*}

The proof of Theorem \ref{thm:identification} is thus completed.

%% file: Appendix/Robustness.tex
\section{Proof of Theorem \ref{thm:robust}}\label{appendix:robust}
The proof of the triply robustness property of the proposed estimator is similar for $\textrm{IDE}(\pi_e,\pi_0)$, $\textrm{IME}(\pi_e,\pi_0)$, $\textrm{DDE}(\pi_e,\pi_0)$ and $\textrm{DME}(\pi_e,\pi_0)$. Here, we take the estimator of IDE as an example. Let $O$ denote a data tuple $(S,A,M,R,S')$, $\rho(S,A,M)=\frac{\sum_a \pi_e(a|S) p(M|S,a)}{p(M|S,A)}$, and $\delta^{\pi}(S,A) = \omega^{\pi}(S)\frac{\pi(A|S)}{\pi_b(A|S)}$ for any policy $\pi$. Without loss of generality, we let $T_i = T$, $\forall i = 1,\cdots, N$. We first reorganize the estimator of IDE into four parts. Recall that $\eta_d = \eta$. Let
\begin{align*}
\phi_{1}(O) &= \eta^{\pi_e}-\eta^{G_e},\\
\phi_{2}(O) &= \delta^{\pi_e}(S,A)\left[R+ \Mean_{\substack{a\sim \pi_e(\bullet|S')\\ m\sim p(\bullet|a,S')}} Q^{\pi_e}(S',a,m)-\Mean_{m\sim p(\bullet|A,S)}Q^{\pi_e}(S,A,m)-\eta^{\pi_e}\right],\\
\phi_{3}(O)  &= \delta^{\pi_e}(S,A) \rho(S,A,M)\frac{\pi_0(A|S)}{\pi_e(A|S)}\{R-r(S,A,M)\},\\
\phi_{4}(O)  &= \delta^{\pi_e}(S,A)\left[\Mean_{a\sim \pi_0(\bullet|S)} r(S,a',M)+\Mean_{\substack{a\sim \pi_e(\bullet|S')\\m\sim p(\bullet|S',a)}} Q^{G_e}(S',a,m)-\Mean_{m\sim p(\bullet|S,A)}Q^{G_e}(S,A,m)-\eta^{G_e}\right].
\end{align*}
Then the proposed MR estimator of IDE is
\begin{align*}
    \textrm{MR-IDE}(\pi_e,\pi_0) =\frac{1}{NT}\sum_{i,t}[\hat{\phi}_{1}(O_{i,t}) +\hat{\phi}_{2}(O_{i,t}) 
    - \hat{\phi}_{3} (O_{i,t})
    -\hat{\phi}_{4}(O_{i,t})].
\end{align*}

The proof of robustness can be divided into four parts. In \textbf{part I}, we show that when $\hat{\pi}_{b}$ and $\hat{\omega}^{\pi_{e}}$ are consistent, the sum of terms involving $Q^{\pi_{e}}$, $Q^{G_e}$, $\eta^{\pi_e}$, and $\eta^{G_e}$ converges to zero by the stationary property. Then, the remaining part of $\textrm{MR-IDE}(\pi_e,\pi_0)$ is
\begin{align}\label{remaining_part}
    \frac{1}{NT}\sum_{i,t}\underbrace{\hat{\delta}^{\pi_e}(S_{i,t},A_{i,t})\left[R_{i,t}-\Mean_{a\sim \pi_0(\bullet|S)} \hat{r}(S_{i,t},a',M_{i,t})\right]}_{\hat{\phi}_{5}(O_{i,t})}-\hat{\phi}_{3}(O_{i,t}).
\end{align}

In \textbf{part II}, we consider the condition $\mathbb{M}_{1}$, where  $\hat{\pi}_{b}$, $\hat{\omega}^{\pi_{e}}$, and $\hat{r}$ are consistent. We show that $\frac{1}{NT}\sum_{i,t}\hat{\phi}_{3} (O_{i,t})$ converged to $0$, and $\frac{1}{NT}\sum_{i,t}\hat{\phi}_{5} (O_{i,t})$ is unbiased to the IS estimator with correctly specified $\pi_{b}$, $\omega^{\pi_{e}}$, and $r$ and thus unbiased and consistent to $\textrm{IDE}(\pi_e,\pi_0)$, using the arguments used in part I. Together with the results from part I, the consistency of our estimator is proved.

In \textbf{part III}, we focus on the condition $\mathbb{M}_{2}$, where $\hat{\pi}_{b}$, $\hat{\omega}^{\pi_{e}}$, and $\hat{p}_m$ are consistent. We show that (\ref{remaining_part}) is consistent to $\textrm{IDE}(\pi_e,\pi_0)$. The consistency is then completed, together with part I.

Finally, in \textbf{part IV}, applying similar arguments in part I, we observe that $\frac{1}{NT}\sum_{i,t}\hat{\phi}_{2} (O_{i,t})$, $\frac{1}{NT}\sum_{i,t}\hat{\phi}_{3} (O_{i,t})$, and $\frac{1}{NT}\sum_{i,t}\hat{\phi}_{4} (O_{i,t})$ converge to 0 respectively, when $\hat{Q}^{\pi_{e}}$, $\hat{Q}^{G_e}$, $\hat{\eta}^{\pi_e}$, $\hat{\eta}^{G_e}$, $\hat{r}$, and $\hat{p}_{m}$ are consistent. Then, we show that $\textrm{MR-IDE}(\pi_e,\pi_0) = \hat{\phi}_{1}$ is consistent to $\textrm{IDE}(\pi_e,\pi_0)$, with consistent $\hat{\eta}^{\pi_e}$ and $\hat{\eta}^{G_e}$. The consistency of the proposed estimator is thus proved, and the proof of triply-robustness is thus completed.

We next detail the proof for each part.

\textbf{Part I.} Condition: $\hat{\pi}_b$ and $\hat{\omega}^{\pi_e}$ are consistent.

First, we focus on the terms involving $Q^{\pi_e}$. Let $f_1(O;\omega^{\pi_e},\pi_b,p_m,Q^{\pi_e})$ denotes
\begin{align*}
    \delta^{\pi_e}(A|S) \left[\Mean_{\substack{a\sim \pi_e(\bullet|S')\\ m\sim p(\bullet|a,S')}}Q^{\pi_e}(S',a,m)-\Mean_{m\sim p(\bullet|A,S)}Q^{\pi_e}(S,A,m)\right].
\end{align*} To show that $\frac{1}{NT}\sum_{i,t}f_1(O_{i,t};\hat{\omega}^{\pi_e},\hat{\pi}_b,\hat{p}_m,\hat{Q}^{\pi_e})$ converges to 0, when $\pi_{b}$ and $\omega^{\pi_{e}}$ are consistent, we decompose it into 
\begin{align*}
    \underbrace{\frac{1}{NT}\sum_{i,t}f_1(O_{i,t};\hat{\omega}^{\pi_e},\hat{\pi}_b,\hat{p}_m,\hat{Q}^{\pi_e})-\frac{1}{NT}\sum_{i,t}f_1(O_{i,t};\omega^{\pi_e},\hat{\pi}_b,\hat{p}_m,\hat{Q}^{\pi_e})}_{\Gamma_1} \\
    +\underbrace{\frac{1}{NT}\sum_{i,t}f_1(O_{i,t};\omega^{\pi_e},\hat{\pi}_b,\hat{p}_m,\hat{Q}^{\pi_e}) - \frac{1}{NT}\sum_{i,t}f_1(O_{i,t};\omega^{\pi_e},\pi_b,\hat{p}_m,\hat{Q}^{\pi_e})}_{\Gamma_2}\\
    +\underbrace{\frac{1}{NT}\sum_{i,t}f_1(O_{i,t};\omega^{\pi_e},\pi_b,\hat{p}_m,\hat{Q}^{\pi_e})}_{\Gamma_3}.
\end{align*} It suffices to show that $\Gamma_1$, $\Gamma_2$, and $\Gamma_3$ all converge to zero in probability.

Let us focus on $\Gamma_1$ first. Under the assumptions that $\Omega^{\pi_e}$, $\mathcal{Q}^{\pi_e}$, $\mathcal{H}_m$, and $\Pi_b$ are all bounded function classes and $\hat{\pi}_b(A_{i,t}|S_{i,t})$ is uniformly bounded away from zero, $|\Gamma_1|$ is upper bounded by 
\begin{align}\label{G1_ub}
    \frac{O(1)}{NT}\sum_{i,t}|\hat{\omega}^{\pi_e}(S_{i,t})-\omega^{\pi_e}(S_{i,t})|, 
\end{align} where $O(1)$ is some positive constant. By Markov's inequality, to prove (\ref{G1_ub}) converges to zero in probability, it suffices to show that 
\begin{align}\label{G1_ub2}
    \frac{1}{NT}\Mean\sum_{i,t}|\hat{\omega}^{\pi_e}(S_{i,t})-\omega^{\pi_e}(S_{i,t})| = o(1).
\end{align}

For any sufficient small constant $\epsilon>0$, let $\Omega^{\pi_e}(\epsilon)$ defines a set of function $\omega$, such that,
\begin{align}
    \Mean_{s\sim p_{\infty}}|\omega(s) - \omega^{\pi_e}(s)|^{2}\leq \epsilon^{2},
\end{align} where $p_{\infty}$ denotes the limiting distribution of state under behavior policy. Since $\hat{\omega}^{\pi_e}$ is consistent and converge to $\omega^{\pi_e}$ in $L_2$-norm, we can show that $\hat{\omega}^{\pi_e} \in \Omega^{\pi_e}(\epsilon)$ with probability approaching to 1 (wpa1) for large $NT$, by Markov's inequality. Therefore, the right-hand side (RHS) of (\ref{G1_ub2}) is upper bounded by 
\begin{align}\label{G1_ub3}
    \frac{1}{NT}\Mean \sup_{\omega\in\Omega^{\pi_e}(\epsilon)}\sum_{i,t}|\omega(S_{i,t})-\omega^{\pi_e}(S_{i,t})|,
\end{align} wpa1. Then, it suffices to show that (\ref{G1_ub3}) is $o_p(1)$.

Implementing the empirical process theory \citep{van1996weak}, we first decompose (\ref{G1_ub3}) into 
\begin{align*}
    \underbrace{\frac{1}{NT}\Mean \sup_{\omega\in\Omega^{\pi_e}(\epsilon)}\left\{\sum_{i,t}|\omega(S_{i,t})-\omega^{\pi_e}(S_{i,t})| - \Mean \sum_{i,t}|\omega(S_{i,t})-\omega^{\pi_e}(S_{i,t})|\right\}}_{\Gamma_4} +
    \underbrace{\frac{1}{NT}\sup_{\omega\in\Omega^{\pi_e}(\epsilon)}\left\{\Mean \sum_{i,t}|\omega(S_{i,t})-\omega^{\pi_e}(S_{i,t})|\right\}}_{\Gamma_5}.
\end{align*} By the definition of $\Omega^{\pi_e}(\epsilon)$ and the Cauchy Schwartz inequality, $\Mean|\omega(S_{i,t})-\omega^{\pi_e}(S_{i,t})| \leq \epsilon$ for any $\omega\in\Omega^{\pi_e}(\epsilon)$. Thus, $\Gamma_5$ is upper bounded by $\epsilon$ and converges to zero when $\epsilon \to 0$ (i.e., $\Gamma_5 = o(1)$). 

Next, we show that $\Gamma_4$ converges to zero as well. Under the assumption that $\Omega^{\pi_e}(\epsilon)$ is a VC-type classes with VC indices upper bounded by $O(N^{k})$ for $k<\frac{1}{2}$ and $\epsilon$ is sufficiently small, using the maximal inequality (See Section 4.2 in \citet{dedecker2002maximal} and Corollary 5.1 in \citet{chernozhukov2014gaussian}), we can show that $\sqrt{NT}\Gamma_4$ converges to zero (i.e., $\sqrt{NT}\Gamma_4 = o_p(1)$). Therefore, we have that, $\Gamma_4 = o_p(\frac{1}{\sqrt{NT}})$. The proof of $\Gamma_1 = o_p(1)$ is then completed.

Similarly, following the steps to prove $\Gamma_1 = o_p(1)$, we can show that $\Gamma_2 = o_p(1)$. Then, it remains to show that $\Gamma_3 = o_p(1)$. By Markov's inequality, it suffices to show that $\Mean(\Gamma_3)=o(1)$. By the definition of $\Gamma_3$, $\Mean(\Gamma_3)$ is upper bounded by 
\begin{align} \label{G3_ub}
    \frac{1}{NT}\Mean \sup_{\tilde{p}\in\mathcal{H}_m, Q\in\mathcal{Q}}\sum_{i,t}f_1(O_{i,t};\omega^{\pi_e},\pi_b,\tilde{p},Q).
\end{align} We first observe that, for any $Q\in\mathcal{Q}^{\pi_e}$ and $ \tilde{p} \in \mathcal{H}_m$, the expectation of $\Gamma_3$ is zero. Specifically, 
\begin{align*}
    &\Mean\Big[\omega^{\pi_e}(S)\frac{\pi_e(A|S)}{\pi_b(A|S)} \Mean_{\substack{a\sim \pi_e(\bullet|S')\\ m\sim \tilde{p}(\bullet|a,S')}}Q(S',a,m)-\Mean_{m\sim \tilde{p}(\bullet|A,S)}Q(S,A,m)\Big]\\
   = &\sum_a\int_{s,m,s'} p(m,s'|a,s)p^{\pi_{e}}(s)\pi_{e}(a|s) \sum_{a'}\int_{m'}Q(s',a',m')\tilde{p}(m'|a',s')\pi_e(a'|s')\\
   &\left.-\sum_a\int_{s,m,s'} p(m,s'|a,s)p^{\pi_{e}}(s)\pi_{e}(a|s)\int_{m'} Q(s,a,m')\tilde{p}(m'|a,s)\right.\\
   =&\sum_{a'}\int_{s',m'} p^{\pi_{e}}(s')\pi_{e}(a'|s')Q(s',a',m')\tilde{p}(m'|a',s') \\
   &\left.- \sum_a\int_{s,m'} p^{\pi_{e}}(s)\pi_{e}(a|s)Q(s,a,m')\tilde{p}(m'|a,s)\right.\\
   = &0.
\end{align*} Then, following the same steps we used to bound (\ref{G1_ub3}), we can show that (\ref{G3_ub}) is $o(1)$. Thus, $\Gamma_3 = o_p(1)$. Together with $\Gamma_1 = o_p(1)$ and $\Gamma_2 = o_p(1)$, we finish the proof of $\frac{1}{NT}\sum_{i,t}f_1(O_{i,t};\hat{\omega}^{\pi_e},\hat{\pi}_b,\hat{p}_m,\hat{Q}^{\pi_e}) = o_p(1)$.

Then we focus on the terms involving $Q^{G_e}$. Let $f_2(O;\omega^{\pi_e},\pi_b,p_m,Q^{G_e})$ denotes
\begin{align*}
    \delta^{\pi_e}(A|S) \left[\Mean_{\substack{a\sim \pi_e(\bullet|S')\\ m\sim p(\bullet|a,S')}}Q^{G_e}(S',a,m)-\Mean_{m\sim p(\bullet|A,S)}Q^{G_e}(S,A,m)\right].
\end{align*} Replacing $Q^{\pi_e}(S,A,m)$ with $Q^{G_e}(S,A,m)$ in the proof of $\frac{1}{NT}\sum_{i,t}f_1(O_{i,t};\hat{\omega}^{\pi_e},\hat{\pi}_b,\hat{p}_m,\hat{Q}^{\pi_e}) = o_p(1)$, we can directly show that $\frac{1}{NT}\sum_{i,t}f_2(O_{i,t};\hat{\omega}^{\pi_e},\hat{\pi}_b,\hat{p}_m,\hat{Q}^{G_e}) = o_p(1)$ as well.

Finally, we need to show that the sum of terms involving $\eta^{\pi_e}$ and $\eta^{G_e}$ converges to zero. Let $f_3(O;\omega^{\pi_e},\pi_b,\eta^{\pi_e},\eta^{G_e})$ denotes
\begin{align*}
\left[1-\omega^{\pi_e}(S)\frac{\pi_e(A|S)}{\pi_b(A|S)}\right](\eta^{\pi_e}-\eta^{G_e}).
\end{align*} For any $\eta_{1}\in\real$ and $\eta_{2}\in\real$, $\frac{1}{NT}\sum_{i,t}f_3(O_{i,t};\omega^{\pi_e},\pi_b,\eta_{1},\eta_{2})$ has mean zero. Specifically,
\begin{align*}
    &\Mean[\eta_1-\eta_2-\omega^{\pi_e}(S)\frac{\pi_e(A|S)}{\pi_b(A|S)}(\eta_1-\eta_2)]\\
    =&\Big\{1-\sum_a\int_s p^{\pi_b}(a,s) \omega^{\pi_e}(s)\frac{\pi_e(a|s)}{\pi_b(a|s)}\Big\}(\eta_1-\eta_2)\\
    =& 0 \times (\eta_1-\eta_2)\\
    =& 0 .
\end{align*} Applying the same arguments in showing that $\Gamma_3=o_p(1)$, we can show that $\frac{1}{NT}\sum_{i,t}f_3(O_{i,t};\omega^{\pi_e},\pi_b,\hat{\eta}^{\pi_e},\hat{\eta}^{G_e}) = o_p(1)$. Then, following the same steps proving that $\Gamma_1=o_p(1)$, we can show that 
\begin{align*}
    \frac{1}{NT}\sum_{i,t}\left\{f_3(O_{i,t};\hat{\omega}^{\pi_e},\hat{\pi}_b,\hat{\eta}^{\pi_e},\hat{\eta}^{G_e})-f_3(O_{i,t};\omega^{\pi_e},\hat{\pi}_b,\hat{\eta}^{\pi_e},\hat{\eta}^{G_e})\right\} = o_p(1),
\end{align*}and 
\begin{align*}
    \frac{1}{NT}\sum_{i,t}\left\{f_3(O_{i,t};\omega^{\pi_e},\hat{\pi}_b,\hat{\eta}^{\pi_e},\hat{\eta}^{G_e})-f_3(O_{i,t};\omega^{\pi_e},\pi_b,\hat{\eta}^{\pi_e},\hat{\eta}^{G_e})\right\} = o_p(1).
\end{align*} Therefore, $\frac{1}{NT}\sum_{i,t}f_3(O_{i,t};\hat{\omega}^{\pi_e},\hat{\pi}_b,\hat{\eta}^{\pi_e},\hat{\eta}^{G_e}) = o_p(1)$. The proof of part I is thus completed.

\textbf{Part II.}
Condition: $\hat{\pi}_{b}(A|S)$, $\hat{\omega}^{\pi_{e}}(S)$, and $\hat{r}$ are consistent.

With true $r$, $\omega^{\pi_e}$, and $\pi_b$, we can show that $\Mean\phi_{3}(O_{i,t};\omega^{\pi_e},\pi_b, \hat{p}_m,r)$ has a mean of zero, as $\Mean [R-r(s,a,m)|S=s,A=a,M=m] = 0$. Then, using the same arguments in showing that $\Gamma_3 = o_p(1)$ in part I, we can show that $\frac{1}{NT}\sum_{i,t}\hat{\phi}_{3} (O_{i,t};\omega^{\pi_e},\pi_b, \hat{p}_m,r) = o_p(1)$. Next, following the same steps proving that $\Gamma_1 = o_p(1)$, we can show that 
\begin{align*}
    \frac{1}{NT}\sum_{i,t}\hat{\phi}_{3} (O_{i,t};\hat{\omega}^{\pi_e},\hat{\pi}_b, \hat{p}_m,\hat{r}) - \frac{1}{NT}\sum_{i,t}\hat{\phi}_{3} (O_{i,t};\omega^{\pi_e},\pi_b, \hat{p}_m,r)= o_p(1).
\end{align*} Therefore, we finish the proof showing that $\frac{1}{NT}\sum_{i,t}\hat{\phi}_{3} (O_{i,t};\hat{\omega}^{\pi_e},\hat{\pi}_b, \hat{p}_m,\hat{r}) = o_p(1)$. Then, it remains to show that $\frac{1}{NT}\sum_{i,t}\hat{\phi}_{5} (O_{i,t};\hat{\omega}^{\pi_e},\hat{\pi}_b, \hat{p}_m,\hat{r})$ is consistent to $\textrm{IDE}(\pi_e,\pi_0)$. Again, applying the arguments used in showing that $\Gamma_1=o_p(1)$, we can show that 
\begin{align*}
    \frac{1}{NT}\sum_{i,t}\hat{\phi}_{5} (O_{i,t};\hat{\omega}^{\pi_e},\hat{\pi}_b, \hat{p}_m,\hat{r}) - \frac{1}{NT}\sum_{i,t}\hat{\phi}_{5} (O_{i,t};\omega^{\pi_e},\pi_b, \hat{p}_m,r) = o_p(1).
\end{align*} Then, it suffices to show that $\frac{1}{NT}\sum_{i,t}\hat{\phi}_{5} (O_{i,t};\omega^{\pi_e},\pi_b, \hat{p}_m,r)$ is consistent to $\textrm{IDE}(\pi_e,\pi_0)$. Specifically,
\begin{align}
    \frac{1}{NT}\sum_{i,t}\hat{\phi}_{5} (O_{i,t};\omega^{\pi_e},\pi_b, \hat{p}_m,r) = \frac{1}{NT}\sum_{i,t}\omega^{\pi_e}\frac{\pi_e(A_{i,t}|S_{i,t})}{\pi_b(A_{i,t}|S_{i,t})}\left[R_{i,t}-\sum_{a}\pi_0(a|S_{i,t})r(S_{i,t},a,M_{i,t})\right].
\end{align} Under the assumption of stationary state process, since the action space is finite, it suffices to show that,
\begin{align}\label{consistency_1}
    \Mean_{\substack{s\sim \hat{p}^{\pi_e}\\m\sim \hat{p}_m}}\omega^{\pi_e}\frac{\pi_e(a|s)}{\pi_b(a|s)}\left[r-\sum_{a'}\pi_0(a'|s)r(s,a',m)\right]\xrightarrow{P}\Mean_{\substack{s\sim p^{\pi_e}\\m\sim p_m}}\omega^{\pi_e}\frac{\pi_e(a|s)}{\pi_b(a|s)}\left[r-\sum_{a'}\pi_0(a'|s)r(s,a',m)\right]
\end{align} for any $a$. By the weak law of large number, we can show that (\ref{consistency_1}) holds when $NT$ is sufficiently large.
Together with the results in part I, we thus complete the proof of Part II.

\textbf{Part III.}
Condition: $\hat{\pi}_{b}(A|S)$, $\hat{\omega}^{\pi_{e}}(S)$, and $\hat{p}_m$ are consistent. 

Applying the same arguments used in showing that $\Gamma_1 = o_p(1)$, we can show that 
\begin{align*}
    \frac{1}{NT}\sum_{i,t}\hat{\phi}_{5} (O_{i,t};\hat{\omega}^{\pi_e},\hat{\pi}_b, \hat{p}_m,\hat{r}) - \frac{1}{NT}\sum_{i,t}\hat{\phi}_{5} (O_{i,t};\omega^{\pi_e},\pi_b, p_m,\hat{r}) = o_p(1),
\end{align*}and 
\begin{align*}
    \frac{1}{NT}\sum_{i,t}\hat{\phi}_{3} (O_{i,t};\hat{\omega}^{\pi_e},\hat{\pi}_b, \hat{p}_m,\hat{r}) - \frac{1}{NT}\sum_{i,t}\hat{\phi}_{3} (O_{i,t};\omega^{\pi_e},\pi_b, p_m,\hat{r}) = o_p(1).
\end{align*} Then, it suffices to show that 
\begin{align}\label{consistency_2}
    \frac{1}{NT}\sum_{i,t}\hat{\phi}_{5} (O_{i,t};\omega^{\pi_e},\pi_b, p_m,\hat{r}) - \frac{1}{NT}\sum_{i,t}\hat{\phi}_{3} (O_{i,t};\omega^{\pi_e},\pi_b, p_m,\hat{r}) \overset{p}{\to} \textrm{IDE}(\pi_e,\pi_0).
\end{align}
The LHS of (\ref{consistency_2}) can be decomposed into two parts. Specifically, it suffices to show that
\begin{align}\label{consistency_3}
    \frac{1}{NT}\sum_{i,t}\delta^{\pi_e}(S_{i,t},A_{i,t})\left\{\Mean_{a'\sim \pi_0(\bullet|S_{i,t})} \hat{r}(S_{i,t},a',M_{i,t}) -  \rho(S_{i,t},A_{i,t},M_{i,t})\frac{\pi_0(A_{i,t}|S_{i,t})}{\pi_e(A_{i,t}|S_{i,t})}\hat{r}(S_{i,t},A_{i,t},M_{i,t})\right\} = o_p(1),
\end{align} and 
\begin{align}\label{consistency_4}
    \frac{1}{NT}\sum_{i,t}\delta^{\pi_e}(S_{i,t},A_{i,t})\left\{R_{i,t} - \rho(S_{i,t},A_{i,t},M_{i,t})\frac{\pi_0(A_{i,t}|S_{i,t})}{\pi_e(A_{i,t}|S_{i,t})}R_{i,t}\right\}\xrightarrow{P}\textrm{IDE}(\pi_e,\pi_0).
\end{align} Following the steps showing that $\Gamma_3 = o_p(1)$ in part I, since the expectation of the LHS of (\ref{consistency_3}) is $0$, we can show that (\ref{consistency_3}) holds. Furthermore, applying the arguments used in showing (\ref{consistency_1}) in part II, we can show that (\ref{consistency_4}) holds. Together with the results in part I, we thus complete the proof of Part III.

\textbf{Part IV.}
Condition: $\hat{Q}^{\pi_{e}}$, $\hat{Q}^{G_e}$, $\hat{\eta}^{\pi_e}$, $\hat{\eta}^{G_e}$, $\hat{r}$, and $\hat{p}_{m}$ are consistent.

As we discussed in the main context, with true $Q^{\pi_{e}}$, $Q^{G_e}$, $\eta^{\pi_e}$, $\eta^{G_e}$, $r$, and $p_{m}$, we can show that $\Mean\hat{\phi}_j(O_{i,t};Q^{\pi_{e}},Q^{G_e},\eta^{\pi_e},\eta^{G_e},r,p_{m},\hat{\omega}^{\pi_e},\hat{\pi}_b) = 0$ for $j=2,3,4$. Then, using the same arguments in showing that $\Gamma_3 = o_p(1)$ in part 1, we can show that 
\begin{align*}
    \frac{1}{NT}\sum_{i,t}\hat{\phi}_j(O_{i,t};Q^{\pi_{e}},Q^{G_e},\eta^{\pi_e},\eta^{G_e},r,p_{m},\hat{\omega}^{\pi_e},\hat{\pi}_b) = o_p(1), \text{ for $j=2,3,4$}.
\end{align*} Then, applying the arguments used in showing that $\Gamma_1 = o_p(1)$, we can further show that 
\begin{align*}
        \frac{1}{NT}\sum_{i,t}\left\{\hat{\phi}_j(O_{i,t};\hat{Q}^{\pi_{e}},\hat{Q}^{G_e},\hat{\eta}^{\pi_e},\hat{\eta}^{G_e},\hat{r},\hat{p}_{m},\hat{\omega}^{\pi_e},\hat{\pi}_b) - \hat{\phi}_j(O_{i,t};Q^{\pi_{e}},Q^{G_e},\eta^{\pi_e},\eta^{G_e},r,p_{m},\hat{\omega}^{\pi_e},\hat{\pi}_b)\right\} = o_p(1),
\end{align*}for $j=2,3,4$. These two results further yields that
\begin{align*}
    \frac{1}{NT}\sum_{i,t}\hat{\phi}_j(O_{i,t};\hat{Q}^{\pi_{e}},\hat{Q}^{G_e},\hat{\eta}^{\pi_e},\hat{\eta}^{G_e},\hat{r},\hat{p}_{m},\hat{\omega}^{\pi_e},\hat{\pi}_b) = o_p(1)
\end{align*} for $j=2,3,4$. Then, it remains to show that $\hat{\phi}_1(\hat{\eta}^{\pi_e},\hat{\eta}^{G_e})$ is consistent to $\textrm{IDE}(\pi_e,\pi_0)$. Applying the arguments used to show $\Gamma_1 = o_p(1)$ again, under the assumption that we have that $\hat{\eta}^{\pi_e}$ and $\hat{\eta}^{G_e}$ are consistent, 
\begin{align*}
    \hat{\phi}_1(\hat{\eta}^{\pi_e},\hat{\eta}^{G_e}) \xrightarrow{P} \hat{\phi}_1(\eta^{\pi_e},\eta^{G_e}) = \textrm{IDE}(\pi_e,\pi_0),
\end{align*} where the equation holds by definition. The proof of part IV is thus completed.



%% file: Appendix/Efficiency.tex
\section{Proof of Theorem \ref{thm:efficiency}}\label{appendix:efficiency}
First, we clarify the assumption of convergence. We required that each of $\hat{Q}^{(\cdot)}, \hat{\omega}^{(\cdot)}, \hat{p}_m, \hat{r}, \hat{\pi}_b$, and $\hat{\eta}^{(\cdot)}$ converges to its corresponding oracle value in $L_2$-norm at a rate of $N^{-k^*}$, for some $k^*>1/4$. Specifically, taking $\hat{\omega}^{\pi_e}$ as an example, we assume that
\begin{align*}
    \sqrt{\Mean_{s\sim p_\infty}|\hat{\omega}^{\pi_e}(s)-\omega^{\pi_e}(s)|} = O_p(N^{-k^*}).
\end{align*}

The proof of the efficiency of the proposed estimator is similar for $\textrm{IDE}(\pi_e,\pi_0)$, $\textrm{IME}(\pi_e,\pi_0)$, $\textrm{DDE}(\pi_e,\pi_0)$, and $\textrm{DME}(\pi_e,\pi_0)$. Here, we take the MR estimator of IDE as an example. Adopting the notation used in the Appendix \ref{appendix:robust}, we have the proposed multiply robust estimator of IDE as
\begin{align*}
    \textrm{MR-IDE}(\pi_e,\pi_0) = \frac{1}{NT}\sum_{i,t}[\hat{\phi}_{1}(O_{i,t})+\hat{\phi}_{2}(O_{i,t}) 
    - \hat{\phi}_{3} (O_{i,t})
    -\hat{\phi}_{4}(O_{i,t})].
\end{align*} Taking the oracle values of the estimators (i.e., $Q^{\pi_{e}},Q^{G_e},\eta^{\pi_e},\eta^{G_e},r,p_{m},\omega^{\pi_e},\pi_b$), we define the oracle estimator as $\textrm{MR-IDE}^*(\pi_e,\pi_0) =\frac{1}{NT}\sum_{i,t}[\hat{\phi}^*_{1}(O_{i,t}) + \hat{\phi}^*_{2}(O_{i,t}) 
    - \hat{\phi}^*_{3} (O_{i,t})
    -\hat{\phi}^*_{4}(O_{i,t})]$.

We decompose the proof into two parts. In part I, we show that the proposed estimator is asymptotically equivalent to the oracle estimator, such that $\textrm{MR-IDE}(\pi_e,\pi_0)-\textrm{MR-IDE}^*(\pi_e,\pi_0) = o_p(\frac{1}{\sqrt{NT}})$. In part II, we show that the oracle estimator is asymptotically normal such that $\sqrt{N}[\textrm{MR-IDE}^*(\pi_e,\pi_0)-\textrm{IDE}(\pi_e,\pi_0)]\xrightarrow{d}N(0,\sigma_T^2)$, where $\sigma_T^2$ is the semiparametric efficiency bound. Noticing that $\psi_2(O_{i,t})$, $\psi_3(O_{i,t})$, and $\psi_4(O_{i,t})$ are the martingale difference sequence with respect to $\{O_{i,t}\}_{0\leq t \leq T-1}$, under the assumption of stationarity, we have that
\begin{align*}
    \sigma^2_T = \frac{1}{T}Var\left[\phi_{2}(O_{t}) 
    - \phi_{3} (O_{t})
    -\phi_{4}(O_{t})\right].
\end{align*} Therefore, we have that 
\begin{align*}
    \sqrt{NT}[\textrm{MR-IDE}^*(\pi_e,\pi_0)-\textrm{IDE}(\pi_e,\pi_0)]\xrightarrow{d}N(0,\sigma^2),
\end{align*} where $\sigma^2 = Var[\phi_{2}(O_{t}) 
    - \phi_{3} (O_{t})
    -\phi_{4}(O_{t})]$. 
Finally, by Slutsky's theorem, the proposed estimator is asymptotically normally distributed with mean $0$ and a variance achieving the semiparametric efficiency bound. Specifically, 
\begin{align*}
    \sqrt{NT}\Big[\textrm{MR-IDE}(\pi_e,\pi_0)-\textrm{IDE}(\pi_e,\pi_0)\Big]\xrightarrow{d}N(0,\sigma^2).
\end{align*} 
In the following, we detail the proof of each part.

\textbf{Part I.} Let $\hat{\psi}=\{\hat{Q}^{\pi_{e}}, \hat{Q}^{G_e}, \hat{\eta}^{\pi_e}, \hat{\eta}^{G_e}, \hat{p}_{m}\}$. We first decompose the $\textrm{MR-IDE}(\pi_e,\pi_0)-\textrm{MR-IDE}^*(\pi_e,\pi_0)$ in to three parts, such that
$\textrm{MR-IDE}(\pi_e,\pi_0)-\textrm{MR-IDE}^*(\pi_e,\pi_0) = \textrm{MR-IDE}^{(1)}(\hat{\psi}) + \textrm{MR-IDE}^{(2)}(\hat{\psi}) + \textrm{MR-IDE}^{(3)}(\hat{\psi},\hat{r})$, where
\begin{align*}
    \textrm{MR-IDE}^{(1)}(\hat{\psi}) = \frac{1}{NT}\sum_{i,t}\Big\{\sum_{j=1}^{2}[\hat{\phi}_{1}(\hat{\psi},\omega^{\pi_e},\pi_b,r)- \hat{\phi}^*_{1} (O_{i,t})]  - \sum_{j=3}^{4}[\hat{\phi}_{j}(O_{i,t};\hat{\psi},\omega^{\pi_e},\pi_b,r) -\hat{\phi}^*_{j}(O_{i,t})]\Big\},
\end{align*}
\begin{multline*}
    \textrm{MR-IDE}^{(2)}(\hat{\psi}) = \frac{1}{NT}\sum_{i,t}\Big\{\sum_{j=1}^{2}[\hat{\phi}_{1}(\hat{\psi},\omega^{\pi_e},\pi_b,\hat{r})- \hat{\phi}_{1} (\hat{\psi},\omega^{\pi_e},\pi_b,r)]\\
    - \sum_{j=3}^{4}[\hat{\phi}_{j}(O_{i,t};\hat{\psi},\omega^{\pi_e},\pi_b,\hat{r}) -\hat{\phi}_{j}(O_{i,t};\hat{\psi},\omega^{\pi_e},\pi_b,r)]\Big\},
\end{multline*} and 
\begin{multline*}
    \textrm{MR-IDE}^{(3)}(\hat{\psi},\hat{r}) = \frac{1}{NT}\sum_{i,t}\Big\{\sum_{j=1}^{2}[\hat{\phi}_{1}(\hat{\psi},\hat{\omega}^{\pi_e},\hat{\pi}_b,\hat{r})- \hat{\phi}_{1} (\hat{\psi},\omega^{\pi_e},\pi_b,\hat{r})] \\- \sum_{j=3}^{4}[\hat{\phi}_{j}(O_{i,t};\hat{\psi},\hat{\omega}^{\pi_e},\hat{\pi}_b,\hat{r}) -\hat{\phi}_{j}(O_{i,t};\hat{\psi},\omega^{\pi_e},\pi_b,\hat{r})]\Big\}.
\end{multline*}
Following the arguments in part I and part II of the proof of Robustness in Appendix \ref{appendix:robust}, the expectation of $\textrm{MR-IDE}^{(2)}(\hat{\psi})$ is  zero. Then, applying the same arguments used in showing that $\Gamma_3 = o_p(\frac{1}{\sqrt{NT}})$ in part I of the proof of Robustness, we can show that $\textrm{MR-IDE}^{(1)}(\hat{\psi}) = o_p(\frac{1}{\sqrt{NT}})$ under the assumption that each component in $\hat{\phi}$ converges to its oracle value in $L_2$ norm at a rate of $N^{-k^*}$ for $k^*>\frac{1}{4}$. 

Then, we focus on showing that $\textrm{MR-IDE}^{(2)}(\hat{\psi}) = o_p(\frac{1}{\sqrt{NT}})$. Noticing that $\textrm{MR-IDE}^{(2)}(\hat{\psi})$ can be further decomposed as
\begin{align*}
    \textrm{MR-IDE}^{(2)}(\hat{\psi}) - \textrm{MR-IDE}^{(2)}(\psi) + \textrm{MR-IDE}^{(2)}(\psi),
\end{align*} it suffices to show that $\textrm{MR-IDE}^{(2)}(\hat{\psi}) - \textrm{MR-IDE}^{(2)}(\psi) = o_p(\frac{1}{\sqrt{NT}})$ and $\textrm{MR-IDE}^{(2)}(\psi)= o_p(\frac{1}{\sqrt{NT}})$. First, similar to the part III of the proof of Theorem \ref{thm:robust}, the expectation of $\textrm{MR-IDE}^{(2)}(\psi)$ is $0$, for any $\hat{r}\in \mathcal{H}_r$. Then, applying the arguments used in showing that $\Gamma_3 = o_p(\frac{1}{\sqrt{NT}})$, we can show that $\textrm{MR-IDE}^{(2)}(\psi,r) = o_p(\frac{1}{\sqrt{NT}})$ under the assumption that $\hat{\omega}^{\pi_e}$ and $\hat{\pi}_b$ converge to their oracle values.  
Then, it remains to show that $\textrm{MR-IDE}^{(2)}(\hat{\psi})-\textrm{MR-IDE}^{(2)}(\psi) = o_p(\frac{1}{\sqrt{NT}})$. It suffices to show that 
\begin{align}\label{consistency_5}
    \frac{1}{NT}\sum_{i,t}[\hat{\phi}_{j}(O_{i,t};\hat{\psi},\omega^{\pi_e},\pi_b,\hat{r}) -\hat{\phi}_{j}(O_{i,t};\hat{\psi},\omega^{\pi_e},\pi_b,r)]-[\hat{\phi}_{j}(O_{i,t};\psi,\omega^{\pi_e},\pi_b,\hat{r}) -\hat{\phi}_{j}(O_{i,t};\psi,\omega^{\pi_e},\pi_b,r)] = o_p(\frac{1}{\sqrt{NT}}),
\end{align} for $j=1,2,3,4$. Here, we prove that the above equation holds for $j=3$ as an example. For $j=1,2,4$, the proof can be completed using similar arguments.

We first observe that the LHS of (\ref{consistency_5}) is upper bounded by
\begin{align*}
    &\frac{1}{NT}\sum_{i,t}|[\hat{\phi}_{j}(O_{i,t};\hat{\psi},\omega^{\pi_e},\pi_b,\hat{r}) -\hat{\phi}_{j}(O_{i,t};\hat{\psi},\omega^{\pi_e},\pi_b,r)]-[\hat{\phi}_{j}(O_{i,t};\psi,\omega^{\pi_e},\pi_b,\hat{r}) -\hat{\phi}_{j}(O_{i,t};\psi,\omega^{\pi_e},\pi_b,r)]|\\
    =& \frac{1}{NT}\sum_{i,t}|\delta^{\pi_e}(S_{i,t},A_{i,t})||\hat{\rho}(S_{i,t},A_{i,t},M_{i,t})-\rho(S_{i,t},A_{i,t},M_{i,t})|\frac{\pi_0(A_{i,t}|S_{i,t})}{\pi_e(A_{i,t}|S_{i,t})}|r(S_{i,t},A_{i,t},M_{i,t})-\hat{r}(S_{i,t},A_{i,t},M_{i,t})|\\
    \leq& \frac{C}{NT}\sum_{i,t}|\hat{\rho}(S_{i,t},A_{i,t},M_{i,t})-\rho(S_{i,t},A_{i,t},M_{i,t})||r(S_{i,t},A_{i,t},M_{i,t})-\hat{r}(S_{i,t},A_{i,t},M_{i,t})|\\
    \leq& \frac{C}{2NT}\sum_{i,t}|\hat{\rho}(S_{i,t},A_{i,t},M_{i,t})-\rho(S_{i,t},A_{i,t},M_{i,t})|^2+\frac{C}{2NT}\sum_{i,t}|r(S_{i,t},A_{i,t},M_{i,t})-\hat{r}(S_{i,t},A_{i,t},M_{i,t})|^2\\
    =&o_p(\frac{1}{\sqrt{NT}}),
\end{align*} where $C$ is some positive constant. The first inequality holds under the assumption that $\Omega^{\pi_e}$ and $\Pi_b$ are bounded function classes of $\omega^{\pi_e}$ and $\pi_b$, respectively. The second inequality holds by applying the Cauchy-Schwartz inequality such that $ab\leq \frac{a^2+b^2}{2}$.
Using the similar arguments used to bound (\ref{G1_ub2}) in part I of the proof of Theorem \ref{thm:robust}, under the assumption that $\hat{p}_m$ and $\hat{r}$ converge to their oracle values respectively in $L_2$ norm at a rate of $O_p(N^{-k^*})$ for some $k^*>1/4$, we can show that the final equality holds. Similarly, we can show that $\textrm{MR-IDE}^{(3)}(\hat{\psi},\hat{r}) = o_p(\frac{1}{\sqrt{NT}})$ as well. The proof of part I is thus completed. 

\textbf{Part II.}
By Central Limit Theorem, when $N\to\infty$, we can show that 
\begin{align*}
    \sqrt{N}[\textrm{MR-IDE}^*(\pi_e,\pi_0)-\textrm{IDE}(\pi_e,\pi_0)]\xrightarrow{d}N(0,\sigma_T^2),
\end{align*} for some variance $\sigma_T^2$. Then it remains to show that $\sigma_T^2$ achieves the asymptotic semiparametric efficiency bound, which is the supreme of the Cramer-Rao lower bounds for all parametric submodels \citep{newey1990semiparametric}.

We first introduce some additional notations. Let $\pi_{b,\theta}$, $p_{m,\theta}$ and $p_{s',r,\theta}$, and $\nu_{\theta}$ be some parametric models parameterized by $\theta$ for $\pi_{b}$, $p_{m}$ and $p_{s',r}$, and $\nu$, and $\mathcal{M}$ denotes the set of all such parametric models. Then, by Theorem 1, $\textrm{IDE}(\pi_e,\pi_0)$ can be represented as a function of $\theta$. We denote the $\textrm{IDE}(\pi_e,\pi_0)$ parameterized by $\theta$ as $\textrm{IDE}_\theta(\pi_e,\pi_0)$. By definition, the Cramer-Rao lower bound for an unbiased estimator is 
\begin{align*}
    CR(\pi_{b,\theta}, p_{m,\theta}, p_{s',r,\theta}, \nu_{\theta}) = \frac{\partial \textrm{IDE}_\theta(\pi_e,\pi_0)}{\partial \theta}\left(\Mean\left\{\frac{\partial l(\{O_t\}_{0\leq t\leq T-1};\theta)}{\partial \theta}\frac{\partial l^{T}(\{O_t\}_{0\leq t\leq T-1};\theta)}{\partial \theta}\right\}\right)^{-1}\frac{\partial \textrm{IDE}_\theta(\pi_e,\pi_0)}{\partial \theta}^{T},
\end{align*} where $l(\{O_t\}_{0\leq t\leq T-1};\theta)$ is the log-likelihood function.

Suppose that there exists some parameter $\theta_0$ such that  $\pi_{b,\theta_0}$, $p_{m,\theta_0}$ and $p_{s',r,\theta_0}$, and $\nu_{\theta_0}$ are the corresponding true models. Then the semiparametric efficiency bound is 
\begin{align}\label{sup_CR_bound}
    \sup_{\mathcal{M}}CR = \sup_{\pi_{b}, p_{m}, p_{s',r}, \nu \in \mathcal{M}}CR(\pi_{b}, p_{m}, p_{s',r}, \nu) = CR(\pi_{b,\theta_0}, p_{m,\theta_0}, p_{s',r,\theta_0}, \nu_{\theta_0}).
\end{align} It suffices to show that $\sigma_T^2 = \sup_{\mathcal{M}}CR$.

On the one hand, from Appendix \ref{appendix:EIF}, we have that 
\begin{align}\label{efficient_score}
    \frac{\partial \textrm{IDE}_{\theta_0}(\pi_e,\pi_0)}{\partial \theta} = \Mean[(\eta^{\pi_e}-\eta^{G_e})S(\bar{O}_{T-1})]+D_1(\theta_0) - D_2(\theta_0),
\end{align} where $\bar{O}_{T-1}$ is the sequence of observations such that $\bar{O}_{T-1} = \{O_1,O_2,\cdots,O_{T-1}\}$, $S(\cdot)$ is the gradient of the log-likelihood function evaluated at $\theta = \theta_0$ (i.e., $\frac{\partial l(\{O_t\}_{0\leq t\leq T-1};\theta_0)}{\partial \theta}$),
\begin{eqnarray*}
     D_1(\theta_0) = \Mean\left[\frac{1}{T}\sum_{t=0}^{T-1}\omega^{\pi_e}(S_t)\frac{\pi_{e}(A_t|S_t)}{\pi_{b,\theta_0}(A_t|S_t)}\{R_t + \Mean^{\pi_e}_{a^*,m^*;\theta_0} Q^{\pi_e}(S_{t+1},a,m) - \Mean_{m;\theta_0} Q^{\pi_e}(S_t,A_t,m) -\eta^{\pi_e} \}S(\bar{O}_{T-1})\right],
\end{eqnarray*} and 
\begin{multline*}
    D_2(\theta_0) = \Mean\Big[\frac{1}{T}\sum_{t=0}^{T-1}\omega^{\pi_e}(S_t)\Big\{\frac{\sum_{a}p_{\theta_0}(M_t|S_t,a)\pi_e(a|S_t)}{p_{\theta_0}(M_t|S_t,A_t)}\frac{\pi_0(A_t|S_t)}{\pi_{b,\theta_0}(A_t|S_t)}[R_t-r_{\theta_0}(S_t,A_t,M_t)]+\frac{\pi_{e}(A_t|S_t)}{\pi_{b,\theta_0}(A_t|S_t)}\\
    \times \{\sum_{a'}r_{\theta_0}(S_t,a',M_t)\pi_0(a'|S_t)-\eta^{G_e} + \Mean_{a,m;\theta_0}^{\pi_e} Q^{G_e}(S_{t+1},a,m) - \Mean_{m;\theta_0} Q^{G_e}(S_t,A_t,m)\}\Big\}S(\bar{O}_{T-1})\Big].
\end{multline*} Adopting the notation used in Appendix \ref{appendix:efficiency}, (\ref{efficient_score}) can be rewritten as
\begin{align*}
    \Mean\left[\left\{\frac{1}{T}\sum_{t}[\phi_{1}(O_{t})+\phi_{2}(O_{t}) 
    - \phi_{3} (O_{t})
    -\phi_{4}(O_{t})]\right\}S(\bar{O}_{T-1})\right].
\end{align*}
Furthermore, since the expectation of a score function is 0, we can show that $\Mean[\textrm{IDE}_{\theta_0}(\pi_e,\pi_0)\times S(\bar{O}_{T-1})] = \textrm{IDE}_{\theta_0}(\pi_e,\pi_0)\times\Mean[S(\bar{O}_{T-1})] = 0$. Therefore, $\frac{\partial \textrm{IDE}_{\theta_0}(\pi_e,\pi_0)}{\partial \theta}$ can be further represented as 
\begin{align*}
    \Mean\left[\left\{ \frac{1}{T}\sum_{t}[\phi_{1}(O_t)+\phi_{2}(O_{t}) 
    - \phi_{3} (O_{t})
    -\phi_{4}(O_{t})]-\textrm{IDE}_{\theta_0}(\pi_e,\pi_0)\right\}S(\bar{O}_{T-1})\right].
\end{align*}
By Cauchy-Schwartz inequality \citep{tripathi1999matrix}, we have that
\begin{align*}
    \sup_{\mathcal{M}}CR \leq& \Mean\left[\left\{  \frac{1}{T}\sum_{t}[\phi_{1}(O_t)+\phi_{2}(O_{t}) 
    - \phi_{3} (O_{t})
    -\phi_{4}(O_{t})]-\textrm{IDE}_{\theta_0}(\pi_e,\pi_0)\right\}^2\right]\\
    =& Var \left\{  \frac{1}{T}\sum_{t}[\phi_{1}(O_t)+\phi_{2}(O_{t}) 
    - \phi_{3} (O_{t})
    -\phi_{4}(O_{t})]-\textrm{IDE}_{\theta_0}(\pi_e,\pi_0)\right\}\\
    =& \sigma_{T}^2.
\end{align*}

On the other hand, by Lemma 20 in \citet{kallus2022efficiently}, there exists model $\mathcal{M}_{\theta'} \in \mathcal{M}$ with sufficiently large number of parameters, having $CR(\pi_{b,\theta'}, p_{m,\theta'}, p_{s',r,\theta'}, \nu_{\theta'})=\sigma_T^2$. Therefore, we have that $\sigma_T^2 = \sup_{\mathcal{M}}CR$. The proof is thus completed.

%% file: Appendix/Derivation.tex
\section{Derivation of Efficient Influence Functions (EIF)}\label{appendix:EIF}
In this section, we focus on deriving the efficient influence function for each component of the average treatment effect. Without loss of generality, we assume that the state, action, mediator and reward are all discrete. While adopting the notations used in the Appendix \ref{appendix:efficiency}, we omit the subscript in $p_m$ and $p_{s',r}$ when there is no confusion. Let $\tau_t$ denote the data trajectory $\{(s_j,a_j,m_j,r_j,s_{j+1})\}_{0\le j \le t}$. 

\subsection{EIF for Immediate Direct Effect}
Let us first focus on the immediate direct effect (IDE). $\textrm{IDE}_{\theta_0}(\pi_e,\pi_0)$ can be represented as

\begin{multline}\label{DE}
	\lim_{T\to \infty}\frac{1}{T}\sum_{t=0}^{T-1} \sum_{\tau_t} \big\{r_t p_{\theta_0}(s_{t+1},r_t|s_t,a_t,m_t)-\sum_{s^*,r^*,a'}r^{*}p_{\theta_0}(s^*,r^*|s_t,a',m_t)\pi_0(a'|s_t)\big\}p_{\theta_0}(m_t|s_t,a_t)\pi_e(a_t|s_t) \\ \times \prod_{j=0}^{t-1}p_{\theta_0}^{\pi_e}(s_{j+1},r_j,m_j,a_j|s_j)\nu_{\theta_0}(s_0),
\end{multline}
where $\nu$ denotes the initial state distribution, and 
\begin{align*}
p_{\theta_0}^{\pi_e}(s_{j+1},r_j,m_j,a_j|s_j)=p_{\theta_0}(s_{j+1},r_j|s_j,a_j,m_j)p_{\theta_0}(m_j|s_j,a_j)\pi_e(a_j|s_j).
\end{align*}
Taking the derivative of (\ref{DE}), we have 
\begin{eqnarray*}
	\frac{\partial \textrm{IDE}_{\theta_0}(\pi_e,\pi_0)}{\partial \theta} = C_1 + D_1 - D_2,
\end{eqnarray*} where
\begin{eqnarray*}
    C_1 = (\ref{DE})\times \triangledown_{\theta} \log(\nu_{\theta_0}(s_0)),
\end{eqnarray*}
\begin{eqnarray*}
	 D_1 = \lim_{T\to \infty}\frac{1}{T}\sum_{t=0}^{T-1} \sum_{\tau_t} r_t \prod_{j=0}^{t}p_{\theta_0}^{\pi_e}(s_{j+1},r_j,m_j,a_j|s_j) \sum_{j=0}^{t}\left[ \triangledown_{\theta} \log p_{\theta_0}^{\pi_e}(s_{j+1},r_j,m_j,a_j|s_j)]\right.\times \nu_{\theta_0}(s_0),
\end{eqnarray*} and
\begin{multline}\label{I2:original}
	 D_2 = \lim_{T\to \infty}\frac{1}{T}\sum_{t=0}^{T-1} \sum_{a',\tau_t}r_t p_{\theta_0}(s_{t+1},r_t|s_t,a_t,m_t)\pi_0(a_t|s_t)p_{\theta_0}(m_t|s_t,a')\pi_e(a'|s_t)\\
	 \times \prod_{j=0}^{t-1}p_{\theta_0}^{\pi_e}(s_{j+1},r_j,m_j,a_j|s_j) \Big\{\triangledown_{\theta} \log p_{\theta_0}^{\pi_e}(s_{t+1},r_t|m_t,s_t,a_t)+\triangledown_{\theta} \log p_{\theta_0}(m_t|s_t, a')\\+\sum_{j=0}^{t-1}\left[ \triangledown_{\theta} \log p_{\theta_0}^{\pi_e}(s_{j+1},r_j,m_j|s_j,a_j)]\right.	 \Big\}\times \nu_{\theta_0}(s_0).
\end{multline}
In the following sections, we will derive $C_1$, $D_1$, and $D_2$, respectively.

\subsubsection{$C_1$}
We first focus on $C_1$. Since the expectation of a score function is zero, we have that
\begin{align*}
    C_1 = \Mean[\textrm{IDE}_{\theta_0}(\pi_e,\pi_0) \times \triangledown_{\theta} \log(\nu_{\theta_0}(s_0))] = \Mean[\textrm{IDE}_{\theta_0}(\pi_e,\pi_0) \times S(\bar{O}_{T-1})] = \Mean[(\eta^{\pi_e} - \eta^{G_e}) \times S(\bar{O}_{T-1})].
\end{align*}

\subsubsection{$D_1$}\label{Derivation:D_1}
We then focus on the derivation of $D_1$. Notice that
\begin{align*}
	 &\lim_{T\to \infty}\frac{1}{T}\sum_{t=0}^{T-1} \sum_{\tau_t} \eta^{\pi_e} \prod_{j=0}^{t}p_{\theta_0}^{\pi_e}(s_{j+1},r_j,m_j,a_j|s_j) \sum_{j=0}^{t}\left[ \triangledown_{\theta} \log p_{\theta_0}^{\pi_e}(s_{j+1},r_j,m_j,a_j|s_j)]\right. \nu_{\theta_0}(s_0),\\
	 = &\lim_{T\to \infty}\frac{1}{T}\sum_{t=0}^{T-1} \eta^{\pi_e} \Mean[\sum_{j=0}^{t} \triangledown_{\theta} \log p_{\theta_0}^{\pi_e}(s_{j+1},r_j,m_j,a_j|s_j)] \times \nu_{\theta_0}(s_0), \\
	 = & 0,
\end{align*}where the last equation holds using the fact that the expectation of a score function is $0$.
Therefore, 
\begin{multline*}
	 D_1 = \lim_{T\to \infty}\frac{1}{T}\sum_{t=0}^{T-1} \sum_{\tau_t} [r-\eta^{\pi_e}] \prod_{j=0}^{t}\left. p_{\theta_0}^{\pi_e}(s_{j+1},r_j,m_j,a_j|s_j)\right.  \sum_{j=0}^{t}\left[ \triangledown_{\theta} \log p_{\theta_0}^{\pi_e}(s_{j+1},r_j,m_j,a_j|s_j)]\right. \nu_{\theta_0}(s_0).
\end{multline*}
Together with the trick of the equality $\star$ (See Appendix \ref{proof:star_equality} for a complete proof of it), we have that
\begin{align}
	 D_1 \stackrel{\star}{=} \lim_{T\to \infty}\frac{1}{T}\sum_{j=0}^{T-1} \sum_{\tau_{j}} &[r-\eta^{\pi_e} + \Mean^{\pi_e}_{a^*,m^*} Q^{\pi_e}(s_{j+1},a^*,m^*)] \prod_{k=0}^{j}\left. p_{\theta_0}^{\pi_{e}}(s_{k+1},r_k,m_k,a_k|s_k)\right. \notag\\
	 &\times \triangledown_{\theta} \log p_{\theta_0}^{\pi_e}(s_{j+1},r_j,m_j,a_j|s_j)\times \nu_{\theta_0}(s_0)\label{I1:2nd}.
\end{align}
Then, we note that 
\begin{align*}
    \sum_{s_0}\prod_{k=0}^{j} p_{\theta_0}^{\pi_{e}}(s_{k+1},r_k,m_k,a_k|s_k)\nu_{\theta_0}(s_0) \stackrel{\star \star}{=} p_{\theta_0}^{\pi_{e}}(s_{j+1},r_j,m_j,a_j|s_j)p^{\pi_e}(s_j),
\end{align*} which is the probability of $\{S_{j+1} = s_{j+1}, R_j = r_j, M_j = m_j, A_j = a_j\}$ under the target polity $\pi_e$.
Further, we notice that
\begin{align*}
    \triangledown_{\theta} \log p_{\theta_0}^{\pi_e}(s_{j+1},r_j,m_j,a_j|s_j) = 
    \triangledown_{\theta}\log p_{\theta_0}(s_{j+1},r_j,m_j|a_j,s_j)
\end{align*}
Using the fact that the expectation of a score function is $0$, we have
\begin{align*}
    \sum_{s_{j+1},r_j,m_j} [p_{\theta_0}(s_{j+1},r_j,m_j|a_j,s_j)\triangledown_{\theta} \log p_{\theta_0}^{\pi_e}(s_{j+1},r_j,m_j|a_j,s_j)] = 0
\end{align*} for any $j$, which follows that
\begin{align*}
	 &\lim_{T\to \infty}\frac{1}{T}\sum_{j=0}^{T-1} \sum_{\tau_{j}} \Mean_{m^*} Q^{\pi_e}(s_{j},a_{j},m^*) p_{\theta_0}^{\pi_e}(s_{j+1},r_j,m_j,a_j|s_j)p^{\pi_e}(s_j) \times \triangledown_{\theta} \log p_{\theta_0}^{\pi_e}(s_{j+1},r_j,m_j|a_j,s_j)\\
	 =&\begin{aligned}\lim_{T\to \infty}\frac{1}{T}\sum_{j=0}^{T-1} \sum_{\tau_{j-1},a_{j},s_{j}} &\Mean_{m^*} Q^{\pi_e}(s_{j},a_{j},m^*) \pi_{e}(a_j|s_j) p^{\pi_e}(s_j)\\ &\times \sum_{s_{j+1},r_j,m_j} [p_{\theta_0}(s_{j+1},r_j,m_j|a_j,s_j)\triangledown_{\theta} \log p_{\theta_0}^{\pi_e}(s_{j+1},r_j,m_j|a_j,s_j)]\end{aligned}\\
	  =& 0
\end{align*}
Thus, combined with the $D_1$ in equation (\ref{I1:2nd}), we have that
\begin{eqnarray*}
	 D_1 = \lim_{T\to \infty}\frac{1}{T}\sum_{j=0}^{T-1} \sum_{\tau_{j}} [r_j-\eta^{\pi_e} + \Mean^{\pi_e}_{a^*,m^*} Q^{\pi_e}(s_{j+1},a^*,m^*) - \Mean_{m^*} Q^{\pi_e}(s_{j},a_j,m^*)]\\ \times p_{\theta_0}^{\pi_e}(s_{j+1},r_j,m_j,a_j|s_j)p^{\pi_e}(s_j) 
	  \triangledown_{\theta} \log p_{\theta_0}(s_{j+1},r_j,m_j|a_j,s_j),\\
	 =\lim_{T\to \infty}\frac{1}{T}\sum_{j=0}^{T-1} \sum_{\tau_{j}} [r_j-\eta^{\pi_e} + \Mean^{\pi_e}_{a^*,m^*} Q^{\pi_e}(s_{j+1},a^*,m^*) - \Mean_{m^*} Q^{\pi_e}(s_{j},a_j,m^*)]\\
	 \times\frac{\pi_{e}(a_j|s_j)p^{\pi_e}(s_j) }{\pi_{b,\theta_0}(a_j|s_j)p^{\pi_b}(s_j)} p_{\theta_0}(s_{j+1},r_j,m_j|a_j,s_j)\pi_{b,\theta_0}(a_j|s_j)p^{\pi_b}(s_j)
	 \triangledown_{\theta} \log p_{\theta_0}(s_{j+1},r_j,m_j|a_j,s_j),\\
	 =\lim_{T\to \infty}\frac{1}{T}\sum_{j=0}^{T-1} \sum_{\tau_{j}} [r_j-\eta^{\pi_e} + \Mean^{\pi_e}_{a^*,m^*} Q^{\pi_e}(s_{j+1},a^*,m^*) - \Mean_{m^*} Q^{\pi_e}(s_{j},a_j,m^*)]\\
	 \times\frac{\pi_{e}(a_j|s_j)p^{\pi_e}(s_j) }{\pi_{b,\theta_0}(a_j|s_j)p^{\pi_b}(s_j)} p_{\theta_0}(s_{j+1},r_j,m_j|a_j,s_j)\pi_{b,\theta_0}(a_j|s_j)p^{\pi_b}(s_j)
	 \triangledown_{\theta} \log p_{\theta_0}^{\pi_b}(s_{j+1},r_j,m_j,a_j,s_j).
\end{eqnarray*}
The second equation holds by substituting $p^{\pi_e}(s_j)$ with $\frac{p^{\pi_e}(s_j)}{p^{\pi_b}(s_j)}p^{\pi_b}(s_j)=\omega^{\pi_e}(s_j)p^{\pi_b}(s_j)$ and $\pi_{e}(a_j|s_j)$ with $\frac{\pi_{e}(a_j|s_j)}{\pi_{b,\theta_0}(a_j|s_j)}\pi_{b,\theta_0}(a_j|s_j)$.
The last equation holds, using the definition of $Q^{\pi_e}(s,a,m)$,
\begin{eqnarray*}
	 \lim_{T\to \infty}\frac{1}{T}\sum_{j=0}^{T-1} \sum_{\tau_{j-1},a_{j},s_{j}} 	\frac{\pi_{e}(a_j|s_j)p^{\pi_e}(s_j) }{\pi_{b,\theta_0}(a_j|s_j)p^{\pi_b}(s_j)} \pi_{b,\theta_0}(a_j|s_j)p^{\pi_b}(s_j)
	 \times \{\triangledown_{\theta} \log \pi_{b,\theta_0}(a_j|s_j) + \log p^{\pi_b}(s_j)\}\\
	 \times
	 \sum_{s_{j+1},r_{j},m_j}[r_j-\eta^{\pi_e} + \Mean^{\pi_e}_{a^*,m^*} Q^{\pi_e}(s_{j+1},a^*,m^*) - \Mean_{m^*} Q^{\pi_e}(s_{j},a_j,m^*)] p_{\theta_0}(s_{j+1},r_j,m_j|a_j,s_j)=0
\end{eqnarray*}
Therefore, implementing the fact that the expectation of a score function is zero and utilizing the Markov property, we obtain that, 
\begin{eqnarray*}
     D_1 = \Mean\left[\omega^{\pi_e}(S)\frac{\pi_{e}(A|S)}{\pi_{b,\theta_0}(A|S)}\{R + \Mean^{\pi_e}_{a,m} Q^{\pi_e}(S',a,m) - \Mean_{m} Q^{\pi_e}(S,A,m) -\eta^{\pi_e} \}S(\bar{O}_{T-1})\right].
\end{eqnarray*}
Since $(S, A, M, R, S')$ is any arbitrary transaction tuple follows the corresponding distribution, we have that
\begin{eqnarray*}
     D_1 = \Mean\left[\frac{1}{T}\sum_{t=0}^{T-1}\omega^{\pi_e}(S_t)\frac{\pi_{e}(A_t|S_t)}{\pi_{b,\theta_0}(A_t|S_t)}\{R_t + \Mean^{\pi_e}_{a,m} Q^{\pi_e}(S_{t+1},a,m) - \Mean_{m} Q^{\pi_e}(S_t,A_t,m) -\eta^{\pi_e} \}S(\bar{O}_{T-1})\right].
\end{eqnarray*}

\subsubsection{$D_2$} \label{Derivation:I2}
Finally, we focus on the derivation of $D_2$. Note that in equation (\ref{I2:original}), $D_2$ can be divided into two parts, where
\begin{multline*}
	 D_2^{(1)} = \lim_{T\to \infty}\frac{1}{T}\sum_{t=0}^{T-1} \sum_{\tau_t}r_t p_{\theta_0}(s_{t+1},r_t|s_t,a_t,m_t)\pi_0(a_t|s_t)\sum_{a'}p_{\theta_0}(m_t|s_t,a')\pi_e(a'|s_t)\\
	 \times \prod_{j=0}^{t-1}p_{\theta_0}^{\pi_e}(s_{j+1},r_j,m_j,a_j|s_j) \triangledown_{\theta} \log p_{\theta_0}^{\pi_e}(s_{t+1},r_t|m_t,s_t,a_t)\times \nu_{\theta_0}(s_0),
\end{multline*}
and
\begin{multline*}
    D_2^{(2)} = \lim_{T\to \infty}\frac{1}{T}\sum_{t=0}^{T-1} \sum_{a_t,m_t,\tau_{t-1}}p_{\theta_0}(m_t|s_t,a_t)\pi_e(a_t|s_t)\sum_{s_{t+1},r_t,a'}r_t p_{\theta_0}(s_{t+1},r_t|s_t,a',m_t)\pi_0(a'|s_t)\\\times\prod_{j=0}^{t-1}p_{\theta_0}^{\pi_e}(s_{j+1},r_j,m_j,a_j|s_j) \Big\{\triangledown_{\theta} \log p_{\theta_0}(m_t|s_t, a_t)+\sum_{j=0}^{t-1}\left[ \triangledown_{\theta} \log p_{\theta_0}^{\pi_e}(s_{j+1},r_j,m_j|s_j,a_j)]\right.\Big\} \nu_{\theta_0}(s_0),
\end{multline*} note that here we switch the summation of $a$ and $a'$ and change the subscript of the summation accordingly.

\textbf{Part I ($D_2^{(1)}$).}
Using the fact that the expectation of a score function is zero, we first obtain that 
\begin{align*}
    &\begin{aligned}\lim_{T\to \infty}\frac{1}{T}\sum_{t=0}^{T-1} \sum_{\tau_t}r_{\theta_0}(s_t,a_t,m_t)p_{\theta_0}(s_{t+1},r_t|s_t,a_t,m_t)\pi_0(a_t|s_t)\sum_{a'}p_{\theta_0}(m_t|s_t,a')\pi_e(a'|s_t)\\
	 \times \prod_{j=0}^{t-1}p_{\theta_0}^{\pi_e}(s_{j+1},r_j,m_j,a_j|s_j) \triangledown_{\theta} \log p_{\theta_0}^{\pi_e}(s_{t+1},r_t|m_t,s_t,a_t)\times \nu_{\theta_0}(s_0) \end{aligned}\\
	 =& \begin{aligned}\lim_{T\to \infty}\frac{1}{T}\sum_{t=0}^{T-1} \sum_{a_t,m_t,\tau_{t-1}}r_{\theta_0}(s_t,a_t,m_t)\pi_0(a_t|s_t)\sum_{a'}p_{\theta_0}(m_t|s_t,a')\pi_e(a'|s_t)\prod_{j=0}^{t-1}p_{\theta_0}^{\pi_e}(s_{j+1},r_j,m_j,a_j|s_j) \\ \times \sum_{s_{t+1},r_t}p_{\theta_0}(s_{t+1},r_t|s_t,a_t,m_t)\triangledown_{\theta} \log p_{\theta_0}^{\pi_e}(s_{t+1},r_t|m_t,s_t,a_t)\times \nu_{\theta_0}(s_0) \end{aligned}\\
	 =& 0
\end{align*}
Therefore, it follows that
\begin{multline*}
    D_2^{(1)} = \lim_{T\to \infty}\frac{1}{T}\sum_{t=0}^{T-1} \sum_{\tau_t}[r_t-r_{\theta_0}(s_t,a_t,m_t)]p_{\theta_0}(s_{t+1},r_t|s_t,a_t,m_t)\pi_0(a_t|s_t)\sum_{a'}p_{\theta_0}(m_t|s_t,a')\pi_e(a'|s_t)\\
	 \times \prod_{j=0}^{t-1}p_{\theta_0}^{\pi_e}(s_{j+1},r_j,m_j,a_j|s_j) \triangledown_{\theta} \log p_{\theta_0}^{\pi_e}(s_{t+1},r_t|m_t,s_t,a_t)\times \nu_{\theta_0}(s_0).
\end{multline*}
Furthermore, since
\begin{multline*}
    \lim_{T\to \infty}\frac{1}{T}\sum_{t=0}^{T-1} \sum_{a_t,m_t,\tau_{t-1}}[\sum_{s_{t+1},r_t}r_t p_{\theta_0}(s_{t+1},r_t|s_t,a_t,m_t)-r_{\theta_0}(s_t,a_t,m_t)]\pi_0(a_t|s_t)\sum_{a'}p_{\theta_0}(m_t|s_t,a')\pi_e(a'|s_t)\\
	 \times \prod_{j=0}^{t-1}p_{\theta_0}^{\pi_e}(s_{j+1},r_j,m_j,a_j|s_j)  \sum_{j=0}^{t-1}\triangledown_{\theta}\log p_{\theta_0}^{\pi_e}(s_{j+1},r_j|m_j,s_j,a_j)\times \nu_{\theta_0}(s_0) = 0, 
\end{multline*} $D_2^{(1)}$ can be further written as
\begin{multline*}
     D_2^{(1)} = \lim_{T\to \infty}\frac{1}{T}\sum_{t=0}^{T-1} \sum_{\tau_t}[r_t-r_{\theta_0}(s_t,a_t,m_t)]p_{\theta_0}(s_{t+1},r_t|s_t,a_t,m_t)\pi_0(a_t|s_t)\sum_{a'}p_{\theta_0}(m_t|s_t,a')\pi_e(a'|s_t)\\
	 \times \prod_{j=0}^{t-1}p_{\theta_0}^{\pi_e}(s_{j+1},r_j,m_j,a_j|s_j)  \sum_{j=0}^{t}\triangledown_{\theta}\log p_{\theta_0}^{\pi_e}(s_{j+1},r_j|m_j,s_j,a_j)\times \nu_{\theta_0}(s_0). 
\end{multline*}
Then, following the same steps in the proof of the equality $D_1$, we obtain that
\begin{multline*}
    D_2^{(1)} \stackrel{\star}{=} \lim_{T\to \infty}\frac{1}{T}\sum_{j=0}^{T-1} \sum_{\tau_j}[r_j-r_{\theta_0}(s_j,a_j,m_j)]p_{\theta_0}(s_{j+1},r_j|s_j,a_j,m_j)\pi_0(a_j|s_j)\sum_{a'}p_{\theta_0}(m_j|s_j,a')\pi_e(a'|s_j)\\
	 \times \prod_{k=0}^{j-1}\left. p_{\theta_0}^{\pi_{e}}(s_{k+1},r_k,m_k,a_k|s_k)\right.\nu_{\theta_0}(s_0)\triangledown_{\theta} \log p_{\theta_0}(s_{j+1},r_j|s_j,a_j,m_j).
\end{multline*}
Similarly, using the equality $\star\star$,
\begin{multline*}
    D_2^{(1)} \stackrel{\star\star}{=} \lim_{T\to \infty}\frac{1}{T}\sum_{j=0}^{T-1} \sum_{s_j,a_j,m_j,r_j,s_{j+1}}[r_j-r_{\theta_0}(s_j,a_j,m_j)]p_{\theta_0}(s_{j+1},r_j|s_j,a_j,m_j) \pi_0(a_j|s_j)\\
    \times \sum_{a'}p_{\theta_0}(m_j|s_j,a')\pi_e(a'|s_j)p^{\pi_{e}}(s_{j})\triangledown_{\theta} \log p_{\theta_0}(s_{j+1},r_j|s_j,a_j,m_j).
\end{multline*}
Replacing $\pi_0(a_j|s_j)$ with $\frac{\pi_0(a_j|s_j)}{\pi_{b,\theta_0}(a_j|s_j)}\pi_{b,\theta_0}(a_j|s_j)$, $p^{\pi_{e}}(s_{j})$ with $\frac{p^{\pi_{e}}(s_{j})}{p^{\pi_{b}}(s_{j})}p^{\pi_{b}}(s_{j}) = \omega^{\pi_e}(s_j)p^{\pi_{b}}(s_{j})$, and $\sum_{a'}p_{\theta_0}(m_j|s_j,a')\pi_e(a'|s_j)$ with $\frac{\sum_{a'}p_{\theta_0}(m_j|s_j,a')\pi_e(a'|s_j)}{p_{\theta_0}(m_j|s_j,a_j)}p_{\theta_0}(m_j|s_j,a_j)$, we obtain that
\begin{multline*}
    D_2^{(1)} = \lim_{T\to \infty}\frac{1}{T}\sum_{j=0}^{T-1} \sum_{s_j,a_j,m_j,r_j,s_{j+1}}\omega^{\pi_e}(s_j)\frac{\sum_{a'}p_{\theta_0}(m_j|s_j,a')\pi_e(a'|s_j)}{p_{\theta_0}(m_j|s_j,a_j)}\frac{\pi_0(a_j|s_j)}{\pi_{b,\theta_0}(a_j|s_j)}[r_j-r_{\theta_0}(s_j,a_j,m_j)]\\
    \times p_{\theta_0}(s_{j+1},r_j|s_j,a_j,m_j)p_{\theta_0}(m_j|s_j,a_j)\pi_{b,\theta_0}(a_j|s_j)p^{\pi_{b}}(s_{j})\triangledown_{\theta} \log p_{\theta_0}(s_{j+1},r_j|s_j,a_j,m_j).
\end{multline*}
Further, since
\begin{multline*}
    \lim_{T\to \infty}\frac{1}{T}\sum_{j=0}^{T-1} \sum_{s_j,a_j,m_j}\omega^{\pi_e}(s_j)\frac{\sum_{a'}p_{\theta_0}(m_j|s_j,a')\pi_e(a'|s_j)}{p_{\theta_0}(m_j|s_j,a_j)}\frac{\pi_0(a_j|s_j)}{\pi_{b,\theta_0}(a_j|s_j)}p_{\theta_0}(m_j|s_j,a_j)\pi_{b,\theta_0}(a_j|s_j)p^{\pi_{b}}(s_{j})\\
    \times\triangledown_{\theta} \log p_{\theta_0}^{\pi_b}(m_j,a_j,s_j)\sum_{r_j,s_{j+1}}[r_j-r_{\theta_0}(s_j,a_j,m_j)]p_{\theta_0}(s_{j+1},r_j|s_j,a_j,m_j)=0,
\end{multline*} we have that 
\begin{multline*}
    D_2^{(1)} = \lim_{T\to \infty}\frac{1}{T}\sum_{j=0}^{T-1} \sum_{s_j,a_j,m_j,r_j,s_{j+1}}\omega^{\pi_e}(s_j)\frac{\sum_{a'}p_{\theta_0}(m_j|s_j,a')\pi_e(a'|s_j)}{p_{\theta_0}(m_j|s_j,a_j)}\frac{\pi_0(a_j|s_j)}{\pi_{b,\theta_0}(a_j|s_j)}[r_j-r_{\theta_0}(s_j,a_j,m_j)]\\
    \times p_{\theta_0}(s_{j+1},r_j|s_j,a_j,m_j)p_{\theta_0}(m_j|s_j,a_j)\pi_{b,\theta_0}(a_j|s_j)p^{\pi_{b}}(s_{j})\triangledown_{\theta} \log p_{\theta_0}^{\pi_b}(s_{j+1},r_j,m_j,a_j,s_j).   
\end{multline*}
Then, combining the fact that the expectation of a score function is zero and the Markov property, we have that
\begin{align}
    D_2^{(1)} = \Mean\Big[\omega^{\pi_e}(S)\frac{\sum_{a}p_{\theta_0}(M|S,a)\pi_e(a|S)}{p_{\theta_0}(M|S,A)}\frac{\pi_0(A|S)}{\pi_{b,\theta_0}(A|S)}[R-r(S,A,M)] S(\bar{O}_{T-1})\Big]\label{I2_1}.
\end{align}

\textbf{Part II ($D_2^{(2)}$).}
Note that, taking the sum over $r_t$ and $s_{t+1}$, $D_2^{(2)}$ can be equally represented as
\begin{multline*}
\lim_{T\to \infty}\frac{1}{T}\sum_{t=0}^{T-1} \sum_{a',a_t,m_t,\tau_{t-1}}r_{\theta_0}(s_t,a',m_t)\pi_0(a'|s_t)p_{\theta_0}(m_t|s_t,a_t)\pi_e(a_t|s_t)\prod_{j=0}^{t-1}p_{\theta_0}^{\pi_e}(s_{j+1},r_j,m_j,a_j|s_j)\\ \times \Big\{\triangledown_{\theta} \log p_{\theta_0}(m_t|s_t, a_t)+\sum_{j=0}^{t-1}\left[ \triangledown_{\theta} \log p_{\theta_0}^{\pi_e}(s_{j+1},r_j,m_j|s_j,a_j)]\right.	 \Big\}\times \nu_{\theta_0}(s_0).
\end{multline*}
Taking the average over $r_t$ and $s_{t+1}$, and noticing that
\begin{multline*}
\lim_{T\to \infty}\frac{1}{T}\sum_{t=0}^{T-1} \sum_{a',a_t,m_t,\tau_{t-1}}r_{\theta_0}(s_t,a',m_t)\pi_0(a'|s_t)p_{\theta_0}(m_t|s_t,a_t)\pi_e(a_t|s_t)\prod_{j=0}^{t-1}p_{\theta_0}^{\pi_e}(s_{j+1},r_j,m_j,a_j|s_j)\\ \times \sum_{s_{t+1},r_t}p_{\theta_0}(s_{t+1},r_t|s_t,a_t,m_t) \triangledown_{\theta} \log p_{\theta_0}(s_{t+1},r_t|s_t,a_t,m_t) \nu_{\theta_0}(s_0)=0,
\end{multline*} we can rewrite the $D_2^{(2)}$ as
\begin{multline*}
    \lim_{T\to \infty}\frac{1}{T}\sum_{t=0}^{T-1} \sum_{a',\tau_t}r_{\theta_0}(s_t,a',m_t)\pi_0(a'|s_t)\prod_{j=0}^{t}p_{\theta_0}^{\pi_e}(s_{j+1},r_j,m_j,a_j|s_j)\nu_{\theta_0}(s_0)\sum_{j=0}^{t}\triangledown_{\theta} \log p_{\theta_0}^{\pi_e}(s_{j+1},r_j,m_j|s_j,a_j).
\end{multline*}
Then, following the steps as we did in deriving $D_1$, we first show that
\begin{align*}
    \lim_{T\to \infty}\frac{1}{T}\sum_{t=0}^{T-1} \sum_{\tau_t}\eta^{G_e}\prod_{j=0}^{t}p_{\theta_0}^{\pi_e}(s_{j+1},r_j,m_j,a_j|s_j)\nu_{\theta_0}(s_0)\sum_{j=0}^{t}\triangledown_{\theta} \log p_{\theta_0}^{\pi_e}(s_{j+1},r_j,m_j|s_j,a_j)=0,
\end{align*}which follows that
\begin{multline*}
    D_2^{(2)} = \lim_{T\to \infty}\frac{1}{T}\sum_{t=0}^{T-1} \sum_{a',\tau_t}[r_{\theta_0}(s_t,a',m_t)\pi_0(a'|s_t)-\eta^{G_e}]\prod_{j=0}^{t}p_{\theta_0}^{\pi_e}(s_{j+1},r_j,m_j,a_j|s_j)\nu_{\theta_0}(s_0)\\\sum_{j=0}^{t}\triangledown_{\theta} \log p_{\theta_0}^{\pi_e}(s_{j+1},r_j,m_j|s_j,a_j).
\end{multline*}
Next, similarly, given the definition of $Q^{G_e}$, following the steps in deriving $D_1$ and combining with the trick of score function together with the Markov property, we obtain that 
\begin{align}\label{I2_2}
    D_2^{(2)} = \Mean[\omega^{\pi_e}(S)\frac{\pi_{e}(A|S)}{\pi_{b,\theta_0}(A|S)}\{\sum_{a'}r_{\theta_0}(S,a',M)\pi_0(a'|S)-\eta^{G_e} + \Mean_{a,m}^{\pi_e} Q^{G_e}(S',a,m) - \Mean_{m} Q^{G_e}(S,A,m)\}S(\bar{O}_{T-1})].
\end{align}
Combining equation(\ref{I2_1}) and equation(\ref{I2_2}), we have that
\begin{multline*}
    D_2 = \Mean\Big[\omega^{\pi_e}(S)\Big\{\frac{\sum_{a}p_{\theta_0}(M|S,a)\pi_e(a|S)}{p_{\theta_0}(M|S,A)}\frac{\pi_0(A|S)}{\pi_{b,\theta_0}(A|S)}[R-r_{\theta_0}(S,A,M)]+\frac{\pi_{e}(A|S)}{\pi_{b,\theta_0}(A|S)}\\
    \times \{\sum_{a'}r_{\theta_0}(S,a',M)\pi_0(a'|S)-\eta^{G_e} + \Mean_{a,m}^{\pi_e} Q^{G_e}(S',a,m) - \Mean_{m} Q^{G_e}(S,A,m)\}\Big\}S(\bar{O}_{T-1})\Big].
\end{multline*}

Since $(S, A, M, R, S')$ is any arbitrary transaction tuple follows the corresponding distribution, we have that
\begin{multline*}
    D_2 = \Mean\Big[\frac{1}{T}\sum_{t=0}^{T-1}\omega^{\pi_e}(S_t)\Big\{\frac{\sum_{a}p_{\theta_0}(M_t|S_t,a)\pi_e(a|S_t)}{p_{\theta_0}(M_t|S_t,A_t)}\frac{\pi_0(A_t|S_t)}{\pi_{b,\theta_0}(A_t|S_t)}[R_t-r_{\theta_0}(S_t,A_t,M_t)]+\frac{\pi_{e}(A_t|S_t)}{\pi_{b,\theta_0}(A_t|S_t)}\\
    \times \{\sum_{a'}r_{\theta_0}(S_t,a',M_t)\pi_0(a'|S_t)-\eta^{G_e} + \Mean_{a,m}^{\pi_e} Q^{G_e}(S_{t+1},a,m) - \Mean_{m} Q^{G_e}(S_t,A_t,m)\}\Big\}S(\bar{O}_{T-1})\Big].
\end{multline*}

\subsubsection{Derivative of $\textrm{IDE}_{\theta_0}(\pi_e,\pi_0)$}
Given $C_1$, $D_1$, and $D_2$, the derivative of $\textrm{IDE}_{\theta_0}(\pi_e,\pi_0)$ is $\eta^{\pi_e}-\eta^{G_e}+I_1-I_2$, where
\begin{eqnarray*}
     I_1 = \Mean[\omega^{\pi_e}(S)\frac{\pi_{e}(A|S)}{\pi_{b,\theta_0}(A|S)}\{R-\eta^{\pi_e} + \Mean^{\pi_e}_{a,m} Q^{\pi_e}(S',a,m) - \Mean_{m} Q^{\pi_e}(S,A,m)\}],
\end{eqnarray*}and
\begin{multline*}
    I_2 = \Mean\Big[\omega^{\pi_e}(S)\Big\{\frac{\sum_{a}p_{\theta_0}(M|S,a)\pi_e(a|S)}{p_{\theta_0}(M|S,A)}\frac{\pi_0(A|S)}{\pi_{b,\theta_0}(A|S)}[R-r_{\theta_0}(S,A,M)]+\frac{\pi_{e}(A|S)}{\pi_{b,\theta_0}(A|S)}\\
    \times \{\sum_{a'}r_{\theta_0}(S,a',M)\pi_0(a'|S) + \Mean_{a,m}^{\pi_e} Q^{G_e}(S',a,m) - \Mean_{m} Q^{G_e}(S,A,m)-\eta^{G_e}\}\Big\}\Big].
\end{multline*}

\subsection{EIF for Immediate Mediator Effect}
Immediate Mediator Effect (IME) can be represented as
\begin{multline}\label{ME}
	\lim_{T\to \infty}\frac{1}{T}\sum_{t=0}^{T-1} \sum_{\tau_t} r_t p_{\theta_0}(s_{t+1},r_t|s_t,a_t,m_t)\pi_0(a_t|s_t)[\sum_{a'}p_{\theta_0}(m_t|s_t,a')\pi_{e}(a'|s_t)-p_{\theta_0}(m_t|a_t,s_t)]\\
	\times \prod_{j=0}^{t-1}\left[p_{\theta_0}^{\pi_e}(s_{j+1},r_j,m_j,a_j|s_j)\right]\nu_{\theta_0}(s_0).
\end{multline}
Taking the derivative of $\textrm{IME}_{\theta_0}(\pi_e,\pi_0)$, we get that 
\begin{eqnarray*}
	\frac{\partial \textrm{IME}_{\theta_0}(\pi_e,\pi_0)}{\partial \theta_{0}} = C_2 + D_2 - D_3,
\end{eqnarray*} where
\begin{eqnarray*}
    C_2 = (\ref{ME})\times \triangledown_{\theta} \log(\nu_{\theta_0}(s_0))= \Mean[\textrm{IME}_{\theta_0}(\pi_e,\pi_0) \times S(\bar{O}_{T-1})] = \Mean[(\eta^{G_e}-\eta^{\pi_{e,0}}) \times S(\bar{O}_{T-1})],
\end{eqnarray*}
$D_2$ is derived in Appendix \ref{Derivation:I2}, and 
\begin{multline*}
    D_3 = \lim_{T\to \infty}\frac{1}{T}\sum_{t=0}^{T-1} \sum_{\tau_t} r_t p_{\theta_0}(s_{t+1},r_t|s_t,a_t,m_t)p_{\theta_0}(m_t|s_t,a_t)\pi_0(a_t|s_t)\prod_{j=0}^{t-1}\left[p_{\theta_0}^{\pi_e}(s_{j+1},r_j,m_j,a_j|s_j)\right]\\
    \times \Big[\sum_{j=0}^{t-1}[\triangledown_{\theta} \log p_{\theta_0}(s_{j+1},r_j,m_j|a_j,s_j)]+\triangledown_{\theta} \log p_{\theta_0}(s_{t+1},r_t,m_t|s_t,a_t)\Big]\nu_{\theta_0}(s_0),
\end{multline*} which can be represented as the sum of two parts. Specifically, $D_3 = D_3^{(1)}+D_3^{(2)}$, where
\begin{multline*}
    D_3^{(1)} = \lim_{T\to \infty}\frac{1}{T}\sum_{t=0}^{T-1} \sum_{\tau_t} r_t p_{\theta_0}(s_{t+1},r_t|s_t,a_t,m_t)p_{\theta_0}(m_t|s_t,a_t)\pi_0(a_t|s_t)\prod_{j=0}^{t-1}\left[p_{\theta_0}^{\pi_e}(s_{j+1},r_j,m_j,a_j|s_j)\right]\\
    \times \triangledown_{\theta} \log p_{\theta_0}(s_{t+1},r_t,m_t|s_t,a_t)\nu_{\theta_0}(s_0),
\end{multline*}
and 
\begin{multline*}
    D_3^{(2)} = \lim_{T\to \infty}\frac{1}{T}\sum_{t=0}^{T-1} \sum_{\tau_t} r_t p_{\theta_0}(s_{t+1},r_t|s_t,a_t,m_t)p_{\theta_0}(m_t|s_t,a_t)\pi_0(a_t|s_t)\prod_{j=0}^{t-1}\left[p_{\theta_0}^{\pi_e}(s_{j+1},r_j,m_j,a_j|s_j)\right]\\
    \times \sum_{j=0}^{t-1}\triangledown_{\theta} \log p_{\theta_0}(s_{j+1},r_j,m_j|a_j,s_j)\nu_{\theta_0}(s_0).
\end{multline*}
\subsubsection{$D_3$} \label{Derivation:I3}
\textbf{Part I ($D_3^{(1)}$).} First, using the fact that the expectation of a score function is $0$, we notice that,
\begin{align*}
    &\begin{aligned}
    \lim_{T\to \infty}\frac{1}{T}\sum_{t=0}^{T-1} \sum_{\tau_t} \Mean_{m^{*}}r_{\theta_0}(s_t,a_t,m^{*})p_{\theta_0}(s_{t+1},r_t|s_t,a_t,m_t)p_{\theta_0}(m_t|s_t,a_t)\pi_0(a_t|s_t)\\ \times \prod_{j=0}^{t-1}\left[p_{\theta_0}^{\pi_e}(s_{j+1},r_j,m_j,a_j|s_j)\right]\triangledown_{\theta} \log p_{\theta_0}(s_{t+1},r_t,m_t|s_t,a_t)\nu_{\theta_0}(s_0),
    \end{aligned}\\
    =&\begin{aligned}
    \lim_{T\to \infty}\frac{1}{T}\sum_{t=0}^{T-1} \sum_{a_t,\tau_{t-1}} \Mean_{m^{*}}r_{\theta_0}(s_t,a_t,m^{*})\pi_0(a_t|s_t)\prod_{j=0}^{t-1}\left[p_{\theta_0}^{\pi_e}(s_{j+1},r_j,m_j,a_j|s_j)\right]\nu_{\theta_0}(s_0)\\
    \times \sum_{s_{t+1},r_t,m_t}p_{\theta_0}(s_{t+1},r_t,m_t|s_t,a_t)\triangledown_{\theta} \log p_{\theta_0}(s_{t+1},r_t,m_t|s_t,a_t),
    \end{aligned}\\
    =&0,
\end{align*} which follows that
\begin{align*}
    D_3^{(1)} &=
    \begin{aligned}
    \lim_{T\to \infty}\frac{1}{T}\sum_{t=0}^{T-1} \sum_{\tau_t} [r_t-\Mean_{m^{*}}r_{\theta_0}(s_t,a_t,m^{*})]p_{\theta_0}(s_{t+1},r_t|s_t,a_t,m_t)p_{\theta_0}(m_t|s_t,a_t)\pi_0(a_t|s_t)\\
    \times \prod_{j=0}^{t-1}\left[p_{\theta_0}^{\pi_e}(s_{j+1},r_j,m_j,a_j|s_j)\right]\nu_{\theta_0}(s_0)
    \triangledown_{\theta} \log p_{\theta_0}(s_{t+1},r_t,m_t|s_t,a_t),
    \end{aligned}\\
    &=\begin{aligned}
    \lim_{T\to \infty}\frac{1}{T}\sum_{t=0}^{T-1} \sum_{s_{t+1},r_t,a_t,m_t,s_t} [r_t-\Mean_{m^{*}}r_{\theta_0}(s_t,a_t,m^{*})]p_{\theta_0}(s_{t+1},r_t,m_t|s_t,a_t)\pi_0(a_t|s_t)\\
    \times p^{\pi_e}(s_t)
    \triangledown_{\theta} \log p_{\theta_0}(s_{t+1},r_t,m_t|s_t,a_t).
    \end{aligned}
\end{align*}
Replacing the $\pi_0(a_t|s_t)$ with $\frac{\pi_0(a_t|s_t)}{\pi_{b,\theta_0}(a_t|s_t)}\pi_{b,\theta_0}(a_t|s_t)$, and $p^{\pi_e}(s_t)$ with $\frac{p^{\pi_e}(s_t)}{p^{\pi_b}(s_t)}p^{\pi_b}(s_t) = \omega^{\pi_e}(s_t)p^{\pi_b}(s_t)$, we obtain that 
\begin{multline*}
    D_3^{(1)} = \lim_{T\to \infty}\frac{1}{T}\sum_{t=0}^{T-1} \sum_{s_{t+1},r_t,a_t,m_t,s_t} \omega^{\pi_e}(s_t)\frac{\pi_0(a_t|s_t)}{\pi_{b,\theta_0}(a_t|s_t)}[r_t-\Mean_{m^{*}}r_{\theta_0}(s_t,a_t,m^{*})]\\ \times p_{\theta_0}(s_{t+1},r_t,m_t|s_t,a_t)\pi_{b,\theta_0}(a_t|s_t)p^{\pi_b}(s_t)\triangledown_{\theta} \log p_{\theta_0}(s_{t+1},r_t,m_t|s_t,a_t).
\end{multline*}
Further, since
\begin{multline*}
    \lim_{T\to \infty}\frac{1}{T}\sum_{t=0}^{T-1} \sum_{a_t,s_t} \omega^{\pi_e}(s_t)\frac{\pi_0(a_t|s_t)}{\pi_{b,\theta_0}(a_t|s_t)}\pi_{b,\theta_0}(a_t|s_t)p^{\pi_b}(s_t)\triangledown_{\theta} \log p_{\theta_0}^{\pi_b}(a_t,s_t)\\ \times \sum_{s_{t+1},r_t, m_t}[r_t-\Mean_{m^{*}}r_{\theta_0}(s_t,a_t,m^{*})] p_{\theta_0}(s_{t+1},r_t,m_t|s_t,a_t)=0,
\end{multline*} we have that
\begin{multline*}
    D_3^{(1)} = \lim_{T\to \infty}\frac{1}{T}\sum_{t=0}^{T-1} \sum_{s_{t+1},r_t,a_t,m_t,s_t} \omega^{\pi_e}(s_t)\frac{\pi_0(a_t|s_t)}{\pi_{b,\theta_0}(a_t|s_t)}[r_t-\Mean_{m^{*}}r_{\theta_0}(s_t,a_t,m^{*})]\\ \times p_{\theta_0}(s_{t+1},r_t,m_t|s_t,a_t)\pi_{b,\theta_0}(a_t|s_t)p^{\pi_b}(s_t)\triangledown_{\theta} \log p_{\theta_0}^{\pi_b}(s_{t+1},r_t,m_t,a_t,s_t).
\end{multline*}
Lastly, combining the fact that the expectation of a score function is $0$ and the Markov property, we finalize the derivation of $D_3^{(1)}$ with 
\begin{align}
    D_3^{(1)} = \Mean\Big[\omega^{\pi_e}(S)\frac{\pi_0(A|S)}{\pi_{b,\theta_0}(A|S)}[R-\Mean_{m}r_{\theta_0}(S,A,m)] S(\bar{O}_{T-1})\Big].\label{D_3_1}
\end{align}

\textbf{Part II ($D_3^{(2)}$).} We first rewrite the $D_3^{(2)}$ as
\begin{multline*}
    \lim_{T\to \infty}\frac{1}{T}\sum_{t=0}^{T-1} \sum_{a_t,m_t,\tau_{t-1}} r_{\theta_0}(s_t,a_t,m_t)p_{\theta_0}(m_t|s_t,a_t)\pi_0(a_t|s_t)\prod_{j=0}^{t-1}\left[p_{\theta_0}^{\pi_e}(s_{j+1},r_j,m_j,a_j|s_j)\right]\\
    \times \sum_{j=0}^{t-1}\triangledown_{\theta} \log p_{\theta_0}(s_{j+1},r_j,m_j|a_j,s_j)\nu_{\theta_0}(s_0).
\end{multline*} Taking the additional average over $s^{*}$, $r^{*}$, $m^{*}$, and $a^{*}$, and noticing that
\begin{multline*}
    \lim_{T\to \infty}\frac{1}{T}\sum_{t=0}^{T-1} \sum_{a_t,m_t,\tau_{t-1}} r_{\theta_0}(s_t,a_t,m_t)p_{\theta_0}(m_t|s_t,a_t)\pi_0(a_t|s_t)\prod_{j=0}^{t-1}\left[p_{\theta_0}^{\pi_e}(s_{j+1},r_j,m_j,a_j|s_j)\right]\\
    \times \sum_{s^{*},r^{*},m^{*},a^{*}}p_{\theta_0}(s^{*},r^{*},m^{*}|s_t,a^*)\pi_e(a^*|s_t)\triangledown_{\theta} \log p_{\theta_0}(s^{*},r^{*},m^{*}|s_t,a^*)\nu_{\theta_0}(s_0) =0,
\end{multline*} we further represent $D_3^{(2)}$ as
\begin{multline*}
    \lim_{T\to \infty}\frac{1}{T}\sum_{t=0}^{T-1} \sum_{\tau_t} \sum_{m^{*},a^*}p_{\theta_0}(m^{*}|s_t,a^*)r_{\theta_0}(s_t,a^*,m^{*})\pi_0(a^*|s_t)\prod_{j=0}^{t}\left[p_{\theta_0}^{\pi_e}(s_{j+1},r_j,m_j,a_j|s_j)\right]\\
    \times \sum_{j=0}^{t}\triangledown_{\theta} \log p_{\theta_0}(s_{j+1},r_j,m_j|a_j,s_j)\nu_{\theta_0}(s_0).
\end{multline*} Note that we change the subscript of the summations accordingly. Then, following the steps we processed to derive the $D_1$, we first show that 
\begin{align*}
    \lim_{T\to \infty}\frac{1}{T}\sum_{t=0}^{T-1} \sum_{\tau_t}\eta^{\pi_{e,0}}\prod_{j=0}^{t}\left[p_{\theta_0}^{\pi_e}(s_{j+1},r_j,m_j,a_j|s_j)\right]\sum_{j=0}^{t}\triangledown_{\theta} \log p_{\theta_0}(s_{j+1},r_j,m_j|a_j,s_j)\nu_{\theta_0}(s_0)=0.
\end{align*}
Therefore, 
\begin{multline*}
    D_3^{(2)} = \lim_{T\to \infty}\frac{1}{T}\sum_{t=0}^{T-1} \sum_{\tau_t} \Big[\sum_{m^{*},a^*}r_{\theta_0}(s_t,a^*,m^{*})p_{\theta_0}(m^{*}|s_t,a^*)\pi_0(a^*|s_t)-\eta^{\pi_{e,0}}\Big]\prod_{j=0}^{t}\left[p_{\theta_0}^{\pi_e}(s_{j+1},r_j,m_j,a_j|s_j)\right]\\
    \times \sum_{j=0}^{t}\triangledown_{\theta} \log p_{\theta_0}(s_{j+1},r_j,m_j|a_j,s_j)\nu_{\theta_0}(s_0).
\end{multline*}
Next, using the equality properties $\star$ and $\star\star$, together with the definition of $Q^{\pi_{e,0}}(s,a,m)$ and the trick of score functions, we can show that
\begin{align*}
    D_3^{(2)} = \lim_{T\to \infty}\frac{1}{T}\sum_{j=0}^{T-1} \sum_{\tau_{j}} &\Big[\sum_{m^{*},a^*}r_{\theta_0}(s_t,a^*,m^{*})p_{\theta_0}(m^{*}|s_t,a^*)\pi_0(a^*|s_t)-\eta^{\pi_{e,0}} + \Mean^{\pi_e}_{a^*,m^*} Q^{\pi_{e,0}}(s_{j+1},a^*,m^*) \\
    &- \Mean_{m^*} Q^{\pi_{e,0}}(s_{j},a_j,m^*)\Big]
	 \frac{\pi_{e}(a_j|s_j)p^{\pi_e}(s_j) }{\pi_{b,\theta_0}(a_j|s_j)p^{\pi_b}(s_j)} p_{\theta_0}(s_{j+1},r_j,m_j|a_j,s_j)\\
	 &\times \pi_{b,\theta_0}(a_j|s_j)p^{\pi_b}(s_j) \triangledown_{\theta} \log p_{\theta_0}^{\pi_b}(s_{j+1},r_j,m_j,a_j,s_j).
\end{align*}
Implementing the fact that the expectation of a score function is zero and utilizing the Markov property, we finally obtain that, 
\begin{multline}\label{D_3_2}
     D_3^{(2)} =
         \Mean\Big[\omega^{\pi_e}(S)\frac{\pi_e(A|S)}{\pi_{b,\theta_0}(A|S)}\{\sum_{a'}\Mean_{m\sim p_{\theta_0}(\bullet|S,a')}r_{\theta_0}(S,a',m)\pi_0(a'|S)\\+\Mean^{\pi_e}_{a,m} Q^{\pi_{e,0}}(S',a,m)
	-\Mean_{m} Q^{\pi_{e,0}}(S,A,m)-\eta^{\pi_{e,0}}\}S(\bar{O}_{T-1})\Big].
\end{multline}
Combining equation (\ref{D_3_1}) and equation (\ref{D_3_2}), we have that
\begin{multline*}
     D_3 = \Mean\Big[\omega^{\pi_e}(S)\Big\{\frac{\pi_0(A|S)}{\pi_{b,\theta_0}(A|S)}[R-\Mean_{m}r_{\theta_0}(S,A,m)]+\frac{\pi_e(A|S)}{\pi_{b,\theta_0}(A|S)}\{\sum_{a'}\Mean_{m\sim p_{\theta_0}(\bullet|S,a')}r_{\theta_0}(S,a',m)\pi_0(a'|S)\\+\Mean^{\pi_e}_{a,m} Q^{\pi_{e,0}}(S',a,m)
	-\Mean_{m} Q^{\pi_{e,0}}(S,A,m)-\eta^{\pi_{e,0}}\}\Big\}S(\bar{O}_{T-1})\Big].
\end{multline*}

Since $(S, A, M, R, S')$ is any arbitrary transaction tuple follows the corresponding distribution, we have that
\begin{multline*}
     D_3 = \Mean\Big[\frac{1}{T}\sum_{t=0}^{T-1}\omega^{\pi_e}(S_t)\Big\{\frac{\pi_0(A_t|S_t)}{\pi_{b,\theta_0}(A_t|S_t)}[R_t-\Mean_{m}r_{\theta_0}(S_t,A_t,m)]+\frac{\pi_e(A_t|S_t)}{\pi_{b,\theta_0}(A_t|S_t)}\{\sum_{a'}\Mean_{m\sim p(\bullet|S_t,a')}r_{\theta_0}(S_t,a',m)\pi_0(a'|S_t)\\+\Mean^{\pi_e}_{a,m} Q^{\pi_{e,0}}(S_{t+1},a,m)
	-\Mean_{m} Q^{\pi_{e,0}}(S_t,A_t,m)-\eta^{\pi_{e,0}}\}\Big\}S(\bar{O}_{T-1})\Big].
\end{multline*}

\subsubsection{Efficient Function}
Given $C_2$, $D_2$, and $D_3$, the efficient influence function for $\textrm{IME}_{\theta_0}(\pi_e,\pi_0)$ is $\eta^{G_e}-\eta^{\pi_{e,0}}+I_2-I_3$, where
\begin{multline*}
    I_2 = \Mean\Big[\omega^{\pi_e}(S)\Big\{\frac{\sum_{a}p_{\theta_0}(M|S,a)\pi_e(a|S)}{p_{\theta_0}(M|S,A)}\frac{\pi_0(A|S)}{\pi_{b,\theta_0}(A|S)}[R-r_{\theta_0}(S,A,M)]+\frac{\pi_{e}(A|S)}{\pi_{b,\theta_0}(A|S)}\\
    \times \{\sum_{a'}r_{\theta_0}(S,a',M)\pi_0(a'|S) + \Mean_{a,m}^{\pi_e} Q^{G_e}(S',a,m) - \Mean_{m} Q^{G_e}(S,A,m)-\eta^{G_e}\}\Big\}\Big],
\end{multline*}and
\begin{multline*}
     I_3 = \Mean\Big[\omega^{\pi_e}(S)\Big\{\frac{\pi_0(A|S)}{\pi_{b,\theta_0}(A|S)}[R-\Mean_{m}r_{\theta_0}(S,A,m)]+\frac{\pi_e(A|S)}{\pi_{b,\theta_0}(A|S)}\{\sum_{a'}\Mean_{m\sim p_{\theta_0}(\bullet|S,a')}r_{\theta_0}(S,a',m)\pi_0(a'|S)\\+\Mean^{\pi_e}_{a,m} Q^{\pi_{e,0}}(S',a,m)
	-\Mean_{m} Q^{\pi_{e,0}}(S,A,m)-\eta^{\pi_{e,0}}\}\Big\}\Big].
\end{multline*}

\input{Appendix/DDE_EIF.tex}

\subsection{Proof of the Equality $\star$} \label{proof:star_equality}

The equality can be proved with the following three steps:

    \textbf{Step 1.} We first exchange the summation of $t$ and $j$ in the first line of the equation $D_1$, which yields that
    \begin{multline}
        D_1 = \lim_{T\to \infty}\frac{1}{T}\sum_{j=0}^{T-1}\sum_{t=j}^{T-1} \sum_{\tau_t} [r_t-\eta^{\pi_e}] \prod_{k=0}^{t}\left. p_{\theta_0}^{\pi_{e}}(s_{k+1},r_k,m_k,a_k|s_k)\right.\\
        \times \left[ \triangledown_{\theta} \log p_{\theta_0}^{\pi_e}(s_{j+1},r_j,m_j,a_j|s_j)]\right.\times \nu_{\theta_0}(s_0).
    \end{multline}
    
    \textbf{Step 2.} Then we split the summation $\sum_{t=j}^{T-1}$ into $t=j$ and $\sum_{t=j+1}^{T-1}$, and split the product $\prod_{k=0}^{t}$ into $\prod_{k=0}^{j}$ and $\prod_{k=j+1}^{t}$, which leads to 
    \begin{multline}
        D_1 = \lim_{T\to \infty}\frac{1}{T}\sum_{j=0}^{T-1}\Big\{\sum_{\tau_j} [r_j-\eta^{\pi_e}] + \sum_{t=j+1}^{T-1} \sum_{\tau_t} [r_t-\eta^{\pi_e}] \prod_{k=j+1}^{t} p_{\theta_0}^{\pi_{e}}(s_{k+1},r_k,m_k,a_k|s_k)\Big\}\\
        \times \prod_{k=0}^{j}\left. p_{\theta_0}^{\pi_{e}}(s_{k+1},r_k,m_k,a_k|s_k)\right. \left[ \triangledown_{\theta} \log p_{\theta_0}^{\pi_e}(s_{j+1},r_j,m_j,a_j|s_j)]\right.\times \nu_{\theta_0}(s_0).
    \end{multline}
    
    \textbf{Step 3.} By the definition of $Q^{\pi_e}$ (see equation (\ref{Q:pi_e})), we have that
    \begin{eqnarray*}
    \sum_{t=j+1}^{T-1} \sum_{\tau_t} [r_t-\eta^{\pi_e}] \prod_{k=j+1}^{t} p_{\theta_0}^{\pi_{e}}(s_{k+1},r_k,m_k,a_k|s_k)\\
    =\Mean^{\pi_e}_{m_{j+1}, a_{j+1}} Q^{\pi_e}(s_{j+1}, a_{j+1},m_{j+1})\\
    =\Mean^{\pi_e}_{a^*,m^*} Q^{\pi_e}(s_{j+1}, a^*,m^*).
    \end{eqnarray*} Substituting this equation, we conclude the proof of $\star$ with that
    \begin{multline*}
        D_1 = \lim_{T\to \infty}\frac{1}{T}\sum_{j=0}^{T-1}\Big\{\sum_{\tau_j} [r_j-\eta^{\pi_e}] + \Mean^{\pi_e}_{a^*,m^*} Q^{\pi_e}(s_{j+1}, a^*,m^*)\Big\}\\
        \times \prod_{k=0}^{j}\left. p_{\theta_0}^{\pi_{e}}(s_{k+1},r_k,m_k,a_k|s_k)\right. \left[ \triangledown_{\theta} \log p_{\theta_0}^{\pi_e}(s_{j+1},r_j,m_j,a_j|s_j)]\right. \nu_{\theta_0}(s_0).
    \end{multline*}

%% file: Appendix/DDE_EIF.tex
\subsection{EIF for Delayed Direct Effect}
Delayed Direct Effect (DDE) can be represented as
\begin{multline}\label{DDE}
    \textrm{DDE}(\pi_e,\pi_0) = \lim_{T\to \infty}\frac{1}{T}\sum_{t=0}^{T-1}\sum_{\tau_t}r_t p(s_{t+1},r_t|s_t,a_t,m_t)p(m_t|s_t,a_t)\pi_0(a_t|s_t)\\
    \times \Big\{\prod_{j=0}^{t-1}p^{\pi_e}(s_{j+1},r_j,m_j,a_j|s_j)
    -\sum_{\bar{a}^*_{t-1}}\prod_{j=0}^{t-1}p(s_{j+1},r_j|s_j,a_j,m_j)\pi_0(a_j|s_j)p(m_j|s_j,a_j^*)\pi_e(a_j^*|s_j)\Big\}\nu(s_0).
\end{multline}
Taking the derivative of $\textrm{DDE}_{\theta_0}(\pi_e,\pi_0)$, we get that 
\begin{eqnarray*}
	\frac{\partial \textrm{DDE}_{\theta_0}(\pi_e,\pi_0)}{\partial \theta_{0}} = C_3 + D_3 - D_4,
\end{eqnarray*} where
\begin{eqnarray*}
    C_3 = (\ref{DDE})\times \triangledown_{\theta} \log(\nu_{\theta_0}(s_0)) = \Mean[\textrm{DDE}_{\theta_0}(\pi_e,\pi_0) \times S(\bar{O}_{T-1})]= \Mean[(\eta^{\pi_{e,0}}-\eta^{G_0}) \times S(\bar{O}_{T-1})],
\end{eqnarray*}
$D_3$ is derived in Appendix \ref{Derivation:I3}, and 
\begin{multline*}
	 D_4 = \lim_{T\to\infty}\frac{1}{T}\sum_{t=0}^{T-1}\sum_{\tau_t,\bar{a}^*_{t-1}}r_t p_{\theta_0}(s_{t+1},r_t,m_t|s_t,a_t)\pi_0(a_t|s_t)\\\prod_{j=0}^{t-1}p_{\theta_0}(s_{j+1},r_j|s_j,a_j,m_j)\pi_0(a_j|s_j)p_{\theta_0}(m_j|s_j,a_j^*)\pi_e(a_j^*|s_j)\nu_{\theta_0}(s_0)\\
    \Big\{ \underbrace{\triangledown_{\theta}\log p_{\theta_0}(s_{t+1},r_{t},m_t|s_t,a_t)}_{D_4^{(1)}}+\underbrace{\sum_{j=0}^{t-1}\triangledown_{\theta}\log p_{\theta_0}(s_{j+1},r_j|s_j,a_j,m_j)}_{D_4^{(2)}} +\underbrace{\sum_{j=0}^{t-1}\triangledown_{\theta}\log p_{\theta_0}^{\pi_e}(m_j|s_j,a^*_j)}_{D_4^{(3)}}\Big\}.
\end{multline*}

\subsubsection{$D_4$}\label{Derivation:I4}


\textbf{Part I ($D_4^{(1)}$).} First, using the fact that the expectation of a score function is $0$, we notice that,
\begin{align*}
    &\begin{aligned}
        \lim_{T\to\infty}\frac{1}{T}\sum_{t=0}^{T-1}\sum_{\tau_t,\bar{a}^*_{t-1}}\Mean_{m'}r_{\theta_0}(s_t,a_t,m') p_{\theta_0}(s_{t+1},r_t,m_t|s_t,a_t)\pi_0(a_t|s_t)\triangledown_{\theta}\log p_{\theta_0}(s_{t+1},r_{t},m_t|s_t,a_t)\\\prod_{j=0}^{t-1}p_{\theta_0}(s_{j+1},r_j|s_j,a_j,m_j)\pi_0(a_j|s_j)p_{\theta_0}(m_j|s_j,a_j^*)\pi_e(a_j^*|s_j)\nu_{\theta_0}(s_0),
    \end{aligned}\\
    =&\begin{aligned}
           \lim_{T\to\infty}\frac{1}{T}\sum_{t=0}^{T-1}\sum_{a_t,\tau_{t-1},\bar{a}^*_{t-1}}\Mean_{m'}r_{\theta_0}(s_t,a_t,m')\pi_0(a_t|s_t)
        \prod_{j=0}^{t-1}p_{\theta_0}(s_{j+1},r_j|s_j,a_j,m_j)\pi_0(a_j|s_j)p_{\theta_0}(m_j|s_j,a_j^*)\pi_e(a_j^*|s_j)\\\sum_{s_{t+1},r_{t},m_{t}}p_{\theta_0}(s_{t+1},r_t,m_t|s_t,a_t)\triangledown_{\theta}\log p_{\theta_0}(s_{t+1},r_{t},m_t|s_t,a_t)\nu_{\theta_0}(s_0),
    \end{aligned}\\
    =&0,
\end{align*} which follows that
\begin{align*}
    D_4^{(1)} &=
    \begin{aligned}
            \lim_{T\to\infty}\frac{1}{T}\sum_{t=0}^{T-1}\sum_{\tau_t,\bar{a}^*_{t-1}}[r_t - \Mean_{m'}r_{\theta_0}(s_t,a_t,m') ]p_{\theta_0}(s_{t+1},r_t,m_t|s_t,a_t)\pi_0(a_t|s_t)\triangledown_{\theta}\log p_{\theta_0}(s_{t+1},r_{t},m_t|s_t,a_t)\\
            \prod_{j=0}^{t-1}p_{\theta_0}(s_{j+1},r_j|s_j,a_j,m_j)\pi_0(a_j|s_j)p_{\theta_0}(m_j|s_j,a_j^*)\pi_e(a_j^*|s_j)
    \nu_{\theta_0}(s_0),
    \end{aligned}\\
    &=\begin{aligned}
            \lim_{T\to\infty}\frac{1}{T}\sum_{t=0}^{T-1}\sum_{s_{t+1},r_{t},m_{t},a_t,s_t}[r_t - \Mean_{m'}r_{\theta_0}(s_t,a_t,m') ]p_{\theta_0}(s_{t+1},r_t,m_t|s_t,a_t)\pi_0(a_t|s_t)p_{\theta_0}^{G}(s_t) \\\triangledown_{\theta}\log p_{\theta_0}(s_{t+1},r_{t},m_t|s_t,a_t).
    \end{aligned}
\end{align*} The last equation holds, since
\begin{align*}
    \sum_{s_0,\tau_{t-1},\bar{a}^*_{t-1}}\prod_{j=0}^{t-1}p_{\theta_0}(s_{j+1},r_j|s_j,a_j,m_j)\pi_0(a_j|s_j)p_{\theta_0}(m_j|s_j,a_j^*)\pi_e(a_j^*|s_j)
    \nu_{\theta_0}(s_0) = p_{\theta_0}^{G}(s_t).
\end{align*}
Replacing the $\pi_0(a_t|s_t)$ with $\frac{\pi_0(a_t|s_t)}{\pi_{b,\theta_0}(a_t|s_t)}\pi_{b,\theta_0}(a_t|s_t)$, and $p_{\theta_0}^{G}(s_t)$ with $\frac{p_{\theta_0}^{G}(s_t)}{p^{\pi_b}(s_t)}p^{\pi_b}(s_t) = \omega_{\theta_0}^{G}(s_t)p^{\pi_b}(s_t)$, we obtain that 
\begin{multline*}
    D_4^{(1)} = \lim_{T\to\infty}\frac{1}{T}\sum_{t=0}^{T-1}\sum_{s_{t+1},r_{t},m_{t},a_t,s_t}\omega_{\theta_0}^{G}(s_t)\frac{\pi_0(a_t|s_t)}{\pi_{b,\theta_0}(a_t|s_t)}[r_t - \Mean_{m'}r_{\theta_0}(s_t,a_t,m') ]\\
    p_{\theta_0}(s_{t+1},r_t,m_t|s_t,a_t)\pi_{b,\theta_0}(a_t|s_t)p^{\pi_b}(s_t) \triangledown_{\theta}\log p_{\theta_0}(s_{t+1},r_{t},m_t|s_t,a_t).
\end{multline*}
Further, since
\begin{multline*}
    \lim_{T\to \infty}\frac{1}{T}\sum_{t=0}^{T-1} \sum_{a_t,s_t} \omega_{\theta_0}^{G}(s_t)\frac{\pi_0(a_t|s_t)}{\pi_{b,\theta_0}(a_t|s_t)}\pi_{b,\theta_0}(a_t|s_t)p^{\pi_b}(s_t)\triangledown_{\theta} \log p_{\theta_0}^{\pi_b}(a_t,s_t)\\ \times \sum_{s_{t+1},r_t, m_t}[r_t-\Mean_{m'}r_{\theta_0}(s_t,a_t,m')] p_{\theta_0}(s_{t+1},r_t,m_t|s_t,a_t)=0,
\end{multline*} we have that
\begin{multline*}
    D_4^{(1)} = \lim_{T\to \infty}\frac{1}{T}\sum_{t=0}^{T-1} \sum_{s_{t+1},r_t,a_t,m_t,s_t} \omega_{\theta_0}^{G}(s_t)\frac{\pi_0(a_t|s_t)}{\pi_{b,\theta_0}(a_t|s_t)}[r_t-\Mean_{m'}r_{\theta_0}(s_t,a_t,m')]\\ \times p_{\theta_0}(s_{t+1},r_t,m_t|s_t,a_t)\pi_{b,\theta_0}(a_t|s_t)p^{\pi_b}(s_t)\triangledown_{\theta} \log p_{\theta_0}^{\pi_b}(s_{t+1},r_t,m_t,a_t,s_t).
\end{multline*}
Lastly, combining the fact that the expectation of a score function is $0$ and the Markov property, we finalize the derivation of $D_4^{(1)}$ with 
\begin{align}
    D_4^{(1)} = \Mean\Big[\omega_{\theta_0}^{G}(S)\frac{\pi_0(A|S)}{\pi_{b,\theta_0}(A|S)}[R-\Mean_{m}r_{\theta_0}(S,A,m)] S(\bar{O}_{T-1})\Big].\label{D_5_1}
\end{align}

\textbf{Part II ($D_4^{(2)}$).} 
Taking the additional average over $s'$, $r'$, $a'$, $m'$, and $\tilde{a}$, and noticing that
\begin{multline*}
    \lim_{T\to \infty}\frac{1}{T}\sum_{t=0}^{T-1} \sum_{a_t,m_t,\tau_{t-1},\bar{a}^*_{t-1}} r_{\theta_0}(s_t,a_t,m_t)p_{\theta_0}(m_t|s_t,a_t)\pi_0(a_t|s_t)\\\times\prod_{j=0}^{t-1}p_{\theta_0}(s_{j+1},r_j|s_j,a_j,m_j)\pi_0(a_j|s_j)p_{\theta_0}(m_j|s_j,a_j^*)\pi_e(a_j^*|s_j)\nu_{\theta_0}(s_0)\\
    \times \sum_{s',r',a',m',\tilde{a}}p_{\theta_0}(s',r'|s_t,a',m')\pi_0(a'|s_t)
    p_{\theta_0}(m'|s_t,\tilde{a}) \pi_e(\tilde{a}|s_t)\triangledown_{\theta}\log p_{\theta_0}(s',r'|s_t,a',m') =0,
\end{multline*} we further represent $D_4^{(2)}$ as
\begin{multline*}
    \lim_{T\to \infty}\frac{1}{T}\sum_{t=0}^{T-1} \sum_{\tau_t,\bar{a}^*_{t}} \sum_{m',a'}p_{\theta_0}(m'|s_t,a')r_{\theta_0}(s_t,a',m')\pi_0(a'|s_t)\\\times\prod_{j=0}^{t}p_{\theta_0}(s_{j+1},r_j|s_j,a_j,m_j)\pi_0(a_j|s_j)p_{\theta_0}(m_j|s_j,a_j^*)\pi_e(a_j^*|s_j)\\
    \times \sum_{j=0}^{t}\triangledown_{\theta} \log p_{\theta_0}(s_{j+1},r_j|s_j,a_j,m_j)\nu_{\theta_0}(s_0).
\end{multline*} 
Note that we change the subscript of the summations accordingly. Then, following the steps we processed to derive the $D_1$, we first show that 
\begin{align*}
    \lim_{T\to \infty}\frac{1}{T}\sum_{t=0}^{T-1} \sum_{\tau_t,\bar{a}^*_{t}}\eta^{G_0}\prod_{j=0}^{t}p_{\theta_0}(s_{j+1},r_j|s_j,a_j,m_j)\pi_0(a_j|s_j)p_{\theta_0}(m_j|s_j,a_j^*)\pi_e(a_j^*|s_j) \\\times\sum_{j=0}^{t}\triangledown_{\theta} \log p_{\theta_0}(s_{j+1},r_j|a_j,s_j,m_j)\nu_{\theta_0}(s_0)=0.
\end{align*}
Therefore, 
\begin{multline*}
    D_4^{(2)} = \lim_{T\to \infty}\frac{1}{T}\sum_{t=0}^{T-1} \sum_{\tau_t,\bar{a}^*_{t}} \Big[\sum_{m',a'}p_{\theta_0}(m'|s_t,a')r_{\theta_0}(s_t,a',m')\pi_0(a'|s_t)-\eta^{G_0}\Big]\\\prod_{j=0}^{t}p_{\theta_0}(s_{j+1},r_j|s_j,a_j,m_j)\pi_0(a_j|s_j)p_{\theta_0}(m_j|s_j,a_j^*)\pi_e(a_j^*|s_j)
    \sum_{j=0}^{t}\triangledown_{\theta} \log p_{\theta_0}(s_{j+1},r_j|a_j,s_j,m_j)\nu_{\theta_0}(s_0).
\end{multline*}
Similar to $\star\star$, we have that
\begin{equation*}
    \sum_{s_0,\tau_{t-1},\bar{a}^*_{t-1}}\prod_{j=0}^{t-1}p_{\theta_0}(s_{j+1},r_j|s_j,a_j,m_j)\pi_0(a_j|s_j)p_{\theta_0}(m_j|s_j,a_j^*)\pi_e(a_j^*|s_j)
    \nu_{\theta_0}(s_0) \stackrel{\star\star\star}{=} p_{\theta_0}^{G}(s_t).
\end{equation*} 
Next, using the equality properties $\star$ and $\star\star\star$, together with the definition of $Q^{G_0}(s,a,m)$ and the trick of score functions, we can show that
    \begin{multline}\label{D4_2_crucial}
        D_4^{(2)} = \lim_{T\to \infty}\frac{1}{T}\sum_{j=0}^{T-1}\sum_{s_{j+1},r_j,a_j,s_j,m_j,a^*_j}\Big[\sum_{m',a'}p_{\theta_0}(m'|s_j,a')r_{\theta_0}(s_j,a',m')\pi_0(a'|s_j)-\eta^{G_0}+ \Mean^{G}_{a,m}Q^{G_0}(s_{j+1},a,m)\Big]\\
        \times p_{\theta_0}(s_{j+1},r_j|s_j,a_j,m_j)\pi_0(a_j|s_j)p_{\theta_0}(m_j|s_j,a_j^*)\pi_e(a_j^*|s_j)p^{G}(s_j)
    \triangledown_{\theta} \log p_{\theta_0}(s_{j+1},r_j|a_j,s_j,m_j).
    \end{multline}

Then, following the steps in deriving $D_1$, we have that
\begin{multline*}
    D_4^{(2)} = \lim_{T\to \infty}\frac{1}{T}\sum_{j=0}^{T-1}\sum_{s_{j+1},r_j,a_j,s_j,m_j}
    \omega_{\theta_0}^{G}(s_j)\frac{\sum_{a'}p_{\theta_0}(m_j|s_j,a')\pi_e(a'|s_j)}{p_{\theta_0}(m_j|s_j,a_j)}\frac{\pi_0(a_j|s_j)}{\pi_{b,\theta_0}(a_j|s_j)}\\
    \times \Big[\sum_{m',a'}p_{\theta_0}(m'|s_j,a')r_{\theta_0}(s_j,a',m')\pi_0(a'|s_j)-\eta^{G_0}+ \Mean^{G}_{a,m}Q^{G_0}(s_{j+1},a,m)-Q^{G_0}(s_{j},a_j,m_j)\Big]\\
    \times p_{\theta_0}(s_{j+1},r_j|s_j,a_j,m_j)p_{\theta_0}(m_j|s_j,a_j)\pi_{b,\theta_0}(a_j|s_j)p^{\pi_{b}}(s_{j})\triangledown_{\theta} \log p_{\theta_0}^{\pi_b}(s_{j+1},r_j,m_j,a_j,s_j).   
\end{multline*}
Finally, combining the fact that the expectation of a score function is zero and the Markov property, we have that
\begin{multline}
    D_4^{(2)} = \Mean\Big[\Big\{\sum_{m',a'}p_{\theta_0}(m'|S,a')r_{\theta_0}(S,a',m')\pi_0(a'|S)-\eta^{G_0}+ \Mean^{G}_{a,m}Q^{G_0}(S',a,m)-Q^{G_0}(S,A,M)\Big\}\\
    \times \omega_{\theta_0}^{G}(S)\frac{\sum_{a'}p_{\theta_0}(M|S,a')\pi_e(a'|S)}{p_{\theta_0}(M|S,A)}\frac{\pi_0(A|S)}{\pi_{b,\theta_0}(A|S)}S(\bar{O}_{T-1})\Big].
\end{multline}

\textbf{Part III ($D_4^{(3)}$).}
Following the same steps used in deriving the equation (\ref{D4_2_crucial}), we can show that,
    \begin{multline}\label{D4_3_1}
        D_4^{(3)} = \lim_{T\to \infty}\frac{1}{T}\sum_{j=0}^{T-1}\sum_{a_j,s_j,m_j,a^*_j}\sum_{s_{j+1},r_j}\Big[\sum_{m',a'}p_{\theta_0}(m'|s_j,a')r_{\theta_0}(s_j,a',m')\pi_0(a'|s_j)-\eta^{G_0}+ \Mean^{G}_{a,m}Q^{G_0}(s_{j+1},a,m)\Big]\\
        \times p_{\theta_0}(s_{j+1},r_j|s_j,a_j,m_j)\pi_0(a_j|s_j)p_{\theta_0}(m_j|s_j,a_j^*)\pi_e(a_j^*|s_j)p^{G}(s_j)
    \triangledown_{\theta} \log p_{\theta_0}(m_j|s_j,a_j^*).
    \end{multline}
Based on the definition of $Q^{G_0}$ and the corresponding Bellman equation, we have that 
\begin{align*}
    \Mean\Big[\sum_{m',a'}p_{\theta_0}(m'|s_j,a')r_{\theta_0}(s_j,a',m')\pi_0(a'|s_j)-\eta^{G_0}+ \Mean^{G}_{a,m}Q^{G_0}(s_{j+1},a,m)|s_j,a_j,m_j\Big] = Q^{G_0}(s_j,a_j,m_j).
\end{align*}
Therefore, (\ref{D4_3_1}) can be rewritten as
    \begin{multline*}
        \lim_{T\to \infty}\frac{1}{T}\sum_{j=0}^{T-1}\sum_{s_j,m_j,a_j}\big[\sum_{a'} Q^{G_0}(s_j,a',m_j)\pi_0(a'|s_j)\big]p_{\theta_0}(m_j|s_j,a_j)\pi_e(a_j|s_j)p^{G}(s_j)
    \triangledown_{\theta} \log p_{\theta_0}(m_j|s_j,a_j).
    \end{multline*}
Notice that
\begin{align*}
    &\begin{aligned}
        \lim_{T\to \infty}\frac{1}{T}\sum_{j=0}^{T-1}\sum_{s_j,m_j,a_j}\Big\{\sum_{a',m_j^*}Q^{G_0}(s_j,a',m_j^*)p_{\theta_0}(m_j^*|s_j,a_j)\pi_0(a'|s_j)\Big\}p_{\theta_0}(m_j|s_j,a_j)\pi_e(a_j|s_j)p^{G}(s_j)
    \\\times \triangledown_{\theta} \log p_{\theta_0}(m_j|s_j,a_j)
    \end{aligned}\\
    =&\begin{aligned}\lim_{T\to \infty}\frac{1}{T}\sum_{j=0}^{T-1}\sum_{s_j,a_j}\Big\{\sum_{a',m_j^*}Q^{G_0}(s_j,a',m_j^*)p_{\theta_0}(m_j^*|s_j,a_j)\pi_0(a'|s_j)\Big\}
        \pi_e(a_j|s_j)p^{G}(s_j)\sum_{m_j}p_{\theta_0}(m_j|s_j,a_j)
    \\\times \triangledown_{\theta} \log p_{\theta_0}(m_j|s_j,a_j
    \end{aligned})\\
    =&0.
\end{align*}
Therefore, we have that
\begin{multline*}
        D_4^{(3)} = \lim_{T\to \infty}\frac{1}{T}\sum_{j=0}^{T-1}\sum_{s_j,m_j,a_j}\Big\{\big[\sum_{a'} Q^{G_0}(s_j,a',m_j)\pi_0(a'|s_j)\big]-\sum_{a',m_j^*}Q^{G_0}(s_j,a',m_j^*)p_{\theta_0}(m_j^*|s_j,a_j)\pi_0(a'|s_j)\Big\}\\
        \times p_{\theta_0}(m_j|s_j,a_j)\pi_e(a_j|s_j)p^{G}(s_j)
    \triangledown_{\theta} \log p_{\theta_0}(m_j|s_j,a_j).
\end{multline*}
Following the same steps we used in getting the final expression of $D_4^{(1)}$, we can show that
\begin{multline}
    D_4^{(3)} = \Mean\Big[\omega_{\theta_0}^{G}(S)\frac{\pi_e(A|S)}{\pi_{b,\theta_0}(A|S)}\Big\{[\sum_{a'} Q^{G_0}(S,a',M)\pi_0(a'|S)\big]-\sum_{a',m'}Q^{G_0}(S,a',m')p_{\theta_0}(m'|S,A)\pi_0(a'|S)\Big\}S(\bar{O}_{T-1})\Big].
\end{multline}
Combining $D_4^{(1)}$, $D_4^{(2)}$, and $D_4^{(3)}$, we have that
\begin{multline*}
     D_4 = \Mean\Big[\omega^{G}(S)\frac{\pi_0(A|S)}{\pi_{b}(A|S)}\Big[\Big\{R-\Mean_{m}r(S,A,m)\Big\}+\rho(S,A,M)\Big\{\Mean^{\pi_0}_{a',m}r(S,a',m)+ \Mean^{G}_{a,m}Q^{G_0}(S',a,m)\\-Q^{G_0}(S,A,M)-\eta^{G_0}\Big\}\Big]
    + \omega^{G}(S)\frac{\pi_e(A|S)}{\pi_{b}(A|S)}\sum_a \pi_0(a|S)\Big\{Q^{G_0}(S,a,M)-\sum_m p(m|S,A)Q^{G_0}(S,a,m)\Big\}\Big].
\end{multline*}

Since $(S, A, M, R, S')$ is any arbitrary transaction tuple follows the corresponding distribution, we have that
\begin{multline*}
     D_4 = \Mean\Big[\frac{1}{T}\sum_{t=0}^{T-1}\omega^{G}(S_t)\frac{\pi_0(A_t|S_t)}{\pi_{b}(A_t|S_t)}\Big[\Big\{R_t-\Mean_{m}r(S_t,A_t,m)\Big\}+\rho(S_t,A_t,M_t)\Big\{\Mean^{\pi_0}_{a',m}r(S_t,a',m)+ \Mean^{G}_{a,m}Q^{G_0}(S_{t+1},a,m)\\-Q^{G_0}(S_t,A_t,M_t)-\eta^{G_0}\Big\}\Big]
    + \omega^{G}(S_t)\frac{\pi_e(A_t|S_t)}{\pi_{b}(A_t|S_t)}\sum_a \pi_0(a|S_t)\Big\{Q^{G_0}(S_t,a,M_t)-\sum_m p(m|S_t,A_t)Q^{G_0}(S_t,a,m)\Big\}\Big].
\end{multline*}

\subsubsection{Efficient Function}
Given $C_3$, $D_3$, and $D_4$, the efficient influence function for $\textrm{DDE}_{\theta_0}(\pi_e,\pi_0)$ is $\eta^{\pi_{e,0}}-\eta^{G_0}+I_3-I_4$, where
\begin{multline*}
    I_3 = \Mean\Big[\omega^{\pi_e}(S)\Big\{\frac{\pi_0(A|S)}{\pi_{b,\theta_0}(A|S)}[R-\Mean_{m}r_{\theta_0}(S,A,m)]+\frac{\pi_e(A|S)}{\pi_{b,\theta_0}(A|S)}\{\sum_{a'}\Mean_{m\sim p_{\theta_0}(\bullet|S,a')}r_{\theta_0}(S,a',m)\pi_0(a'|S)\\+\Mean^{\pi_e}_{a,m} Q^{\pi_{e,0}}(S',a,m)
	-\Mean_{m} Q^{\pi_{e,0}}(S,A,m)-\eta^{\pi_{e,0}}\}\Big\}\Big],
\end{multline*}and
\begin{multline*}
     I_4 = \Mean\Big[\omega^{G}(S)\frac{\pi_0(A|S)}{\pi_{b}(A|S)}\Big[\Big\{R-\Mean_{m}r(S,A,m)\Big\}+\rho(S,A,M)\Big\{\Mean^{\pi_0}_{a',m}r(S,a',m)+ \Mean^{G}_{a,m}Q^{G_0}(S',a,m)\\-Q^{G_0}(S,A,M)-\eta^{G_0}\Big\}\Big]
    + \omega^{G}(S)\frac{\pi_e(A|S)}{\pi_{b}(A|S)}\sum_a \pi_0(a|S)\Big\{Q^{G_0}(S,a,M)-\sum_m p(m|S,A)Q^{G_0}(S,a,m)\Big\}\Big].
\end{multline*}

\subsection{EIF for Delayed Mediator Effect}
Delayed Mediator Effect (DME) can be represented as
\begin{multline}\label{DME}
    \textrm{DME}(\pi_e,\pi_0) = \lim_{T\to \infty}\frac{1}{T}\sum_{t=0}^{T-1} \sum_{\tau_t}r_t p(s_{t+1},r_t|s_t,a_t,m_t)p(m_t|s_t,a_t)\pi_0(a_t|s_t)\\
    \times \Big\{\sum_{\bar{a}^*_{t-1}}\prod_{j=0}^{t-1}p(s_{j+1},r_j|s_j,a_j,m_j)\pi_0(a_j|s_j)p(m_j|s_j,a_j^*)\pi_e(a_j^*|s_j)-
    \prod_{j=0}^{t-1}p^{\pi_0}(s_{j+1},r_j,m_j,a_j|s_j)\Big\}\nu(s_0).
\end{multline}
Taking the derivative of $\textrm{DME}_{\theta_0}(\pi_e,\pi_0)$, we get that 
\begin{eqnarray*}
	\frac{\partial \textrm{DME}_{\theta_0}(\pi_e,\pi_0)}{\partial \theta_{0}} = C_4 + D_4 - D_5,
\end{eqnarray*} where
\begin{eqnarray*}
    C_4 = (\ref{DME})\times \triangledown_{\theta} \log(\nu_{\theta_0}(s_0)) = \Mean[\textrm{DME}_{\theta_0}(\pi_e,\pi_0) \times S(\bar{O}_{T-1})]= \Mean[(\eta^{G_0}-\eta^{\pi_0}) \times S(\bar{O}_{T-1})],
\end{eqnarray*}
$D_4$ is derived in Appendix \ref{Derivation:I4}, and 
\begin{eqnarray*}
	 D_5 = \lim_{T\to \infty}\frac{1}{T}\sum_{t=0}^{T-1} \sum_{\tau_t} r_t \prod_{j=0}^{t}p_{\theta_0}^{\pi_0}(s_{j+1},r_j,m_j,a_j|s_j) \sum_{j=0}^{t}\left[ \triangledown_{\theta} \log p_{\theta_0}^{\pi_0}(s_{j+1},r_j,m_j,a_j|s_j)]\right.\times \nu_{\theta_0}(s_0),
\end{eqnarray*}
Notice that $D_5$ is similar as $D_1$, and can be derived similarly as $D_1$ by replacing the $\pi_e$ in $D_1$ with $\pi_0$. Therefore, with the definition of $Q^{\pi_0}(s,a,m)$, we can show that
\begin{eqnarray*}
     D_5 = \Mean\Big[\omega^{\pi_0}(S)\frac{\pi_0(A|S)}{\pi_{b,\theta_0}(A|S)}\{R + \sum_{a'}\Mean_{m} Q^{\pi_0}(S',a',m)\pi_0(a'|S') - \Mean_{m} Q^{\pi_0}(S,A,m)-\eta^{\pi_0}\}S(\bar{O}_{T-1})\Big].
\end{eqnarray*}
Since $(S, A, M, R, S')$ is any arbitrary transaction tuple follows the corresponding distribution, we have that
\begin{eqnarray*}
     D_5 = \Mean\Big[\frac{1}{T}\sum_{t=0}^{T-1}\omega^{\pi_0}(S_t)\frac{\pi_0(A_t|S_t)}{\pi_{b,\theta_0}(A_t|S_t)}\{R_t + \sum_{a'}\Mean_{m} Q^{\pi_0}(S_{t+1},a',m)\pi_0(a'|S_{t+1}) - \Mean_{m} Q^{\pi_0}(S_t,A_t,m)-\eta^{\pi_0}\}S(\bar{O}_{T-1})\Big].
\end{eqnarray*}

\subsubsection{Efficient Function}
Given $C_4$, $D_4$, and $D_5$, the efficient influence function for $\textrm{DME}_{\theta_0}(\pi_e,\pi_0)$ is $\eta^{G_0}-\eta^{\pi_0}+I_4-I_5$, where
\begin{multline*}
     I_4 = \Mean\Big[\omega^{G}(S)\frac{\pi_0(A|S)}{\pi_{b}(A|S)}\Big[\Big\{R-\Mean_{m}r(S,A,m)\Big\}+\rho(S,A,M)\Big\{\Mean^{\pi_0}_{a',m}r(S,a',m)+ \Mean^{G}_{a,m}Q^{G_0}(S',a,m)\\-Q^{G_0}(S,A,M)-\eta^{G_0}\Big\}\Big]
    + \omega^{G}(S)\frac{\pi_e(A|S)}{\pi_{b}(A|S)}\sum_a \pi_0(a|S)\Big\{Q^{G_0}(S,a,M)-\sum_m p(m|S,A)Q^{G_0}(S,a,m)\Big\}\Big].
\end{multline*} and 
\begin{eqnarray*}
     I_5 = \Mean\Big[\omega^{\pi_0}(S)\frac{\pi_0(A|S)}{\pi_{b,\theta_0}(A|S)}\{R + \sum_{a'}\Mean_{m} Q^{\pi_0}(S',a',m)\pi_0(a'|S') - \Mean_{m} Q^{\pi_0}(S,A,m)-\eta^{\pi_0}\}\Big].
\end{eqnarray*}

%% file: Appendix/Toy_Examples.tex
\section{Settings for Numerical Examples}
\subsection{Toy Example 1 \& Toy Example 2}\label{toy settings}
\textbf{Settings.} We consider a scenario with discrete states, actions, mediators, and rewards. We set time $T=50$, and $S_{0}$ for each trajectory is sampled from a Bernoulli distribution with a mean probability of $0.5$. Denote the sigmoid function as $expit(\cdot)$. Following the behavior policy, the action $A_{t} \in \{0,1\}$ is sampled from a Bernoulli distribution, where $\prob(A_{t} = 1|S_{t}) = expit(1.0-2.0S_{t})$. Observing $S_t$ and $A_{t}$, the mediator $M_{t} \in \{0,1\}$ is drawn from a Bernoulli distribution with $\prob(M_{t} = 1|S_{t}, A_{t}) = expit(1.0-1.5S_{t}+2.5A_{t})$. The distributions of $R_{t}$ and $S_{t+1}$ are both Bernoulli and conditional on $S_{t}$, $A_{t}$, and $M_{t}$. Specifically, the reward distribution of $R_t\in\{0,10\}$ satisfies that $\prob(R_{t} = 10|S_{t}, A_{t}, M_{t}) = expit(1.0+2.0S_{t}-1.0A_{t}-2.5M_{t})$, while the distribution of next state $S_{t+1}\in \{0,1\}$ satisfies that $\prob(S_{t+1} = 1|S_{t}, A_{t}, M_{t}) = expit(.5+3.0S_{t}-2.5A_{t}-.5M_{t})$. We are interested in estimating the treatment effect of the target policy $\pi_e$, which applies a treatment with $\prob(A_{t} = 1|S_{t}) = expit(1.5+1.0S_{t})$, compared to the control policy $\pi_0$, which always applies no treatment (i.e., $\prob(A_{t} = 1)=0$). Monte Carlo (MC) simulations are used to calculate the oracle distributions of $\omega^{(\cdot)}$ and $Q^{(\cdot)}$, and the oracle values of $\eta^{(\cdot)}$, $\textrm{IDE}(\pi_e,\pi_0)$, $\textrm{IME}(\pi_e,\pi_0)$, $\textrm{DDE}(\pi_e,\pi_0)$, and $\textrm{DME}(\pi_e,\pi_0)$. Based on 40K simulated trajectories with 1K observations each, we obtained that $\textrm{IDE} = -1.277$, $\textrm{IME} = -1.222$, $\textrm{DDE} = -2.982$, and $\textrm{DME} = -.085$. Considering the true distributions of $Q^{\pi_e}$, $Q^{\pi_e,a0}$, $Q^{\pi_e,a0*}$, $Q^{G}$, and $Q^{a_0}$, we approximate each of them by assuming linear equation models \citep{shi2020statistical}. 

\textbf{Misspecification.} To misspecify the $\omega^{\pi_e}$, we add .25 to the $\omega^{\pi_e}(S_{t}=1)$ and subtract .25 from the $\omega^{\pi_e}(S_{t}=0)$. Similarly, we subtract .3 from the $\omega^{a_0}(S_{t}=1)$ ($\omega^{G}(S_{t}=1)$) and add .3 to the $\omega^{a_0}(S_{t}=0)$ ($\omega^{G}(S_{t}=0)$). For $Q^{(\cdot)}$, and $r$ functions, we inject Gaussian noises into each parameter involved in the true model. For the misspecification of $p_m$ and $\pi_b$, we multiply the true value by a random variable drawn from a bounded uniform distribution and then clip the probabilities to ensure that they are within the range of .01 and .99.

\subsection{Semi-Synthetic Data}\label{synthetic settings}
The spaces for reward, state, and mediator are continuous, and the action space is binary. Specifically, the semi-synthetic data is generated as follows. The initial states are i.i.d. sampled from the standard normal distribution. $\pi_b$ follows a Bernoulli distribution, satisfying that $\prob(A_t = 1) = \prob(A_t = 0) = .5$. We consider a 2-dimensional mediator, where $M_{t,1}$ and $M_{t,2}$ are independent and normally distributed with a standard deviation of 2. While the mean of $M_{t,1}$ is $\sqrt{|S_t|}+(A_t-.5)$, the mean of $M_{t,2}$ is $.5(A_t-.5)*\sqrt{|S_t|}-.5 S_t$. We set $R_t = S_{t+1}$, and $R_t$ is drwan from a normal distribution with a mean of $.75[S_t + \sqrt{|S_t|}+(1+\sqrt{|M_{t1}|+|M_{t2}|})(A_t-.5)]+1.5(M_{t1}+M_{t2})$ and a standard deviation of 2. The control policy always takes action $A_t = 0$, while the target policy follows a Bernoulli distribution with $\prob(A_t = 1) = expit(.7*S_{t})$. Monte Carlo (MC) simulations are used to calculate the oracle the oracle values of $\eta^{(\cdot)}$, $\textrm{IDE}(\pi_e,\pi_0)$, $\textrm{IME}(\pi_e,\pi_0)$, $\textrm{DDE}(\pi_e,\pi_0)$, and $\textrm{DME}(\pi_e,\pi_0)$. Based on 20K simulated trajectories with 6400 observations each, we obtained that $\textrm{IDE} = 2.680$, $\textrm{IME} = 3.654$, $\textrm{DDE} = 1.244$, and $\textrm{DME} = .689$.

\section{Baseline Estimators} \label{baseline}
\subsection{Baseline\_DM}
Following the definitions of the direct and indirect effect in \citet{robins1992identifiability}, the first set of baseline estimators we considered are constructed by inputting estimated probability functions directly. Specifically, the estimator for $\textrm{IDE}(\pi_e,\pi_0)$ is
\begin{equation*}
    \frac{1}{NT}\sum_{i,t,a,m}\Big\{r(S_{i,t},a,m)-\sum_{a'}r(S_{i,t},a',m)\pi_0(a'|S_{i,t})\Big\}p(m|S_{i,t},a)\pi_e(a|S_{i,t}),
\end{equation*} and the estimator for $\textrm{IME}(\pi_e,\pi_0)$ is 
\begin{equation*}
    \frac{1}{NT}\sum_{i,t,a,m} r(S_{i,t},a,m)\Big\{\sum_{a'}p(m|S_{i,t},a')\pi_e(a'|S_{i,t})-p(m|S_{i,t},a)\Big\}\pi_0(a|S_{i,t}).
\end{equation*} 

\subsection{Baseline\_IPW}
The inverse probability weighting estimators, proposed in \citet{lange2012simple} and \citet{hong2010ratio}, are the second set of baseline estimators. Specifically, the estimator for $\textrm{IDE}(\pi_e,\pi_0)$ is
\begin{equation*}
    \frac{1}{NT}\sum_{i,t}\frac{\pi_e(A_{i,t}|S_{i,t}) - \pi_0(A_{i,t}|S_{i,t})\rho(S,A,M)}{\pi_b(A_{i,t}|S_{i,t})}R_{i,t}
\end{equation*} and the estimator for $\textrm{IME}(\pi_e,\pi_0)$ is 
\begin{equation*}
   \frac{1}{NT}\sum_{i,t}\frac{\pi_0(A_{i,t}|S_{i,t})}{\pi_b(A_{i,t}|S_{i,t})}\{\rho(S,A,M)-1\}R_{i,t}.
\end{equation*}

\subsection{Baseline\_MR}
Finally, the multiply robust estimators, proposed in \citet{tchetgen2012semiparametric}, are the third set of baseline estimators. Specifically, the estimator for $\eta^{\pi}$ is
\begin{equation*}
    \frac{1}{NT}\sum_{i,t}\frac{\pi(A_{i,t}|S_{i,t})}{\pi_b(A_{i,t}|S_{i,t})}[R_{i,t}-r(S_{i,t},A_{i,t})]+r(S_{i,t},\pi)
\end{equation*} and the estimator for $\eta^{G_e}$ is 
\begin{align*}
    \frac{1}{NT}\sum_{i,t}&\Big\{\frac{\pi_0(A_{i,t}|S_{i,t})}{\pi_b(A_{i,t}|S_{i,t})}\rho(S,A,M)[R_{i,t}-r(S_{i,t},M_{i,t},A_{i,t})]+\\&\frac{\pi_e(A_{i,t}|S_{i,t})}{\pi_b(A_{i,t}|S_{i,t})}[r(S_{i,t},\pi_0,M_{i,t})-\int_{m}p(m|S_{i,t},A_{i,t})r(S_{i,t},\pi_0,m)] + r(S_{i,t},\pi_0,\pi_e)\Big\},
\end{align*} where $r(S_{i,t},\pi_0,\pi_e) = \sum_{a,a',m}r(S_{i,t},a',m)\pi_0(a'|S_{i,t})p(m|S_{i,t},a)\pi_e(a|S_{i,t})$. Plugging in these estimators, we then get the corresponding estimators for $\textrm{IDE}(\pi_e,\pi_0)$ and $\textrm{IME}(\pi_e,\pi_0)$. 

%% file: Appendix/optimal_policy.tex
\section{Estimating Optimal Policy}\label{opt_policy}
To estimate the optimal policy, we first estimate a $Q$ function based on the observational data, following the same methods described in Section \ref{nuisance}. Specifically, let 
\begin{equation*}
    Q(s,a,m) = \sum_{t\geq0}\Mean[R_t-\eta],
\end{equation*} which leads to a Bellmen equation model, such that
\begin{equation*}
    Q(S_t,A_t,M_t) = \sum\Mean[R_t+ \sum_{a}\int_{m}\Mean Q(S_{t+1}, a, m)-\eta].
\end{equation*}We then approximate the $Q$ function using linear sieves. Finally, the estimated optimal policy is defined as 
\begin{equation*}
    \hat{\pi}^{opt}(s) = \arg\max_{a\in \mathcal{A}}\sum_{a}\int_m\Mean \hat{Q}(s,a,m).
\end{equation*}

It is worth noting that we used cross-validation to estimate the ATE of $\hat{\pi}_{opt}$. To be more specific, we divide the observed trajectories into two folds. In each round $k$, we first estimate the $\hat{\pi}^k_{opt}$ based on the trajectories within fold k, and then estimate the ATE of $\hat{\pi}^k_{opt}$ on another fold of trajectories.

%% file: Appendix/Additional_Experiments.tex
\section{Additional Numerical Experiments} \label{additional_Exp}
In this section, we conducted additional numerical experiments to evaluate the estimation performance of the proposed estimators under diverse settings.

\subsection{Performance Under Tabular Setting}\label{additional:tabular}
\begin{figure}[ht]
    \centering
    \includegraphics[width=.85\columnwidth]{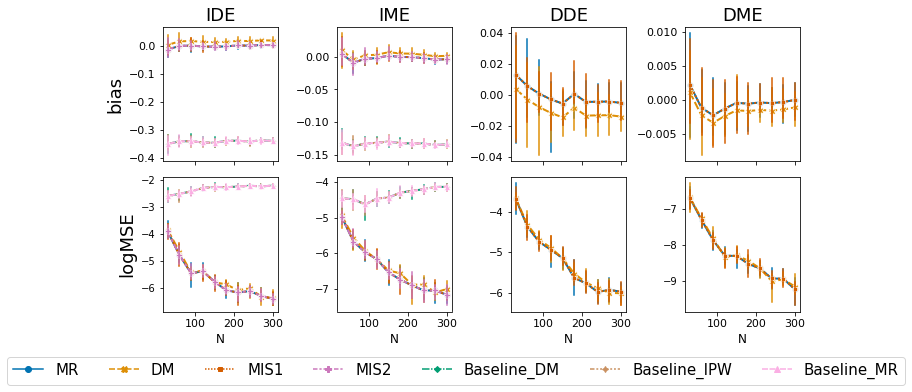}
    \caption{The bias and logMSE of various estimators under the setting with discrete spaces. The results are aggregated over 200 random seeds.}
    \label{fig:additional1}
\end{figure}

First, we investigate the estimation performance under the tabular setting as discussed in the toy examples (see Section \ref{toy_2}). Results are summarized in Figure \ref{fig:additional1}. Similar to what we concluded from the semi-synthetic simulation in Section \ref{semi_syn}, all three sets of the proposed estimators provide unbiased estimation for all four effect components, with the MSE decreasing continuously as the sample size increases. In contrast, all baseline estimators ignoring the fact of state transition continue to yield biased estimates no matter how large the sample size is.

\subsection{Impact of Variance on Performance} \label{additional:different variance}
\begin{figure}[ht]
    \centering
    \includegraphics[width=.85\columnwidth]{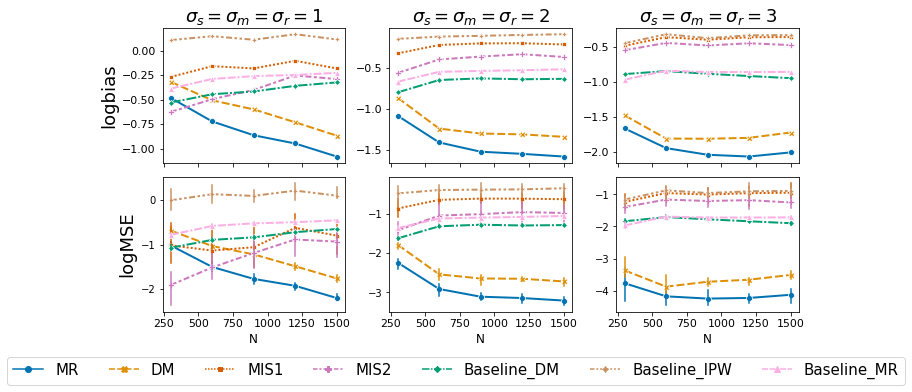}
    \caption{The logBias and the logMSE of estimators for IDE, under different data generation scenarios. Fix T = 50.}
    \label{fig:additional2_1}
\end{figure}

\begin{figure}[ht]
    \centering
    \includegraphics[width=.85\columnwidth]{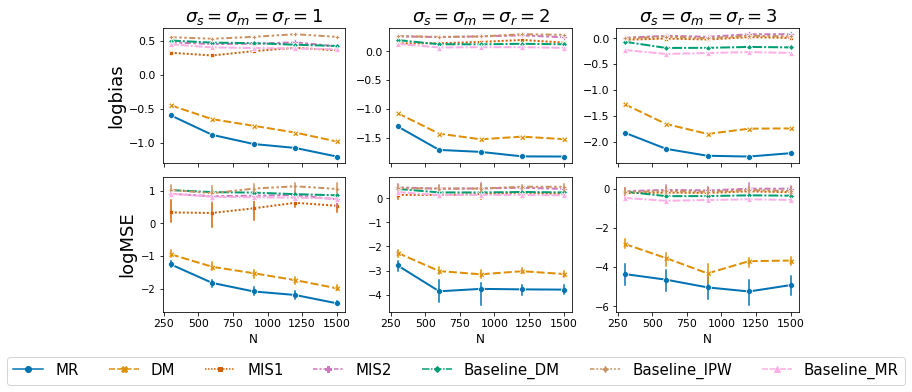}
    \caption{The logBias and the logMSE of estimators for IME, under different data generation scenarios. Fix T = 50.}
    \label{fig:additional2_2}
\end{figure}

\begin{figure}[ht]
    \centering
    \includegraphics[width=.65\columnwidth]{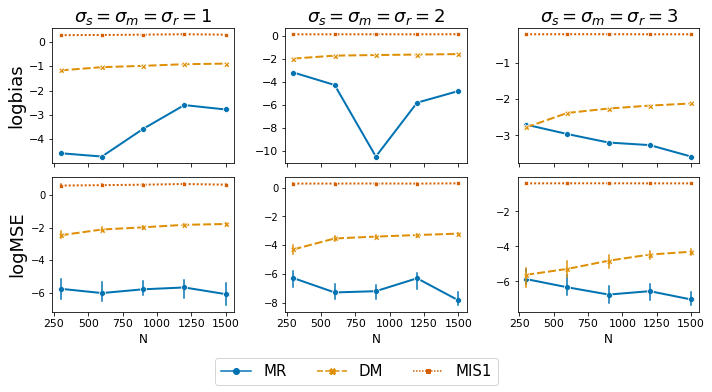}
    \caption{The logBias and the logMSE of estimators for DDE, under different data generation scenarios. Fix T = 50.}
    \label{fig:additional2_3}
\end{figure}

\begin{figure}[ht]
    \centering
    \includegraphics[width=.65\columnwidth]{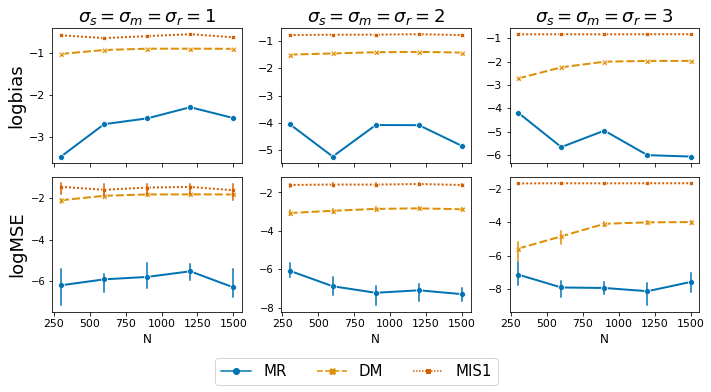}
    \caption{The logBias and the logMSE of estimators for DME, under different data generation scenarios. Fix T = 50.}
    \label{fig:additional2_4}
\end{figure}
Second, under the semi-synthetic data setting that we used in Section \ref{semi_syn}, we further vary the variances of the random noises of states, mediators, and rewards. Specifically, we considered three settings where the standard deviation of the random noise takes values of 1, 2, and 3, i.e., $\sigma_m = \sigma_s =  \sigma_r = 1$, $\sigma_m = \sigma_s =  \sigma_r = 2$, and $\sigma_m = \sigma_s = \sigma_r = 3$. Results are summarized in Figure \ref{fig:additional2_1}, Figure \ref{fig:additional2_2}, Figure \ref{fig:additional2_3}, and Figure \ref{fig:additional2_4}. For each choice of the variance, the proposed MR estimators always achieve the smallest bias and MSE. While all the baseline estimators and MIS estimators continue to provide biased estimates with non-decreasing MSE, we observe that the difference between DM and MR estimators becomes smaller as the variance increases.

\subsection{Multidimensional State and Mediator} \label{additional:multi}
\begin{figure}[ht]
    \centering
    \includegraphics[width=.85\columnwidth]{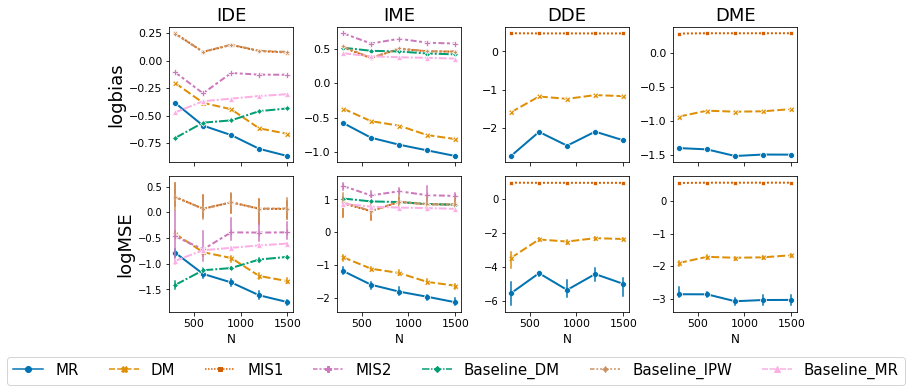}
    \caption{The logBias and the logMSE of estimators, under settings with multidimensional state. Fix T = 100.}
    \label{fig:additional3_1}
\end{figure}

\begin{figure}[ht]
    \centering
    \includegraphics[width=.85\columnwidth]{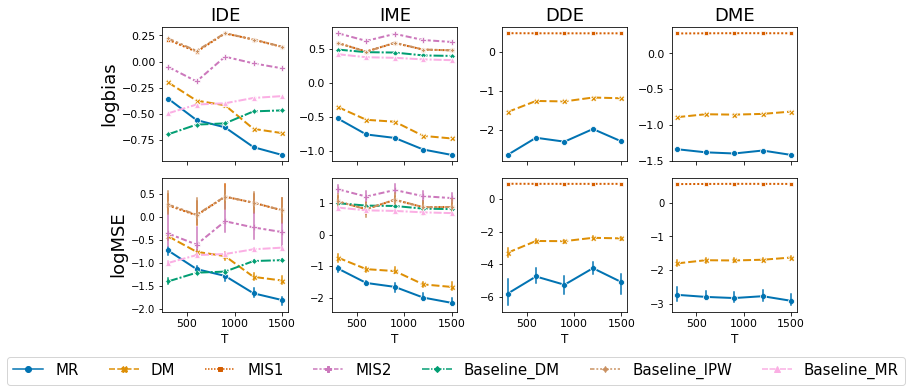}
    \caption{The logBias and the logMSE of estimators, under settings with multidimensional state. Fix N = 100.}
    \label{fig:additional3_2}
\end{figure}
Finally, we investigate the estimation performance in a more complicated system where both the state and mediator are multi-dimensional. Specifically, we set the number of state variables to be 2 and the number of mediators to be 2. The data generating mechanism is as follows:
\begin{align*} 
S_{0, i} & \sim \mathcal{N}(0,1), \text { for } i \in\{1,2\}; \\ A_t & \sim \pi_b=\operatorname{Bernoulli}(.5); \\ 
M_{t 1} & \sim \mathcal{N}\left(.5\left(\sqrt{\left|S_{t, 1}\right|}+\sqrt{\left|S_{t, 2}\right|}\right)+\left(A_t-.5\right), 1\right); \\ 
M_{t 2} & \sim \mathcal{N}\left(-.25\left(S_{t, 1}+S_{t, 2}\right)+.25\left(A_t-.5\right)\left(\sqrt{\left|S_{t, 1}\right|}+\sqrt{\left|S_{t, 2}\right|}\right), 1\right); \\
S_{t+1,1} & \sim \mathcal{N}\left(.75\left\{S_{t, 1}+\sqrt{\left|S_{t, 1}\right|}+\left(1+\sqrt{\left|M_{t 1}\right|+\left|M_{t 2}\right|}\right)\left(A_t-.5\right)\right\}+1.5\left\{M_{t 1}+M_{t 2}\right\}, 1\right); \\ 
S_{t+1,2} & \sim \mathcal{N}\left(.75\left\{S_{t, 2}+\sqrt{\left|S_{t, 2}\right|}+\left(1+\sqrt{\left|M_{t 1}\right|+\left|M_{t 2}\right|}\right)\left(A_t-.5\right)\right\}+1.5\left\{M_{t 1}+M_{t 2}\right\}, 1\right); \\ 
R_t & \sim \mathcal{N}\left(.75\left\{.5\left(S_{t, 1}+S_{t, 2}+\sqrt{\left|S_{t, 1}\right|}+\sqrt{\left|S_{t, 2}\right|}\right)+\left(1+\sqrt{\left|M_{t 1}\right|+\left|M_{t 2}\right|}\right)\left(A_t-.5\right)\right\}+\frac{3}{2}\left\{M_{t 1}+M_{t 2}\right\}, 1\right); \\ 
\pi_e & \sim \operatorname{Bernoulli}\left(\operatorname{expit}\left(.3\left(S_{t, 1}+S_{t, 2}\right)\right)\right); \\ 
a_0 & =0.
\end{align*}

The results are summarized in Figure \ref{fig:additional3_1} and Figure \ref{fig:additional3_2}, which show the same trend as the semi-synthetic simulation in Section \ref{semi_syn}.